\documentclass[sigconf]{acmart}
\usepackage[T1]{fontenc}
\usepackage{booktabs}
\usepackage{subcaption}
\usepackage{adjustbox}
\usepackage{graphicx}
\usepackage{tabularx}
\usepackage{longtable}
\usepackage{placeins}

\setcopyright{none}
\renewcommand\footnotetextcopyrightpermission[1]{}

\acmConference{}{}{}
\acmBooktitle{}
\acmPrice{}
\acmDOI{}
\acmISBN{}

\usepackage{tikz}
\usetikzlibrary{arrows.meta, shapes.geometric, positioning}
\usepackage{xeCJK}
\usepackage{zxjatype}
\setCJKsansfont{ipaexg.ttf}
\usepackage{array}
\usepackage{booktabs}
\usepackage{tabularx}
\usepackage{multirow}
\usepackage{arydshln}

\begin{document}
\raggedbottom
\title{Proposing Topic Models and Evaluation Frameworks for Analyzing Associations with External Outcomes: An Application to Leadership Analysis Using Large-Scale Corporate Review Data}
%
%
\author{Yura Yoshida}
\affiliation{
\institution{Accenture Japan Ltd}
\country{Japan}
}
\author{Masato Kanai}
\affiliation{
\institution{Accenture Japan Ltd}
\institution{Kyoto University Institute for the Future of Human Society}
\country{Japan}
}
\author{Masataka Nakayama}
\affiliation{
\institution{Kyoto University Institute for the Future of Human Society}
\country{Japan}
}
\author{Haruki Ohsawa}
\affiliation{
\institution{OpenWork Inc.}
\country{Japan}
}
\author{Yukiko Uchida}
\affiliation{
\institution{Kyoto University Institute for the Future of Human Society}
\country{Japan}
}
\author{Arata Yuminaga}
\author{Gakuse Hoshina}
\affiliation{
\institution{Accenture Japan Ltd}
\country{Japan}
}
\author{Nobuo Sayama}
\affiliation{
\institution{Integral Corporation}
\country{Japan}
}
\begin{abstract}
Analyzing topics extracted from text data in relation to external outcomes is important across a wide range of fields, including computational social science, organizational research, and marketing. However, in existing topic modeling methods, it is difficult to simultaneously achieve topic interpretability, which is important for interpreting relationships with external outcomes; topic specificity, defined as alignment with specific and concrete actions or characteristics; and polarity stance consistency, defined as the absence of mixed positive and negative evaluations within the same topic. Focusing on leadership analysis based on corporate review data, this study proposes a method that leverages large language models to generate topics with interpretability, topic specificity, and polarity stance consistency, while simultaneously introducing an evaluation framework suitable for external outcome analysis. The proposed evaluation framework explicitly positions topic specificity and polarity stance consistency, which have not been sufficiently addressed in the existing literature, as evaluation criteria, and examines the validity of automated evaluation methods applied to existing metrics. This framework enables a multidimensional examination of the characteristics of generated topics. Analyses using reviews posted by current and former employees on OpenWork, one of the largest corporate review platforms in Japan, showed that the proposed method simultaneously achieves interpretability, topic specificity, and polarity stance consistency. In analyses of external outcomes such as employee morale, the method generated topics with consistently higher explanatory power than existing methods. This study newly proposes extended methodological approaches and evaluation criteria for topic analysis aimed at understanding relationships with external outcomes, and demonstrates their potential to generalize to a wide range of application domains with similar analytical requirements.

\end{abstract}

\maketitle              
\section{Introduction}
\subsection{Leadership, Performance, and Employee Engagement}
Improving organizational performance has long been a central managerial objective, and in recent years, psychological aspects of employees, such as vitality and work engagement, have also been recognized as playing a key role in organizational performance~\cite{Harter2002,Judge2001}. Based on these findings, it is important to simultaneously examine factors that promote both performance outcomes and employees' psychological well-being.

In this context, leadership has been widely recognized as a key factor influencing both organizational performance and employee engagement. Leadership at multiple organizational levels, ranging from top executives to middle managers and direct supervisors, is associated with employees' high performance and favorable psychological states~\cite{Judge2004,Montano2017}.

Since the 1930s, leadership has been a central topic in management and psychology, and diverse theoretical frameworks have been developed, primarily in Western contexts~\cite{House1997,Solansky2017}. Early trait and behavioral theories later evolved into more complex frameworks, such as transformational and transactional leadership~\cite{DeRue2011}. Building on these foundations, various measurement approaches have been developed, including leadership scales and multidimensional frameworks~\cite{Avolio1999,WarnerSoderholm2020}. As a result, a substantial body of empirical research has accumulated, and meta-analytic studies have synthesized findings from individual studies (e.g.,~\cite{DeRue2011,Judge2004,Montano2017}). 

However, although leadership research has produced a substantial body of knowledge, several challenges remain. Existing evidence is heavily concentrated in Western cultural contexts, analyses often rely on specific theoretical frameworks, and cross-firm studies tend to abstract leadership characteristics into broad categories. These limitations motivate the need for data-driven, cross-company approaches that enable fine-grained analysis of leadership behaviors across diverse cultural contexts.

\subsection{Approach of This Study}
This study aims to address these three challenges.
To this end, we employ large-scale text data accumulated on the corporate review platform "OpenWork" and applies topic modeling to reviews posted by current and former employees, primarily focusing on Japanese companies~\cite{openwork}. This approach enables an examination of the relationships between leadership characteristics reflected in the reviews and corporate performance as well as employee morale. It is made possible by the combination of user-generated, objective cross-company data such as OpenWork reviews and recent advances in natural language processing techniques, including large language models (LLMs). The review data analyzed in this study reflect employees' spontaneous evaluations of their companies, enabling a detailed and realistic understanding of leadership practices in Japanese companies from the employees’ perspective. Advances in LLMs make it possible to conduct highly accurate and flexible analyses of large-scale unstructured data efficiently.

\section{Related Work}
\subsection{Limitations of Existing Leadership Research}
Despite the accumulation of empirical evidence reviewed above, at least three important challenges remain in leadership research.

First, leadership research has predominantly relied on studies conducted in Western organizations, as reflected in major theoretical frameworks and many meta-analyses~\cite{House1997,Solansky2017,DeRue2011,Judge2004,Montano2017,Schimmelpfennig2025}. Although some theories have discussed leadership characteristics expected in Japanese organizations (e.g.,~\cite{Misumi1985}), and cultural differences in other-orientation among high-ranking individuals between Western and East Asian societies have been documented~\cite{GobelMiyamoto2023}, systematic empirical evidence from non-Western contexts remains limited.

Second, leadership research has largely been structured around established theoretical frameworks (e.g.,~\cite{DeRue2011}). While theory-driven approaches have contributed to the validation and refinement of existing theories, they are prone to mismatches between theoretical frameworks and empirical methods~\cite{Dinh2014}, which may obscure influence processes that are not fully considered within a given theory. Although a small number of recent studies have begun to explore inductive approaches using free-text data (e.g.,~\cite{tonidandel2022leadership}), research that comprehensively identifies leadership characteristics through data-driven analyses of large-scale text remains scarce.

Third, insights derived from cross-firm leadership research tend to remain at an abstract level. This is partly because many cross-company studies rely primarily on meta-analytic approaches to integrate heterogeneous findings, which has been pointed out to encourage the abstraction of leadership characteristics into broad meta-categories, such as task-oriented and relationship-oriented leadership~\cite{Yukl2019}. While such abstraction facilitates cross-study comparison, it constrains detailed examination of specific leadership behaviors across firms.

\subsection{Technical Requirements for Outcome-Oriented Topic Modeling}
In applied settings considered in this study, technical requirements include topic representations that ensure interpretability, specificity and polarity stances  as well as an evaluation framework to assess these characteristics. To use topic modeling for analyzing relationships with external variables and to connect the findings to practical discussions, the extracted topics must be interpretable to humans and meaningfully linked to imaginable and specific behaviors, attributes, or actions. Moreover, when evaluative orientations—such as positive and negative stances—are mixed within a topic, interpretation of its relationship with external variables becomes difficult. In such cases, effects on external outcomes may cancel out, making consistency in polarity stance within topics a critical requirement. For example, even if a topic named "decision making" shows a positive correlation with firm performance, the topic may contain a mixture of positive documents praising rapid decision making and negative documents criticizing slow decision making. In such cases, opposing polarity stance may offset one another in their associations with external variables. As a result, even when a relationship between the topic and external variables is detected, it remains difficult to translate the result into actionable practical implications.
Accordingly, in applied settings considered in this study, technical requirements include topic representations that ensure interpretability, specificity and polarity stances as well as an evaluation framework to assess these characteristics.

\subsection{Limitations of Existing Topic Modelings}
Topic modeling aims to extract latent thematic structures embedded in collections of text and to summarize document corpora in a low-dimensional topic space, and has developed as a representative approach for large-scale text analysis. Early studies proposed probabilistic generative models based on the bag-of-words assumption, such as Latent Dirichlet Allocation (LDA)~\cite{blei2003lda}, which infer topic distributions from word co-occurrence patterns. However, when these topic modeling approaches are used as text-mining technologies,
several technical challenges remain. To address these challenges, some extensions of topic modeling have been proposed.



Probabilistic models based on the bag-of-words assumption, such as the Structural Topic Model (STM), have been widely used for estimating relationships between topics and external variables, as they provide a framework well suited for statistical inference that explicitly accounts for uncertainty~\cite{roberts2014stm}. 
The STM was developed to incorporate document-level covariates into topic models,
allowing researchers to statistically estimate how topic prevalence and content
vary as a function of external information. As a result, STM can capture positive and negative nuances within topics while allowing uncertainty-aware statistical evaluation based on variational posterior inference. However, because topics are represented as probabilistic distributions over word co-occurrences, STM cannot directly capture semantic relationships that depend on word order or contextual information. Consequently, such models may generate incoherent topics~\cite{lau2014interpretable}.

To address the challenge of producing semantically intuitive topic representations,
interpretation-oriented approaches such as BERTopic, which leverages document
embeddings and distributed representations, and TopicGPT, which extracts topics
in natural language using large language models, have been developed
~\cite{grootendorst2022bertopic,pham2024topicgpt}. However, because these methods are not primarily designed to explicitly separate polarity stance, positive and negative descriptions may coexist within the same topic.

\subsection{Limitations of Exisiting Evaluation Metrics for Topic Modeling}
A wide range of metrics have been used to automatically evaluate topic models, including statistical goodness-of-fit measures and topic coherence metrics. However, recent studies have pointed out that these automatic evaluation metrics do not necessarily align well with human interpretations, leading to renewed scrutiny of how interpretability should be evaluated.

In particular, topic model evaluation has traditionally relied on automatic metrics that assess interpretability based on word co-occurrence patterns within topics. While such metrics offer the advantage of quantitatively assessing lexical coherence, prior studies have raised concerns about the validity of automated coherence metrics in capturing human interpretations~\cite{lau2014interpretable}. In contrast, human evaluation of semantic validity and interpretability can achieve high reliability, it is difficult to automate, which limits its applicability to large-scale datasets or comparisons across many experimental conditions.
Moreover, in analyses examining relationships between topics and external variables, it is important that topics can be meaningfully linked to specific behaviors or attributes and that their polarity orientations are consistent. To the best of our knowledge, however, no prior studies have explicitly defined these properties as evaluation criteria for topic models.
Taken together, evaluating topic modeling approaches for analyses involving external variables requires not only reliable automation of interpretability assessment but also the incorporation of topic specificity and polarity stance consistency as explicit evaluation criteria.

\section{Objectives and Contributions}
This study first develops a topic modeling methodology and an automated evaluation framework for analyzing relationships with external variables, explicitly incorporating not only interpretability but also topic specificity and polarity stance consistency. This enables topics derived from text data to be interpreted in a more practically meaningful manner. Furthermore, as an applied demonstration, we examine leadership characteristics that may contribute to firm performance and employee morale using large-scale employee experience review data. Through this empirical analysis, we demonstrate the practical usefulness of the proposed approach for social science research, while also providing insights that are practically valuable for companies.

\section{Method}
\subsection{Proposed Topic Modeling Framework}
\subsubsection{Overview of the Proposed Topic Modeling Framework}
\begin{figure*}[t]
  \centering
  \includegraphics[
    width=\textwidth,
    trim=1mm 75mm 1mm 60mm,
    clip
  ]{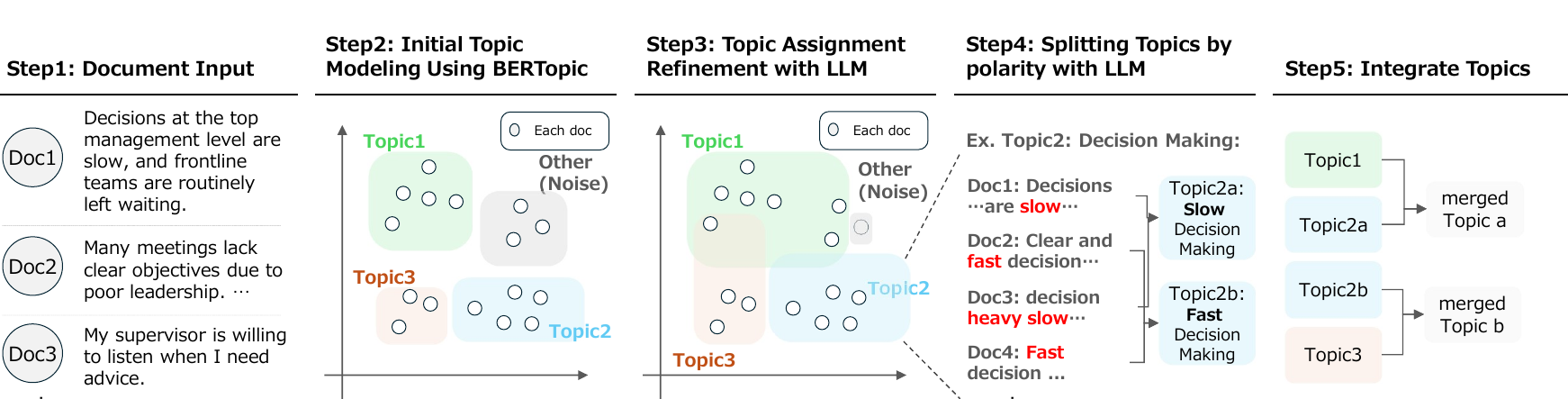}
  \caption{Overview of the proposed topic modeling framework.}
  \label{fig:proposed_topic_model_workflow}
\end{figure*}

In this study, we propose a topic modeling approach that simultaneously achieves interpretability as well as topic specificity and polarity stance consistency. Figure~\ref{fig:proposed_topic_model_workflow} provides an overview of the proposed framework. Specifically, the method proceeds as follows: (1) a collection of documents is provided as the initial data; (2) initial topics are generated using BERTopic; (3) An LLM is used to assign each document to zero or more initial topics, allowing documents to be associated with multiple topics or with none. (4) topics are split by polarity using LLMs; and (5) semantically related topics are integrated.

\subsubsection{Initial Topic Modeling Using BERTopic}
We apply document embeddings to the input documents to obtain continuous vector representations. To improve computational efficiency, principal component analysis (PCA) is applied to the embedded document vectors to reduce their dimensionality.

Using the dimension-reduced document embeddings as input, we generate an initial set of topics with BERTopic. For each initial topic, we provide an LLM with representative keywords and exemplar documents, and the model outputs a topic name and a topic description.
\subsubsection{Topic Assignment Refinement with LLM}
Next, the collection of generated topic names and topic descriptions is provided to an LLM together with all documents, and the model is used to reassign each document to topics.

BERTopic uses HDBSCAN for topic clustering, which is known to produce a substantial number of outlier documents (topic~$-1$)
\citep{grootendorst2022bertopic,McInnes2017,BERTopicOutlierReduction}.
We therefore use an LLM to reassign these documents to topics instead of discarding them as noise. Moreover, because a single document may be relevant to multiple topics, topic assignment is performed in a soft-clustering manner.

\subsubsection{Splitting Topics by polarity with LLM}
A randomly sampled set of documents is drawn from each topic and evaluated using an LLM to determine whether the topic contains documents with differing polarity stances; if at least one such document is identified, the topic is split into polarity-specific topics accordingly. Subsequently, documents associated with the pre-split topic are assigned, based on their content, to one of the polarity-specific topics or to neither of them.

\subsubsection{Topic Integration}
\label{topic_integration}
After splitting topics by polarity, we integrate semantically similar topics to obtain a more compact set of topic representations. During this process, to prevent the erroneous merging of topics with opposing polarity stances, we define topic similarity by jointly considering semantic proximity and polarity stance similarity.
First, for each topic, we construct a multi-view topic representation by taking a weighted average of the embeddings of the topic name, topic description, and the documents assigned to the topic, following prior work on LLM-guided multi-view cluster representations~\cite{pattnaik2024improving}. The semantic distance between topics $i$ and $j$, denoted as $d_{i,j}^{\text{meaning}}$, is defined as the cosine distance between their topic representation vectors.

Next, we define the polarity stance similarity between topics, denoted as $s_{i,j}^{\text{stance}}$, which quantifies the degree to which the polarity stances of topics $i$ and $j$ are aligned. This score is automatically evaluated using the G-Eval framework, which employs an LLM as an evaluator and takes the topic name and topic description as input~\cite{liu2023geval}. Larger values of $s_{i,j}^{\text{stance}}$ indicate greater similarity in polarity stance between the two topics.
Polarity stance information is intended solely as secondary information for determining whether topics should be integrated. It is therefore important to avoid cases in which topics with differing polarity stances are nevertheless considered similar solely because they are semantically related. To this end, we introduce a semantic distance threshold $\tau^{\text{meaning}}$ and incorporate stance information into the distance measure only for topic pairs that are sufficiently close in semantic space.
We define the Heaviside function $H_{i,j}$ as
\[
H_{i,j} =
\begin{cases}
1 & \text{if } d_{i,j}^{\text{meaning}} \le \tau^{\text{meaning}}, \\
0 & \text{otherwise}.
\end{cases}
\]
Using this function, the overall distance between topics $i$ and $j$ is defined as
\[
d_{i,j} = d_{i,j}^{\text{meaning}} - \left(1 - s_{i,j}^{\text{stance}}\right) H_{i,j}.
\]
As a result, topic pairs are ordered by distance, from semantically close and stance-aligned pairs, through semantically close but stance-misaligned pairs, to semantically distant pairs.

All prompts used in this study are provided in Appendix~\ref{prompts}.

\subsection{Proposed Evaluation Metrics}
In this study, we introduce additional evaluation metrics to complement conventional measures used in topic model evaluation, such as coherence and topic diversity metrics based on word co-occurrence and frequency statistics. Specifically, we incorporate metrics for topic label alignment and semantic–based topic diversity. More importantly, we introduce evaluation metrics that focus on topic specificity and polarity stance consistency, which constitute the core criteria of the proposed evaluation framework.
\subsubsection{Topic Label Alignment}
\label{topic_label_alignment}
Topic Label Alignment is a metric that evaluates the semantic alignment between the content of a topic label and the content of the document collection assigned to that topic. As discussed above, coherence metrics based on lexical consistency have been shown to be insufficient for fully capturing topic interpretability~\cite{lau2014interpretable}. Doogan et al.~\cite{doogan-2021-topic} further demonstrate that such metrics do not necessarily align with human evaluations, and argue that interpretability should instead be evaluated in terms of how descriptively a "Topic Word-set" represents its corresponding "Topic Document-collection". In this study, we operationalize this notion of alignment between topic-representative information and topic documents by automatically evaluating, using an LLM, the semantic alignment between a topic name, a topic description, and the documents assigned to that topic.
\subsubsection{Semantic-based Topic Diversity}
\label{semantic_based_topic_diversity}
We define Semantic-based Topic Diversity as a metric for evaluating the overall semantic diversity of a set of topics. Specifically, for all pairs of topics, an LLM is used to automatically assess their semantic similarity based on the topic names and topic descriptions. Topic pairs with similarity scores of 9 or higher on a 10-point scale are regarded as semantically equivalent. Based on these semantically equivalent topic pairs, topics are grouped into clusters of semantically similar topics, while topics with no identified similarities form singleton clusters. Finally, the Semantic-Based Topic Diversity score is computed as the ratio of the number of unique topic clusters to the total number of topics.
\subsubsection{Specificity}
\label{specificity}
We define Specificity as an evaluation metric that assesses whether a topic is expressed in a manner that enables readers to clearly imagine a specific situation or action. This metric is informed by prior discussions of \textit{imaginability} in Altarriba et al.~\cite{altarriba1999imaginability} and the notion of \textit{specific situations} proposed by Mischel~\cite{mischel1968personality}. Based on these perspectives, we define topic specificity along two aspects. First, we assess whether the topic representation allows readers to clearly envision the relevant actor and the specific state or behavioral change involved. For example, "a supervisor interrupting subordinates during weekly meetings" represents a high level of specificity, whereas "problems in organizational communication" is insufficiently specific. Second, we assess whether the topic describes a narrowly defined situation, condition, or phenomenon. For instance, "workflow delays caused by slow approval processes" exhibits higher specificity than a general description such as "declining work efficiency." Based on these two aspects, we use an LLM to automatically evaluate the extent to which a topic name and topic description are expressed with sufficient specificity.
\subsubsection{Polarity Stance Consistency}
\label{polarity_stance_consisitency}
We define Polarity Stance Consistency as an evaluation metric that assesses whether a single topic gives rise to mutually opposing polarity interpretations. Specifically, this metric reflects whether a topic name and description are expressed in an ambiguous way that can simultaneously refer to states with opposing polarity, such as presence versus absence, high versus low degrees, or strong versus weak levels. For example, expressions such as "rapid decision making" exhibit clear polarity stance consistency, whereas more ambiguous topic names such as "decision making" may simultaneously imply both positive and negative stance. In this study, polarity stance consistency is automatically evaluated using an LLM based on the topic name and topic description.

\section{Experiments}
\subsection{Input Data and Dataset Construction}
\label{input_data_and_dataset_construction}
This study used two types of data are required: employee experience reviews that contains corporate leader characteristics, and firm-level financial indicators used to measure organizational outcomes.

First, we analyzed employee experience review data posted on OpenWork, one of the largest corporate review platforms in Japan~\cite{openwork}. OpenWork contains freely written reviews submitted by current and former employees. The reviews provide rich accounts of organizational realities, including characteristics of leaders such as CEOs and managers. In addition to the textual content, each review includes five-point employee ratings (e.g., employee morale) as well as associated metadata, such as company identifiers, posting dates, and contributor attributes. Reviews posted between 2017 and 2024 were included in the analysis.
According to OpenWork's privacy policy, users consent to the provision of their data for third-party use after appropriate processing and for academic research purposes.The data used in this study were anonymized by OpenWork prior to being provided to the authors.
From each review text, we used an LLM to extract multiple passages referring to leaders, allowing multiple extractions per review. At the same time, the extracted passages were classified along two dimensions: (i) leader type, distinguishing between top executives (e.g., CEOs) and non-top leaders below the CEO level, and (ii) leader characteristics, categorizing each passage as referring to behaviors, attitudes, or abilities.

Second, we obtained corporate financial data through Japan’s
Electronic Disclosure for Investors’ NETwork (EDINET), a disclosure system provided by Japan’s Financial Services Agency~\cite{edinet}. In this study, we used return on assets (ROA) as an external outcome variable representing firm performance. ROA was calculated based on annual securities reports disclosed on EDINET, defined as net income divided by total assets.

Based on these steps, final set of input documents
was constructed from reviews posted for 1356 Japanese publicly listed firms. Table~\ref{tab:leader-extraction-count} reports the document counts of the extracted
leadership-related texts across the six groups. To ensure consistent temporal coverage between the review data and corporate financial data, the reviews were restricted to those posted for Japanese publicly listed firms for which financial statement data can be obtained for all fiscal years from 2017 to 2024. The LLM-based extraction was evaluated along category alignment for leader type and leader characteristic, and semantic consistency between the original and extracted texts, and was found to be largely valid (see Appendix).

\begin{table}[t]
\centering
\normalsize
\setlength{\tabcolsep}{5pt}
\renewcommand{\arraystretch}{0.95}
\caption{Counts of extracted leadership-related documents}
\label{tab:leader-extraction-count}
\begin{tabular}{@{}
    >{\raggedright\arraybackslash}p{0.28\columnwidth}
    >{\raggedright\arraybackslash}p{0.28\columnwidth}
    >{\centering\arraybackslash}p{0.24\columnwidth}
@{}}
\toprule
\textbf{Leader Type} &
\textbf{Leader Characteristic} &
\textbf{Document Count} \\
\midrule
\multirow{3}{*}{\textbf{Top}}
 & \textbf{Behavior} & 17,504 \\
 & \textbf{Attitude} & 8,737  \\
 & \textbf{Ability}  & 2,783  \\
\addlinespace[1.0mm]
\multirow{3}{*}{\textbf{Non-top}}
 & \textbf{Behavior} & 59,176 \\
 & \textbf{Attitude} & 12,833 \\
 & \textbf{Ability}  & 9,165  \\
\bottomrule
\end{tabular}
\end{table}
\subsection{Application of the Proposed Topic Modeling Method}
In this analysis, we applied the proposed topic modeling approach under the following settings. 
For Step~(1), \textit{Input Documents}, each of the six text
groups was treated as a separate set of input documents for the topic modeling (Table~\ref{tab:leader-extraction-count}).
For Step~(2), \textit{Initial Topic Modeling Using BERTopic}, we generated document embeddings for the leader-related text corpus described above using \textit{text-embedding-3-large} provided by OpenAI, resulting in 3,072-dimensional vector representations for each document. To improve computational efficiency, PCA was applied to the embedding vectors, reducing their dimensionality to 450 while preserving 90\% of the explained variance. Using the dimension-reduced document embeddings as input, we performed clustering with HDBSCAN to generate an initial set of topics. In this process, the minimum cluster size was set to 100, as each topic is required to contain a sufficient number of samples to allow for subsequent aggregation of topic frequencies at the firm-year level and for outcome regression analyses. All other hyperparameters were set to the default values of the bertopic library.\cite{grootendorst2022bertopic}
For Step~(3), \textit{Topic Assignment Refinement with LLM}, we extracted top-10 most probable words, and 30 representative documents for each topic using Maximal Marginal Relevance (MMR; \cite{Carbonell1998}) with $\lambda = 0.5$. These top-10 words and representative documents were provided as input to an LLM to generate topic names and a topic descriptions.
For Step~(4), \textit{Spliting Topics by Polarity with LLM}, we randomly sampled 50 documents from each topic and used an LLM to assess whether the topic contained descriptions with differing polarity stances, and topics exhibiting such polarity differences were then split accordingly.
For Step~(5), \textit{Topic Integration}, polarity stance similarity between topics was computed using the G-Eval framework, and topic integration was performed based on the resulting overall distance measure. The threshold for semantic distance $\tau^{\text{meaning}}$ was set to the bottom 1\% of the empirical distribution, and this threshold was used to determine whether stance-based adjustment should be applied when constructing the overall distance measure. Then, using the resulting topic–topic distance matrix, hierarchical clustering based on Ward's method was applied to merge topics. \cite{Ward1963,Johnson1967}. After topic integration, topic names and topic descriptions were regenerated using the same LLM-based naming procedure employed in Step~(4). When selecting representative documents for this renaming process, we set $\lambda = 0$ in MMR and sampled 30 documents to ensure that the resulting topic name and description comprehensively capture the full range of documents assigned to each integrated topic.
Throughout the entire pipeline, we used GPT-4.1-mini as the LLM.
\subsection{Results: Evaluation of Topic Models}
\subsubsection{Validation of the Proposed Evaluation Metrics}
To verify whether the evaluation metrics proposed in Section 4.2 provide judgments aligned with human assessments, we examined the degree of agreement between LLM-based automatic evaluations and human ratings. Specifically, for Topic Label Alignment, Specificity, Polarity Stance Consistency, and Semantic-based Topic Diversity, we computed the intraclass correlation coefficient, ICC(2,2) between LLM-based ratings and human ratings. Since GPT-4.1-mini was used in the topic modeling stage, we employed Gemini-2.5-Flash for LLM-based automatic evaluation in order to avoid potential bias and ensure fairness. Both the LLM and human annotators were provided with identical evaluation criteria, and each metric was assessed on 50 test cases. The resulting agreement scores are reported in Table~\ref{tab:icc_results}. The results indicate that Polarity Stance Consistency, Topic Label Alignment, and semantic-based Topic Diversity exhibit high levels of agreement between LLM and human evaluations, suggesting that these metrics are sufficiently reliable. Although Specificity shows relatively lower agreement compared to the other three metrics, its reliability remains within an acceptable range for practical use.
\begin{table}[t]
    \centering
    \caption{Inter-rater reliability between LLM and human evaluators}
    \begin{tabular}{lll}
        \toprule
        Metric & ICC(2,2) & 95\% CI \\
        \midrule
        Polarity Stance Consistency & 0.902 & [0.83, 0.94] \\
        Topic Label Alignment      & 0.895 & [0.82, 0.94] \\
        Diversity                  & 0.824 & [0.68, 0.90] \\
        Specificity                & 0.729 & [0.51, 0.85] \\
        \bottomrule
    \end{tabular}
    \label{tab:icc_results}

    \vspace{1mm}
    \begin{minipage}{\linewidth}
    \footnotesize
    \raggedright
    \textit{Notes:}Inter-rater reliability was assessed using ICC(2,2). Values in brackets indicate 95\% confidence intervals.
    \end{minipage}
\end{table}
\subsubsection{benchmarking proposed topic model}
In this section, we conducted a benchmark evaluation of the proposed method.
As evaluation metrics, we employed Coherence (NPMI) and Bag-of-Words–based Topic Diversity, which have been conventionally used in topic modeling research, together with the previously introduced metrics: Topic Label Alignment, semantic-based topic diversity, specificity, and polarity stance consistency. Finally, we also assessed the extent to which the proposed method improves the explanatory power with respect to external outcome variables.
\subsubsection{Competitors}
We compared the effectiveness of our method with the following baselines. As a bag-of-words–based method, we used Non-negative Matrix Factorization (NMF). As an embedding-based method, we used BERTopic, corresponding to the pipeline up to Step~(2) in Figure~\ref{fig:proposed_topic_model_workflow}. In addition, we included an LLM-based method that applies topic relabeling, corresponding to the pipeline up to Step~(3) in Figure~\ref{fig:proposed_topic_model_workflow}. For NMF, we used the default parameter settings. For the other methods, we adopted the same experimental settings as described above. 
\subsubsection{Evaluation metric settings}
For topic coherence, we adopted Normalized Pointwise Mutual Information (NPMI). For both NPMI and topic diversity, the topic representations were constructed using the top 10 words for each topic. For LLM-based automatic evaluation, we used Gemini-2.5-flash.
\subsubsection{Topic coherence evaluation}
Across both metrics, the proposed method generally achieved higher Topic Label Alignment and Coherence than the benchmark models (Table~\ref{tab:metrics-topic quality-npmi}). These results suggest that the generated topics are not only lexically coherent, but that the corresponding topic names and descriptions also descriptively represent the associated document collections.
\begin{table}[t]
\centering
\scriptsize
\setlength{\tabcolsep}{3pt}
\renewcommand{\arraystretch}{0.95}
\caption{Topic Coherence Evaluation Results}
\label{tab:metrics-topic quality-npmi}

\begin{tabular}{>{\raggedright\arraybackslash}p{0.05\textwidth}
                >{\raggedright\arraybackslash}p{0.05\textwidth}
                >{\raggedright\arraybackslash}p{0.07\textwidth}
                *{3}{>{\centering\arraybackslash}p{0.03\textwidth}}
                *{3}{>{\centering\arraybackslash}p{0.03\textwidth}}}
\toprule
\multirow{2}{*}{\textbf{Type.}} &
\multirow{2}{*}{\textbf{Char.}} &
\multirow{2}{*}{\textbf{Model}} &
\multicolumn{3}{c}{\textbf{Topic Label Alignment}} &
\multicolumn{3}{c}{\textbf{NPMI}} \\
\cmidrule(lr){4-6} \cmidrule(lr){7-9}
& & & \textbf{Tmin} & \textbf{Tmid} & \textbf{Tmax}
      & \textbf{Tmin} & \textbf{Tmid} & \textbf{Tmax} \\
\midrule
\multirow{4}{*}{\textbf{Top}} & \multirow{4}{*}{\textbf{Behavior}} & NMF & \textbf{0.75} & \textbf{0.67} & 0.62 & -0.12 & -0.26 & -0.29 \\
& & BERTopic & 0.50 & 0.53 & 0.54 & -0.10 & -0.15 & -0.17 \\
& & Relabel & 0.62 & \textbf{0.67} & – & \textbf{-0.05} & \textbf{-0.06} & – \\
& & Split(Proposed) & 0.65 & 0.66 & \textbf{0.74} & -0.06 & \textbf{-0.06} & \textbf{-0.06} \\
\addlinespace[1.5mm]
\multirow{4}{*}{\textbf{Top}} & \multirow{4}{*}{\textbf{Attitude}} & NMF & 0.45 & \textbf{0.66} & 0.66 & -0.12 & -0.23 & -0.31 \\
& & BERTopic & \textbf{0.69} & 0.60 & 0.52 & -0.14 & -0.19 & -0.18 \\
& & Relabel & 0.56 & 0.63 & – & \textbf{-0.08} & \textbf{-0.08} & – \\
& & Split(Proposed) & 0.53 & 0.62 & \textbf{0.70} & \textbf{-0.08} & \textbf{-0.08} & \textbf{-0.08} \\
\addlinespace[1.5mm]
\multirow{4}{*}{\textbf{Top}} & \multirow{4}{*}{\textbf{Ability}} & NMF & 0.65 & \textbf{0.83} & 0.62 & -0.15 & -0.29 & -0.35 \\
& & BERTopic & \textbf{0.85} & 0.58 & 0.53 & -0.21 & -0.16 & -0.21 \\
& & Relabel & 0.52 & 0.72 & – & \textbf{-0.11} & \textbf{-0.10} & – \\
& & Split(Proposed) & 0.69 & 0.64 & \textbf{0.76} & -0.17 & -0.11 & \textbf{-0.12} \\
\addlinespace[1.5mm]
\multirow{4}{*}{\textbf{Non-top}} & \multirow{4}{*}{\textbf{Behavior}} & NMF & 0.48 & 0.55 & 0.56 & -0.08 & -0.20 & -0.25 \\
& & BERTopic & 0.53 & 0.51 & 0.54 & -0.04 & -0.09 & -0.13 \\
& & Relabel & 0.53 & – & – & \textbf{-0.03} & – & – \\
& & Split(Proposed) & \textbf{0.54} & \textbf{0.60} & \textbf{0.60} & -0.04 & \textbf{-0.01} & \textbf{-0.01} \\
\addlinespace[1.5mm]
\multirow{4}{*}{\textbf{Non-top}} & \multirow{4}{*}{\textbf{Attitude}} & NMF & 0.51 & 0.63 & 0.58 & -0.04 & -0.21 & -0.26 \\
& & BERTopic & \textbf{0.65} & \textbf{0.66} & 0.62 & -0.09 & -0.12 & -0.14 \\
& & Relabel & 0.55 & – & – & \textbf{-0.07} & – & – \\
& & Split(Proposed) & 0.64 & \textbf{0.66} & \textbf{0.65} & -0.09 & \textbf{-0.06} & \textbf{-0.06} \\
\addlinespace[1.5mm]
\multirow{4}{*}{\textbf{Non-top}} & \multirow{4}{*}{\textbf{Ability}} & NMF & 0.52 & 0.60 & 0.60 & -0.10 & -0.21 & -0.28 \\
& & BERTopic & 0.44 & 0.57 & 0.58 & -0.08 & -0.12 & -0.18 \\
& & Relabel & 0.54 & \textbf{0.64} & – &\textbf{ -0.05} & \textbf{-0.04} & – \\
& & Split(Proposed) & \textbf{0.59} & 0.61 & \textbf{0.74} & -0.07 & \textbf{-0.04} & \textbf{-0.05} \\
\addlinespace[1.5mm]
\bottomrule
\end{tabular}
{\footnotesize
\begin{minipage}{\linewidth}
\raggedright
\textit{Common notes for Tables~3--6:}
 This table reports results by model and topic granularity. Type. denotes leader type. Char. denotes leader characteristic. The number of topics prior to integration was divided into five equally sized bins. The first, third, and fifth bins were selected as representative topic counts and correspond to Tmin, Tmid, and Tmax, respectively, with Tmax representing the original (non-integrated) number of topics. Scores are reported for Tmin, Tmid, and Tmax. "–" indicates that the corresponding value is not available. The largest values in each metric and settings are \textbf{bolded}.
 \end{minipage}
}
\end{table}
\subsubsection{Topic diversity evaluation}
While our method underperformed NMF and BERTopic in terms of Bag-of-Words–based Topic Diversity, it achieved comparable performance in Semantic-based Topic Diversity (Table~\ref{tab:metrics-topic diversity-topic diversity based on top words}). This suggests that our method is capable of capturing subtle semantic distinctions between topics, even when they share similar high-frequency words.
\subsubsection{Specificity and Polarity Stance Consistency evaluation}
The proposed method consistently achieved higher Specificity than NMF and BERTopic, and slightly higher Specificity than Relabeling with LLM (Table~\ref{tab:metrics-topic specificity-topic stance}). Moreover, the proposed method consistently attained the highest Polarity Stance Consistency across all comparison models.
\begin{table}[t]
\centering
\scriptsize
\setlength{\tabcolsep}{3pt}
\renewcommand{\arraystretch}{0.95}
\caption{Topic Diversity Evaluation Results}
\label{tab:metrics-topic diversity-topic diversity based on top words}

\begin{tabular}{>{\raggedright\arraybackslash}p{0.05\textwidth}
                >{\raggedright\arraybackslash}p{0.05\textwidth}
                >{\raggedright\arraybackslash}p{0.07\textwidth}
                *{3}{>{\centering\arraybackslash}p{0.03\textwidth}}
                *{3}{>{\centering\arraybackslash}p{0.03\textwidth}}}
\toprule
\multirow{2}{*}{\textbf{Type.}} &
\multirow{2}{*}{\textbf{Char.}} &
\multirow{2}{*}{\textbf{Model}} &
\multicolumn{3}{c}{\textbf{Semantic-based}} &
\multicolumn{3}{c}{\textbf{Bag-of-Words-based}} \\
\cmidrule(lr){4-6} \cmidrule(lr){7-9}
& & & \textbf{Tmin} & \textbf{Tmid} & \textbf{Tmax}
      & \textbf{Tmin} & \textbf{Tmid} & \textbf{Tmax} \\
\midrule
\multirow{4}{*}{\textbf{Top}} & \multirow{4}{*}{\textbf{Behavior}} & NMF & 0.27 & 0.18 & 0.29 & 0.45 & 0.49 & 0.48 \\
& & BERTopic & \textbf{0.50} & \textbf{0.66} & \textbf{0.54} & \textbf{0.54} & \textbf{0.71} & \textbf{0.73} \\
& & Relabel & 0.45 & 0.42 & – & 0.43 & 0.28 & – \\
& & Split(Proposed) & 0.45 & 0.52 & 0.47 & 0.44 & 0.29 & 0.25 \\
\addlinespace[1.5mm]
\multirow{4}{*}{\textbf{Top}} & \multirow{4}{*}{\textbf{Attitude}} & NMF & \textbf{1.00} & 0.50 & 0.41 & \textbf{0.68} & 0.55 & 0.51 \\
& & BERTopic & 0.50 & 0.53 & 0.58 & 0.38 & \textbf{0.65} & \textbf{0.68} \\
& & Relabel & \textbf{1.00} & 0.88 & – & 0.48 & 0.36 & – \\
& & Split(Proposed) & 0.80 & \textbf{0.94} & \textbf{0.78} & 0.50 & 0.34 & 0.28 \\
\addlinespace[1.5mm]
\multirow{4}{*}{\textbf{Top}} & \multirow{4}{*}{\textbf{Ability}} & NMF & \textbf{1.00} & 0.20 & \textbf{0.62} & 0.80 & \textbf{0.64} & 0.60 \\
& & BERTopic & – & 0.75 & 0.57 & \textbf{1.00} & 0.60 & \textbf{0.63} \\
& & Relabel & \textbf{1.00} & 0.75 & – & 0.80 & 0.45 & – \\
& & Split(Proposed) & \textbf{1.00} & \textbf{0.80} & \textbf{0.62} & 0.60 & 0.38 & 0.31 \\
\addlinespace[1.5mm]
\multirow{4}{*}{\textbf{Non-top}} & \multirow{4}{*}{\textbf{Behavior}} & NMF & 0.77 & 0.66 & 0.46 & 0.50 & 0.44 & 0.37 \\
& & BERTopic & 0.84 & \textbf{0.75} & 0.61 & \textbf{0.69} & \textbf{0.65} & \textbf{0.61} \\
& & Relabel & 0.87 & – & – & 0.31 & – & – \\
& & Split(Proposed) & \textbf{0.90} & 0.70 & \textbf{0.66} & 0.30 & 0.21 & 0.17 \\
\addlinespace[1.5mm]
\multirow{4}{*}{\textbf{Non-top}} & \multirow{4}{*}{\textbf{Attitude}} & NMF & 0.62 & 0.64 & 0.40 & 0.61 & 0.54 & 0.54 \\
& & BERTopic & 0.71 & 0.75 & 0.68 & \textbf{0.64} & \textbf{0.64} & \textbf{0.68} \\
& & Relabel & \textbf{0.88} & – & – & 0.42 & – & – \\
& & Split(Proposed) & \textbf{0.88} & \textbf{0.88} & \textbf{0.83} & 0.47 & 0.33 & 0.27 \\
\addlinespace[1.5mm]
\multirow{4}{*}{\textbf{Non-top}} & \multirow{4}{*}{\textbf{Ability}} & NMF & \textbf{0.67} & 0.37 & 0.25 & 0.58 & 0.51 & 0.53 \\
& & BERTopic & 0.60 & \textbf{0.56} & \textbf{0.65} & \textbf{0.64} & \textbf{0.64} & \textbf{0.69
}\\
& & Relabel & \textbf{0.67} & 0.39 & – & 0.47 & 0.31 & – \\
& & Split(Proposed) & 0.50 & 0.32 & 0.41 & 0.42 & 0.28 & 0.23 \\
\addlinespace[1.5mm]
\bottomrule
\end{tabular}
\end{table}
\begin{table}[t]
\centering
\scriptsize
\setlength{\tabcolsep}{3pt}
\renewcommand{\arraystretch}{0.95}
\caption{Specificity and Polarity Stance Consistency Evaluation Results }
\label{tab:metrics-topic specificity-topic stance}

\begin{tabular}{>{\raggedright\arraybackslash}p{0.05\textwidth}
                >{\raggedright\arraybackslash}p{0.05\textwidth}
                >{\raggedright\arraybackslash}p{0.07\textwidth}
                *{3}{>{\centering\arraybackslash}p{0.03\textwidth}}
                *{3}{>{\centering\arraybackslash}p{0.03\textwidth}}}
\toprule
\multirow{2}{*}{\textbf{Type.}} &
\multirow{2}{*}{\textbf{Char.}} &
\multirow{2}{*}{\textbf{Model}} &
\multicolumn{3}{c}{\textbf{Specificity}} &
\multicolumn{3}{c}{\textbf{Polarity Stance Consistency}} \\
\cmidrule(lr){4-6} \cmidrule(lr){7-9}
& & & \textbf{Tmin} & \textbf{Tmid} & \textbf{Tmax}
      & \textbf{Tmin} & \textbf{Tmid} & \textbf{Tmax} \\
\midrule
\multirow{4}{*}{\textbf{Top}} & \multirow{4}{*}{\textbf{Behavior}} & NMF & 0.23 & 0.27 & 0.35 & 0.57 & 0.61 & 0.63 \\
& & BERTopic & \textbf{0.42} & 0.36 & 0.37 & 0.84 & 0.79 & 0.86 \\
& & Relabel & 0.37 & 0.35 & – & 0.80 & 0.82 & – \\
& & Split(Proposed) & 0.41 & \textbf{0.45} & \textbf{0.40} & \textbf{0.91} & \textbf{0.91} & \textbf{0.95} \\
\addlinespace[1.5mm]
\multirow{4}{*}{\textbf{Top}} & \multirow{4}{*}{\textbf{Attitude}} & NMF & 0.16 & 0.14 & 0.14 & 0.78 & 0.54 & 0.61 \\
& & BERTopic & \textbf{0.25} & 0.18 & \textbf{0.21} & 0.50 & 0.66 & 0.74 \\
& & Relabel & 0.16 & 0.23 & – & 0.46 & 0.86 & – \\
& & Split(Proposed) & 0.18 & \textbf{0.25} & 0.20 & \textbf{0.82} & \textbf{0.89} & \textbf{0.98} \\
\addlinespace[1.5mm]
\multirow{4}{*}{\textbf{Top}} & \multirow{4}{*}{\textbf{Ability}} & NMF & \textbf{0.10} & \textbf{0.10} & \textbf{0.11} & 0.30 & 0.44 & 0.46 \\
& & BERTopic & \textbf{0.10} & \textbf{0.10} & 0.09 & 0.40 & 0.75 & 0.81 \\
& & Relabel & \textbf{0.10} & \textbf{0.10} & – & 0.70 & \textbf{0.85} & – \\
& & Split(Proposed) & \textbf{0.10} & \textbf{0.10} & 0.09 & \textbf{1.00} & 0.60 & \textbf{1.00} \\
\addlinespace[1.5mm]
\multirow{4}{*}{\textbf{Non-top}} & \multirow{4}{*}{\textbf{Behavior}} & NMF & 0.33 & 0.35 & 0.37 & 0.47 & 0.47 & 0.48 \\
& & BERTopic & 0.34 & 0.41 & 0.40 & 0.58 & 0.70 & 0.69 \\
& & Relabel & 0.43 & – & – & 0.69 & – & – \\
& & Split(Proposed) & \textbf{0.47} & \textbf{0.52} & \textbf{0.54} & \textbf{0.82} & \textbf{0.84} & \textbf{0.91} \\
\addlinespace[1.5mm]
\multirow{4}{*}{\textbf{Non-top}} & \multirow{4}{*}{\textbf{Attitude}} & NMF & 0.17 & 0.16 & 0.16 & 0.74 & 0.54 & 0.65 \\
& & BERTopic & 0.11 & 0.19 & 0.22 & 0.41 & 0.54 & 0.66 \\
& & Relabel & \textbf{0.29} & – & – & \textbf{0.80} & – & – \\
& & Split(Proposed) & 0.14 & \textbf{0.21} & \textbf{0.26} & 0.74 & \textbf{0.63} & \textbf{0.94} \\
\addlinespace[1.5mm]
\multirow{4}{*}{\textbf{Non-top}} & \multirow{4}{*}{\textbf{Ability}} & NMF & 0.12 & \textbf{0.15} & 0.13 & 0.37 & 0.58 & 0.58 \\
& & BERTopic & 0.10 & 0.14 & \textbf{0.14} & 0.52 & 0.57 & 0.64 \\
& & Relabel & \textbf{0.15} & 0.11 & – & 0.52 & 0.66 & – \\
& & Split(Proposed) & 0.12 & \textbf{0.15} & 0.10 & \textbf{0.68} & \textbf{0.78} & \textbf{0.95} \\
\addlinespace[1.5mm]
\bottomrule
\end{tabular}
\end{table}

\subsection{Aggregation of Topic Frequencies}
To analyze the relationship between the extracted topics and outcome variables in subsequent sections, we aggregated topic frequencies at the firm-year level for the final set of topics. In this study, the topic frequency is defined, for each firm-year combination, as the proportion of posts associated with a given topic relative to the total number of posts. 
\subsection{Results: Benchmarking the Explanatory Power of Leadership Topics for Outcomes}
To assess whether the topics constructed by the proposed method are more effective for outcome-related analyses than those generated by conventional methods, we benchmarked their explanatory power for external outcomes in terms of external outcomes.
Specifically, we compared the explanatory power of topic frequencies across different topic modeling approaches for ROA and employee morale. For all topic modeling approaches, topic frequencies were aggregated using the procedure described above. We modeled ROA and employee morale according to Equation~\ref{calc_explanatory_power} and computed the explanatory power of each topic modeling approach. To control for firm size, we included the logarithm of the number of employees as a control variable. In addition, to account for the influence of year-specific and industry-specific macro factors, the dependent variables for ROA and employee morale were defined as deviations from their respective year-industry means. The model is given by
\begin{equation}
Y_{i,t} = \sum_k \beta_k f_{i,t,k} + \alpha \log(\text{Employees}_{i,t}) + \varepsilon_{i,t},
\label{calc_explanatory_power}
\end{equation}
where $Y_{i,t}$ denoted the outcome variable for firm $i$ in year $t$, defined as the deviation of ROA or employee morale from the corresponding year--industry average, and $f_{i,t,k}$ represented the frequency of posts assigned to topic $k$. The regression model was estimated separately for each topic using subsamples of firms that meet minimum thresholds for the annual number of 10 posts, resulting in a sample of 373 firms. The model was estimated using Elastic Net regression ~\cite{ZouHastie2005}. To quantify explanatory power, we compared a full model that included all topic frequencies and control variables with a baseline model containing only the control variables, and computed the incremental contribution of topic frequencies. This incremental explanatory power was measured using the partial coefficient of determination (partial $R^2$).
\begin{table}[t]
\centering
\scriptsize
\setlength{\tabcolsep}{3pt}
\renewcommand{\arraystretch}{0.95}
\caption{Explanation Power for Outcomes Measures}
\label{tab:explanation rate-elasticnet-thr10}

\begin{tabular}{>{\raggedright\arraybackslash}p{0.05\textwidth}
                >{\raggedright\arraybackslash}p{0.05\textwidth}
                >{\raggedright\arraybackslash}p{0.07\textwidth}
                *{3}{>{\centering\arraybackslash}p{0.03\textwidth}}
                *{3}{>{\centering\arraybackslash}p{0.03\textwidth}}}
\toprule
\multirow{2}{*}{\textbf{Type.}} &
\multirow{2}{*}{\textbf{Char.}} &
\multirow{2}{*}{\textbf{Model}} &
\multicolumn{3}{c}{\textbf{ROA}} &
\multicolumn{3}{c}{\textbf{Employees Morale}} \\
\cmidrule(lr){4-6} \cmidrule(lr){7-9}
& & & \textbf{Tmin} & \textbf{Tmid} & \textbf{Tmax}
      & \textbf{Tmin} & \textbf{Tmid} & \textbf{Tmax} \\
\midrule
\multirow{4}{*}{\textbf{Top}} & \multirow{4}{*}{\textbf{Behavior}} & NMF & 0.004 & 0.006 & 0.005 & 0.011 & 0.013 & 0.016 \\
& & BERTopic & \textbf{0.009} & \textbf{0.017} & \textbf{0.018} & 0.000 & 0.021 & 0.022 \\
& & Relabel & 0.006 & 0.006 & – & 0.012 & 0.049 & – \\
& & Split(Proposed) & 0.005 & 0.008 & 0.009 & \textbf{0.026} & \textbf{0.053} & \textbf{0.048} \\
\addlinespace[1.5mm]
\multirow{4}{*}{\textbf{Top}} & \multirow{4}{*}{\textbf{Attitude}} & NMF & 0.007 & 0.008 & 0.008 & 0.011 & 0.015 & 0.014 \\
& & BERTopic & \textbf{0.011} & 0.007 & 0.011 & 0.000 & 0.009 & 0.015 \\
& & Relabel & 0.010 & 0.011 & – & 0.000 & 0.026 & – \\
& & Split(Proposed) & 0.010 & \textbf{0.015} & \textbf{0.015} & \textbf{0.018} & \textbf{0.033} & \textbf{0.032} \\
\addlinespace[1.5mm]
\multirow{4}{*}{\textbf{Top}} & \multirow{4}{*}{\textbf{Ability}} & NMF & \textbf{0.005} & 0.003 & 0.003 & 0.013 & 0.012 & 0.011 \\
& & BERTopic & 0.004 & 0.005 & 0.002 & 0.007 & 0.017 & 0.015 \\
& & Relabel & \textbf{0.005} & 0.005 & – & \textbf{0.022} & \textbf{0.020} & – \\
& & Split(Proposed) & \textbf{0.005} & \textbf{0.010} & \textbf{0.010} & 0.006 & 0.017 & \textbf{0.021} \\
\addlinespace[1.5mm]
\multirow{4}{*}{\textbf{Non-top}} & \multirow{4}{*}{\textbf{Behavior}} & NMF & 0.010 & 0.025 & 0.017 & 0.030 & 0.042 & 0.038 \\
& & BERTopic & 0.018 & \textbf{0.030} & \textbf{0.033} & 0.030 & 0.035 & 0.035 \\
& & Relabel & \textbf{0.024} & – & – & 0.040 & – & – \\
& & Split(Proposed) & 0.018 & 0.024 & 0.030 & \textbf{0.042} & \textbf{0.059} & \textbf{0.061} \\
\addlinespace[1.5mm]
\multirow{4}{*}{\textbf{Non-top}} & \multirow{4}{*}{\textbf{Attitude}} & NMF & 0.000 & \textbf{0.008} & 0.001 & 0.004 & 0.010 & 0.013 \\
& & BERTopic & \textbf{0.006} & 0.006 & \textbf{0.009} & 0.000 & 0.000 & 0.006 \\
& & Relabel & 0.001 & – & – & \textbf{0.010} & – & – \\
& & Split(Proposed) & 0.000 & 0.000 & 0.001 & 0.009 & \textbf{0.016} & \textbf{0.022} \\
\addlinespace[1.5mm]
\multirow{4}{*}{\textbf{Non-top}} & \multirow{4}{*}{\textbf{Ability}} & NMF & 0.005 & 0.008 & \textbf{0.009} & 0.002 & 0.010 & 0.012 \\
& & BERTopic & 0.000 & 0.001 & 0.000 & 0.002 & 0.000 & 0.002 \\
& & Relabel & 0.004 & \textbf{0.013} & – & \textbf{0.014} & \textbf{0.014} & – \\
& & Split(Proposed) & \textbf{0.007} & 0.007 & 0.000 & 0.011 & 0.012 & \textbf{0.016} \\
\addlinespace[1.5mm]
\bottomrule
\end{tabular}
\end{table}

Table~\ref{tab:explanation rate-elasticnet-thr10} shows that the proposed method achieved consistently higher explanatory power for employee morale compared with the baseline models. By contrast, for ROA, the proposed method didn't exhibit a consistent advantage across datasets. This may be because leadership-related topics exhibit weaker contemporaneous associations with ROA than with employee morale. In particular, changes in leadership-related perceptions may not be immediately reflected in financial indicators and can affect firm performance with a time lag. As a result, the corresponding effect sizes for ROA may be relatively small, which can make differences across topic modeling methods more difficult to detect.

\subsection{Results: Relationship Between Topics and External Outcomes}
To examine the relationship between leadership characteristics and firm performance as well as employee morale, we estimated regression models relating topic frequencies to ROA and employee morale. As discussed, analyses of explanatory power indicate that, in this dataset, models without topic integration generally achieved higher explanatory performance for both ROA and employee morale. Accordingly, the outcome analyses in this section were conducted using topic frequencies without applying topic integration. To identify which types of leadership characteristics are associated with higher ROA and employee morale, we conducted analyses relationship between each topic frequency and outcomes. for firm $i$, year $t$, and topic $k$, we estimated the following regression model:
\begin{equation}
Y_{i,t} = \beta_k f_{i,t,k} + \alpha \log(\text{Employees}_{i,t}) + \varepsilon_{i,t},
\end{equation}
where $Y_{i,t}$ denoted the outcome variable for firm $i$ in year $t$, defined as the deviation of ROA or employee morale from the corresponding year-industry average.
The variable $f_{i,t,k}$ represents the frequency of posts associated with topic $k$ for firm $i$ in year $t$, and $\log(\text{Employees}_{i,t})$ denotes the logarithm of the number of employees. The coefficient $\beta_k$ captures the association between topic frequency and the outcome, $\alpha$ represents the effect of firm size, and $\varepsilon_{i,t}$ is the error term.
The regression model was estimated separately for each topic using subsamples of firms that meet minimum thresholds for the annual number of posts (5, 10, and 15 posts per firm–year), resulting in sample sizes of 628, 373, and 258 firms, respectively. Table~\ref{tab:relationship_of_leadership_topic_and_outcomes} reports only topics that are robust, defined as being statistically significant at the 5\% level across at least two threshold specifications, and based on at least 100 posts assigned to the topic. For all but one topic—namely, "lack of managerial capability" (10 out of 11 topics in Table~\ref{tab:relationship_of_leadership_topic_and_outcomes}), the results are broadly consistent with prior leadership research reporting associations between leadership characteristics and firm performance or employee psychological outcomes (e.g., see \citealp{DeRue2011,Judge2004,Montano2017}), while providing more specific behavioral interpretations.

\begin{table*}[t]
\centering
\footnotesize
\setlength{\tabcolsep}{2pt}
\renewcommand{\arraystretch}{0.92}
\caption{Relationship between leadership topics and firm outcomes}
\label{tab:relationship_of_leadership_topic_and_outcomes}
\begin{adjustbox}{max width=\textwidth}
\begin{tabular}{>{\raggedright\arraybackslash}p{0.08\textwidth}>{\raggedright\arraybackslash}p{0.10\textwidth}>{\raggedright\arraybackslash}p{0.21\textwidth}>{\raggedright\arraybackslash}p{0.42\textwidth}>{\centering\arraybackslash}p{0.12\textwidth}>{\centering\arraybackslash}p{0.12\textwidth}>{\centering\arraybackslash}p{0.06\textwidth}>{\raggedright\arraybackslash}p{0.10\textwidth}}
\toprule
Leader Type & Characteristic & Topic Name & Topic Description & ROA & Employees Morale & N & Sig. thr. (5\%) \\
\midrule
Non-top & Ability & Lack of managerial capability
& Content indicating that managers are perceived as lacking managerial competence or operational skills 
& 68.21 (18.12)*** & 2.97 (0.95)*** & 249 & 5, 10, 15 \\
\addlinespace[0.3em]
\hdashline
\addlinespace[0.3em]

Non-top & Attitude & Lack of motivation and engagement$^a$
& Content related to managers’ lack of motivation or enthusiasm, or their passive attitudes toward work
& -31.81 (11.70)*** & -1.62 (0.62)*** & 545 & 10, 15 \\
\addlinespace[0.3em]
\hdashline
\addlinespace[0.3em]

Non-top & Attitude & Proactive mindset and behavior
& Content describing managers demonstrating high motivation, sense of responsibility, and growth orientation, and proactively engaging in tasks and employee development
& 31.95 (15.17)** & 2.67 (0.80)*** & 325 & 5, 10, 15 \\
\addlinespace[0.3em]
\hdashline
\addlinespace[0.3em]

Non-top & Behavior & Proactive project execution and task management$^a$ 
& Reviews describing leaders actively driving projects forward while managing and coordinating tasks 
& 32.39 (12.65)** & 1.72 (0.67)*** & 581 & 5, 10, 15 \\
\addlinespace[0.3em]
\hdashline
\addlinespace[0.3em]

Non-top & Behavior & Strict evaluation and disciplinary guidance 
& Attitudes in which branch managers or supervisors emphasize performance and discipline through strict guidance or pressure 
& -21.65 (11.28)* & -1.29 (0.60)** & 329 & 5, 15 \\
\addlinespace[0.3em]
\hdashline
\addlinespace[0.3em]

Non-top & Behavior & Proactive career support
& Reviews describing supervisors actively supporting subordinates’ career development through consultations and meetings to help achieve growth and aspirations
& 51.75 (11.95)*** & 3.20 (0.63)*** & 639 & 5, 10, 15 \\
\addlinespace[0.3em]
\hdashline
\addlinespace[0.3em]

Non-top & Behavior & Managerial dysfunction 
& Situations in which managers fail to adequately manage tasks or supervise subordinates, lacking presence or effective delegation of authority 
& -11.34 (6.43)* & -1.44 (0.34)*** & 2008 & 5, 15 \\
\addlinespace[0.3em]
\hdashline
\addlinespace[0.3em]

Non-top & Behavior & Early development of young leaders 
& Reviews describing environments where employees are given leadership responsibilities early and accumulate leadership experience 
& 45.76 (15.48)*** & 2.80 (0.82)*** & 344 & 5, 10, 15 \\
\addlinespace[0.3em]
\hdashline
\addlinespace[0.3em]

Top & Ability & Strong leadership capability
& Positive evaluations of top executives’ superior leadership capabilities, including expertise, decision-making, and strategic planning
& 41.47 (8.64)*** & 3.02 (0.46)*** & 845 & 5, 10, 15 \\
\addlinespace[0.3em]
\hdashline
\addlinespace[0.3em]

Top & Ability & Advanced managerial capability and foresight of top executives
& Evaluations highlighting top executives’ strong leadership, expertise, foresight, and broad practical capabilities
& 27.88 (8.58)*** & 2.91 (0.45)*** & 930 & 5, 10, 15 \\
\addlinespace[0.3em]
\hdashline
\addlinespace[0.3em]

Top & Ability & High leadership capability
& Evaluations of top executives’ strong leadership and high capability
& 45.04 (9.27)*** & 3.40 (0.49)*** & 748 & 5, 10, 15 \\
\addlinespace[0.3em]
\hdashline
\addlinespace[0.3em]

Top & Attitude & Charisma of top executives
& A topic capturing the charismatic leadership style of top executives, characterized by personal appeal and strong character
& 22.92 (9.27)** & 2.52 (0.49)*** & 757 & 10, 15 \\
\addlinespace[0.3em]
\hdashline
\addlinespace[0.3em]

Top & Attitude & Responsibility avoidance and self-preservation orientation
& Attitudes in which top executives avoid responsibility and prioritize protecting their own position and compensation
& -70.37 (25.78)*** & -6.02 (1.37)*** & 106 & 5, 10 \\
\addlinespace[0.3em]
\hdashline
\addlinespace[0.3em]

Top & Behavior & Active dialogue and communication 
& Behavioral patterns in which the CEO directly engages in dialogue and actively communicates with employees 
& 33.70 (8.66)*** & 3.04 (0.46)*** & 1147 & 5, 10, 15 \\
\bottomrule
\end{tabular}
\end{adjustbox}
\vspace{1mm}
\begin{minipage}{\textwidth}
\footnotesize
\textit{Notes:} For each entry, the reported value is the estimated coefficient, with the standard error shown in parentheses.$^{*}$, $^{**}$, and $^{***}$ indicate statistical significance at the 10\%, 5\%, and 1\% levels, respectively. Reported coefficients correspond to the reviewer-count threshold of 10 posts per firm-year. $N$ denotes the number of posts assigned to each topic. The column "Sig. thr. (5\%)" reports the reviewer-count thresholds (5, 10, 15) at which the coefficient is statistically significant at the 5\% level. Topics are included only if they are statistically significant at the 5\% level for both outcomes in at least 2 different thresholds. Topic names may be manually renamed based on the topic descriptions for interpretability.
\end{minipage}
\end{table*}
For example, it is shown that executive extraversion is positively associated with firm performance~\cite{Judge2004}. Consistent with this finding, the positive associations identified in this study between the leader behavior topic "active dialogue and communication" and both firm performance (ROA) and employee morale not only support prior evidence but also provide complementary insights by illustrating how an abstract personality trait such as extraversion manifests as specific leadership behaviors within organizations that are linked to performance outcomes. In contrast, the topic "lack of managerial capability" yields results that are not necessarily consistent with prior studies (see, e.g., \citealp{BloomVanReenen2007,BloomSadunVanReenen2016MAT}). However, since the present analysis is based on observational data and is limited to examining correlations rather than estimating causal relationships, these findings can be interpreted as reflecting the possibility of reverse causality.

\section{Conclusion}
This study demonstrated that it is possible to construct topic representations that simultaneously satisfy interpretability, specificity, and polarity stance consistency, and that the resulting topics can explain external outcomes as well as or better than those generated by conventional methods. These findings indicate that designing topics with interpretability, specificity, and polarity stance consistency helps prevent the dilution of relationships between topics and outcomes, thereby enabling the derivation of more actionable insights.

In addition to the methodological contribution described above, this study further contributes by proposing evaluation metrics for specificity and polarity stance consistency, as well as by operationalizing automatic evaluation methods for Specificity, Polarity Stance Consistency, Topic Label Alignment, and Semantic-Based Topic Diversity.

From the perspective of leadership research that motivated this study, these findings lead to the following three points.
First, core findings from prior leadership research largely accumulated in Western cultural contexts were broadly replicated in Japanese firms. For example, leadership behaviors typically associated with transactional leadership, such as "proactive project execution and task management", were identified as beneficial.
Second, by adopting a data-driven analysis of employee experience review data, this study presents patterns of leadership without being constrained by any theoretical frameworks.
Third, whereas prior cross-firm studies—often centered on meta-analyses—have tended to treat leadership characteristics as abstract meta-categories, the proposed approach enables these characteristics to be decomposed into specific behaviors, attitudes, and abilities at the level of individual topics. 
Taken together, the proposed topic modeling method and evaluation framework are not limited to leadership research and are applicable to a wide range of domains that analyze relationships between topics and external outcomes.

\section{Limitations and Future Work}
First, as the analysis in this study is primarily based on Japanese data, the generalizability of the findings is limited, and it is unclear whether similar patterns hold in other cultural contexts. Future work should incorporate multilingual and multicultural review data from other countries to enable cross-cultural comparisons.

Second, methodological refinement remains possible in linking topic representations to external outcomes. While this study adopts deterministic topic assignment, future work could employ probabilistic frameworks, such as STM, to represent topic–document associations as continuous values and enable finer-grained analyses.

Third, the evaluation of robustness is limited. Due to constraints in computational cost and execution time, validation was limited to a small set of model configurations and datasets. Future research should conduct more comprehensive evaluations across a wider range of LLMs and experimental conditions.



%
%
\bibliographystyle{ACM-Reference-Format}
\bibliography{references}

@article{blei2003lda,
  title   = {Latent Dirichlet Allocation},
  author  = {Blei, David M. and Ng, Andrew Y. and Jordan, Michael I.},
  journal = {Journal of Machine Learning Research},
  volume  = {3},
  pages   = {993--1022},
  year    = {2003}
}

@article{roberts2014stm,
  title   = {Structural Topic Models for Open-Ended Survey Responses},
  author  = {Roberts, Margaret E. and Stewart, Brandon M. and Tingley, Dustin},
  journal = {American Journal of Political Science},
  volume  = {58},
  number  = {4},
  pages   = {1064--1082},
  year    = {2014},
  doi = {10.1111/ajps.12103}
}

@article{grootendorst2022bertopic,
  title   = {BERTopic: Neural Topic Modeling with a Class-Based TF-IDF Procedure},
  author  = {Grootendorst, Maarten},
  journal = {arXiv preprint arXiv:2203.05794},
  year    = {2022},
  howpublished = {\url{https://github.com/MaartenGr/BERTopic}},
}

@misc{BERTopicOutlierReduction,
  title        = {Outlier reduction},
  author       = {Grootendorst, Maarten},
  howpublished = {BERTopic Documentation},
  year         = {2022},
  url          = {https://maartengr.github.io/BERTopic/getting_started/outlier_reduction/outlier_reduction.html}
}

@article{McInnes2017,
  title   = {HDBSCAN: Hierarchical Density Based Clustering},
  author  = {McInnes, Leland and Healy, John and Astels, Steve},
  journal = {Journal of Open Source Software},
  volume  = {2},
  number  = {11},
  pages   = {205},
  year    = {2017},
  doi     = {10.21105/joss.00205}
}

@article{pham2024topicgpt,
  title   = {TopicGPT: A Prompt-based Topic Modeling Framework},
  author  = {Pham, Chau Minh and Hoyle, Alexander and Sun, Ming and Resnik, Phillip and Iyyer, Mohit},
  url= {https://arxiv.org/abs/2311.01449},
  year    = {2024}
}

@article{tonidandel2022leadership,
  title   = {Using structural topic modeling to gain insight into challenges faced by leaders},
  author  = {Tonidandel, Scott and Summerville, Karoline M. and Gentry, William A. and Young, Stephen F.},
  journal = {The Leadership Quarterly},
  year    = {2022},
  volume  = {33},
  number  = {5},
  pages   = {101576},
  doi     = {10.1016/j.leaqua.2021.101576}
}

@inproceedings{lau2014interpretable,
  title     = {Machine Reading Tea Leaves: Automatically Evaluating Topic Coherence and Topic Model Quality},
  author    = {Lau, Jey Han and Newman, David and Baldwin, Timothy},
  booktitle = {Proceedings of the 14th Conference of the European Chapter of the Association for Computational Linguistics},
  pages     = {530--539},
  year      = {2014}
}

@article{Ward1963,
  author  = {Ward, Joe H.},
  title   = {Hierarchical Grouping to Optimize an Objective Function},
  journal = {Journal of the American Statistical Association},
  volume  = {58},
  number  = {301},
  pages   = {236--244},
  year    = {1963}
}

@article{Johnson1967,
  author  = {Johnson, Stephen C.},
  title   = {Hierarchical Clustering Schemes},
  journal = {Psychometrika},
  volume  = {32},
  pages   = {241--254},
  year    = {1967}
}

@article{Harter2002,
  author  = {Harter, James K. and Schmidt, Frank L. and Hayes, Theodore L.},
  title   = {Business-Unit-Level Relationship between Employee Satisfaction, Employee Engagement, and Business Outcomes: A Meta-Analysis},
  journal = {Journal of Applied Psychology},
  volume  = {87},
  number  = {2},
  pages   = {268--279},
  year    = {2002},
  doi     = {10.1037/0021-9010.87.2.268}
}

@article{Judge2001,
  author  = {Judge, Timothy A. and Thoresen, Carl J. and Bono, Joyce E. and Patton, Gregory K.},
  title   = {The Job Satisfaction--Job Performance Relationship: A Qualitative and Quantitative Review},
  journal = {Psychological Bulletin},
  volume  = {127},
  number  = {3},
  pages   = {376--407},
  year    = {2001},
  doi     = {10.1037/0033-2909.127.3.376}
}

@article{Judge2004,
  author  = {Judge, Timothy A. and Piccolo, Ronald F.},
  title   = {Transformational and Transactional Leadership: A Meta-Analytic Test of Their Relative Validity},
  journal = {Journal of Applied Psychology},
  volume  = {89},
  number  = {5},
  pages   = {755--768},
  year    = {2004},
  doi     = {10.1037/0021-9010.89.5.755}
}

@article{Montano2017,
  author  = {Montano, Diego and Reeske, Anna and Franke, Franziska and H{\"u}ffmeier, Joachim},
  title   = {Leadership, Followers' Mental Health and Job Performance in Organizations: A Comprehensive Meta-Analysis from an Occupational Health Perspective},
  journal = {Journal of Organizational Behavior},
  volume  = {38},
  pages   = {327--350},
  year    = {2017},
  doi     = {10.1002/job.2124}
}

@article{House1997,
  author  = {House, Robert J. and Aditya, Ram N.},
  title   = {The Social Scientific Study of Leadership: Quo Vadis?},
  journal = {Journal of Management},
  volume  = {23},
  number  = {3},
  pages   = {409--473},
  year    = {1997}
}

@article{Solansky2017,
  author  = {Solansky, Stephanie and Gupta, Vipin and Wang, Jia},
  title   = {Ideal and Confucian Implicit Leadership Profiles in China},
  journal = {Leadership \& Organization Development Journal},
  volume  = {38},
  number  = {2},
  doi = {10.1016/j.leaqua.2021.101576},
  pages   = {164--177},
  year    = {2017}
}

@article{DeRue2011,
  author  = {DeRue, Scott D. and Nahrgang, Jennifer D. and Wellman, Ned and Humphrey, Stephen E.},
  title   = {Trait and Behavioral Theories of Leadership: An Integration and Meta-Analytic Test of Their Relative Validity},
  journal = {Personnel Psychology},
  volume  = {64},
  number  = {1},
  pages   = {7--52},
  year    = {2011},
  doi     = {10.1111/j.1744-6570.2010.01201.x}
}

@article{Avolio1999,
  author  = {Avolio, Bruce J. and Bass, Bernard M. and Jung, Dong I.},
  title   = {Re-examining the Components of Transformational and Transactional Leadership Using the Multifactor Leadership Questionnaire},
  journal = {Journal of Occupational and Organizational Psychology},
  volume  = {72},
  number  = {4},
  pages   = {441--462},
  year    = {1999},
  doi     = {10.1348/096317999166789}
}

@article{WarnerSoderholm2020,
  author  = {Warner-Soderholm, Gillian and Minelgaite, Inga and Littrell, Romie Frederick},
  title   = {From LBDQXII to LBDQ50: Preferred Leader Behavior Measurement Across Cultures},
  journal = {Journal of Management Development},
  volume  = {39},
  number  = {1},
  pages   = {68--81},
  year    = {2020},
  doi     = {10.1108/JMD-03-2019-0067}
}

@article{Misumi1985,
  author  = {Misumi, Jyuji and Peterson, Mark F.},
  title   = {The Performance--Maintenance (PM) Theory of Leadership: Review of a Japanese Research Program},
  journal = {Administrative Science Quarterly},
  volume  = {30},
  number  = {2},
  pages   = {198--223},
  year    = {1985},
  doi = {10.2307/2393105}
}

@article{GobelMiyamoto2023,
  author  = {Gobel, Matthias S. and Miyamoto, Yuri},
  title   = {Self- and Other-Orientation in High Rank: A Cultural Psychological Approach to Social Hierarchy},
  journal = {Personality and Social Psychology Review},
  pages   = {54--80},
  volume  = {28},
  number  = {1},
  year    = {2023},
  doi     = {10.1177/10888683231172252}
}

@article{Schimmelpfennig2025,
  author  = {Schimmelpfennig, Robin and Elb{\ae}k, Christian and Mitkidis, Panagiotis and Singh, Anisha and Roberson, Quinetta},
  title   = {The ``WEIRDEST'' Organizations in the World? Assessing the Lack of Sample Diversity in Organizational Research},
  journal = {Journal of Management},
  volume  = {51},
  number  = {6},
  pages   = {2460--2487},
  year    = {2025},
  doi     = {10.1177/01492063241305577}
}

@article{Dinh2014,
  author  = {Dinh, Jessica E. and Lord, Robert G. and Gardner, William L. and Meuser, Jeremy D. and Liden, Robert C. and Hu, Jinyu},
  title   = {Leadership Theory and Research in the New Millennium: Current Theoretical Trends and Changing Perspectives},
  journal = {The Leadership Quarterly},
  volume  = {25},
  number  = {1},
  pages   = {36--62},
  year    = {2014},
  doi     = {10.1016/j.leaqua.2013.11.005}
}

@article{Yukl2019,
  author  = {Yukl, Gary and Mahsud, Raza and Prussia, Gregory E. and Hassan, Shafiq},
  title   = {Effectiveness of Broad and Specific Leadership Behaviors},
  journal = {Personnel Review},
  volume  = {48},
  number  = {3},
  pages   = {774--783},
  year    = {2019},
  doi     = {10.1108/PR-03-2018-0100}
}

@inproceedings{doogan-2021-topic,
  title     = {Topic Model or Topic Twaddle? Re-evaluating Semantic Interpretability Measures},
  author    = {Doogan, Caitlin and Buntine, Wray},
  editor    = {Toutanova, Kristina and Rumshisky, Anna and Zettlemoyer, Luke and Hakkani-Tur, Dilek and Beltagy, Iz and Bethard, Steven and Cotterell, Ryan and Chakraborty, Tanmoy and Zhou, Yichao},
  booktitle = {Proceedings of the 2021 Conference of the North American Chapter of the Association for Computational Linguistics: Human Language Technologies},
  month     = jun,
  year      = {2021},
  address   = {Online},
  publisher = {Association for Computational Linguistics},
  url       = {https://aclanthology.org/2021.naacl-main.300/},
  doi       = {10.18653/v1/2021.naacl-main.300},
  pages     = {3824--3848}
}

@article{altarriba1999imaginability,
  title   = {Concreteness, context availability, and imageability ratings and word associations for abstract, concrete, and emotion words},
  author  = {Altarriba, Jeanette and Bauer, Laurie M. and Benvenuto, Claudia},
  journal = {Behavior Research Methods, Instruments, \& Computers},
  volume  = {31},
  number  = {4},
  pages   = {578--602},
  year    = {1999},
  doi     = {10.3758/BF03200738}
}

@book{mischel1968personality,
  title     = {Personality and Assessment},
  author    = {Mischel, Walter},
  year      = {1968},
  publisher = {Wiley},
  address   = {New York}
}

@article{BloomVanReenen2007,
  author  = {Bloom, Nicholas and Van Reenen, John},
  title   = {Measuring and Explaining Management Practices Across Firms and Countries},
  journal = {The Quarterly Journal of Economics},
  volume  = {122},
  number  = {4},
  pages   = {1351--1408},
  year    = {2007},
  doi     = {10.1162/qjec.2007.122.4.1351}
}

@techreport{BloomSadunVanReenen2016MAT,
  title       = {Management as a Technology?},
  author      = {Bloom, Nicholas and Sadun, Raffaella and Van Reenen, John},
  institution = {National Bureau of Economic Research},
  type        = {Working Paper},
  number      = {22327},
  year        = {2016},
  doi         = {10.3386/w22327},
  url         = {https://www.nber.org/papers/w22327}
}

@article{ZouHastie2005,
  author  = {Zou, Hui and Hastie, Trevor},
  title   = {Regularization and Variable Selection via the Elastic Net},
  journal = {Journal of the Royal Statistical Society: Series B (Statistical Methodology)},
  volume  = {67},
  number  = {2},
  pages   = {301--320},
  year    = {2005},
  doi     = {10.1111/j.1467-9868.2005.00503.x}
}

@article{liu2023geval,
  author  = {Liu, Yang and Iter, Dan and Xu, Yichong and Wang, Shuohang and Xu, Ruochen and Zhu, Chenguang},
  title   = {G-Eval: NLG Evaluation using GPT-4 with Better Human Alignment},
  journal = {arXiv preprint arXiv:2303.16634},
  year    = {2023}
}

@inproceedings{Carbonell1998,
  author    = {Carbonell, Jaime and Goldstein, Jade},
  title     = {The Use of MMR, diversity-based reranking for reordering documents and producing summaries},
  booktitle = {Proceedings of the 21st Annual International ACM SIGIR Conference on Research and Development in Information Retrieval},
  pages     = {335--336},
  year      = {1998},
  publisher = {ACM},
  doi       = {10.1145/290941.291025}
}

@misc{openwork,
  author       = {{OpenWork Inc.}},
  title        = {{OpenWork}: Japanese Corporate Review Platform},
  year         = {2025},
  howpublished = {\url{https://www.openwork.jp/}},
}

@misc{edinet,
  author       = {{Financial Services Agency of Japan}},
  title        = {{EDINET}: Electronic Disclosure for Investors' NETwork},
  year         = {2025},
  howpublished = {\url{https://disclosure2.edinet-fsa.go.jp/}},
}

@inproceedings{pattnaik2024improving,
  title     = {Improving Hierarchical Text Clustering with LLM-guided Multi-view Cluster Representation},
  author    = {Pattnaik, Anup and George, Cijo and Tripathi, Rishabh Kumar and Vutla, Sasanka and Vepa, Jithendra},
  booktitle = {Proceedings of the 2024 Conference on Empirical Methods in Natural Language Processing (Industry Track)},
  year      = {2024},
  url       = {https://aclanthology.org/2024.emnlp-industry.54}
}

\appendix
\section{Human Evaluation of Extraction Precision}
\FloatBarrier
\begin{table}
\centering
\normalsize
\setlength{\tabcolsep}{5pt}
\renewcommand{\arraystretch}{0.95}
\caption{Precision of leadership-related document extraction evaluated by human, by category}
\label{tab:leader-extraction-precision}
\begin{tabular}{@{}
    >{\raggedright\arraybackslash}p{0.40\columnwidth}
    >{\centering\arraybackslash}p{0.24\columnwidth}
    >{\centering\arraybackslash}p{0.24\columnwidth}
@{}}
\toprule
\multirow{2}{*}{\textbf{Category}} &
\multicolumn{2}{c}{\textbf{Precision}} \\
\cmidrule(lr){2-3}
 & \textbf{Pre/Post Meaning Consistency}
 & \textbf{Category Alignment} \\
\midrule
\textbf{Top}        & 1.00 & 0.88 \\
\textbf{Non-top}    & 0.90 & 1.00 \\
\textbf{Behavior}   & 0.90 & 0.96 \\
\textbf{Attitude}   & 0.86 & 0.92 \\
\textbf{Ability}    & 0.82 & 0.78 \\
\bottomrule
\end{tabular}
\vspace{1mm}
\begin{minipage}{\linewidth}
\footnotesize
\textit{Notes:} Precision values are based on human evaluation of LLM-extracted documents along two dimensions: (1) \textit{Pre/Post Semantic Consistency}, which measures semantic consistency between the original review text and the extracted passages before and after processing, and (2) \textit{Category Alignment}, which measures the alignment between the assigned categories and human judgments.

\end{minipage}
\end{table}

\FloatBarrier
\section{Libraries and Parameter Settings for the Proposed Topic Modeling}

This appendix lists the main Python libraries used in this study, along with their official URLs.

\begin{itemize}
  \item Incremental PCA was implemented using \textit{IncrementalPCA} from scikit-learn:
  \url{https://scikit-learn.org/stable/}.

  \item Topic integration employed hierarchical clustering via \textit{fcluster} (criterion: \textit{maxclust}) in \textit{SciPy}:
  \url{https://docs.scipy.org/doc/scipy/}.

  \item NMF was implemented using \textit{NMF} from scikit-learn:
  \url{https://scikit-learn.org/stable/}.

  \item Automated evaluation was conducted using \textit{deepeval}:
  \url{https://docs.confident-ai.com/}.

  \item Panel regressions were estimated using \textit{PanelOLS} from \textit{linearmodels.panel}:
  \url{https://github.com/bashtage/linearmodels}.

  \item Elastic net regularization was applied using \textit{ElasticNetCV} from scikit-learn:
  \url{https://scikit-learn.org/stable/}.
\end{itemize}

\FloatBarrier
\clearpage
\section{Prompts Used in this Study}
\label{prompts}
\newcommand{\promptcell}[1]{%
  \begin{minipage}[t]{\linewidth}
    \ttfamily\footnotesize\raggedright
    #1
  \end{minipage}%
}

\begin{table*}[tbp]
\centering
\caption{Prompt used for extracting leadership documents (see Section~\ref{input_data_and_dataset_construction})}
\label{tab:extract-leadership-document}
\small
\setlength{\tabcolsep}{10pt}
\renewcommand{\arraystretch}{1.15}
\begin{tabularx}{\textwidth}{@{} X X @{}}
\toprule
\textbf{English Prompt (Translation)} & \textbf{Japanese Prompt (Original)} \\
\midrule

\promptcell{%
\# Task\\
From the given text, extract instances of the specified extraction target at the minimum granularity, and classify them according to the provided classification guidelines.\\[0.8ex]
\# Requirements\\
- If multiple instances of the extraction target are present, extract all of them.\\
- If no instances of the extraction target are found, return an empty list.\\
- Avoid speculative or overly interpretive reasoning, and base the extraction strictly on information explicitly stated in the text.\\[0.8ex]
\# Metadata for the given text\\
\{input\_text\_metadata\}\\[0.8ex]
\# Given text\\
\{input\_text\}\\[0.8ex]
\# Extraction target\\
\{extraction\_target\}\\[0.8ex]
\# Supplementary definition of the extraction target\\
\{extraction\_target\_supplement\}\\[0.8ex]
\# Classification guidelines\\
\{classification\_guideline\}\\[0.8ex]
\# Output format\\
Output the results in JSON using the following schema.\\
\{output\_json\_schema\}%
}
&
\promptcell{%
\# タスク\\
与えられた文章から\#抽出対象を最小粒度で抽出し、\#分類仕様に従って分類してください。\\[0.8ex]
\# 要件\\
・抽出対象の記述が複数ある場合は、全て抽出してください。\\
・抽出対象の記述がない場合は、空リストを返してください。\\
・飛躍した解釈や過度な推測を避け、文章に明確に記載されている内容に基づいて抽出してください。\\[0.8ex]
\# 与えられた文章に関するメタ情報\\
\{input\_text\_metadata\}\\[0.8ex]
\# 与えられた文章\\
\{input\_text\}\\[0.8ex]
\# 抽出対象\\
\{extraction\_target\}\\[0.8ex]
\# 抽出対象の補足定義\\
\{extraction\_target\_supplement\}\\[0.8ex]
\# 分類仕様\\
\{classification\_guideline\}\\[0.8ex]
\# 出力形式\\
以下の形式でJSONを出力してください。\\
\{output\_json\_schema\}%
}
\\

\bottomrule
\end{tabularx}
\vspace{1mm}
\begin{minipage}{\textwidth}
\footnotesize
\textit{Notes:}\textit{\{extraction\_target\}} denotes leader-related attributes that are explicitly described in the employee experienced review text (not wishes/ideals) and attributable to an individual leader (not general employees, policies, or organizational features; passive statements are excluded). \textit{\{extraction\_target\_supplement\}} provides extraction constraints (no speculative inference; preserve meaning; split into minimal concise units). \textit{\{classification\_guideline\}} specifies labels for \textit{target\_leader\_layer} (\textit{top}/\textit{non\_top}/\textit{unknown}) and \textit{element\_type} (\textit{behavior}/\textit{attitude}/\textit{ability}/\textit{other}); use \textit{unknown}/\textit{other} when evidence is insufficient. \textit{\{input\_text\_metadata\}} includes the firm name and review category and notes that the reviewer is not necessarily a leader. The model outputs JSON following \textit{\{output\_json\_schema\}}, optionally setting flags such as \textit{implicit\_extraction}, \textit{change\_meaning}, and \textit{is\_past}.
\end{minipage}
\end{table*}

\begin{table*}[tbp]
\centering
\caption{Prompt used in Step~2 of Figure~\ref{fig:proposed_topic_model_workflow}}
\label{tab:topic-naming-prompt}
\small
\setlength{\tabcolsep}{10pt}
\renewcommand{\arraystretch}{1.15}
\begin{tabularx}{\textwidth}{@{} X X @{}}
\toprule
\textbf{English Prompt (Translation)} & \textbf{Japanese Prompt (Original)} \\
\midrule

\promptcell{%
\# Task\\
For a single topic generated by the topic model, determine an appropriate topic name by referring to the topic's top words and representative documents.\\[0.8ex]
\# Requirements\\
- In addition to the topic name (\texttt{topic\_name}), provide a short description of the topic (\texttt{topic\_short\_description}).\\
- Output the result in JSON format.\\[0.8ex]
\# Supplementary naming guidelines\\
- The topic name should be a noun phrase.\\
- The topic name should be concise; avoid redundant expressions such as ``A and B'' or ``A and B-related,'' and keep the number of words to a minimum.\\
- The topic name should comprehensively reflect the content of the representative documents.\\
- The topic name should be as specific as possible.\\
- The topic name should be consistent with the overall context inferred from the metadata of the topic modeling corpus.\\
- The topic description should consist of approximately one sentence and serve as a supplementary explanation of the topic name.\\[0.8ex]
\# Metadata of the topic modeling corpus\\
\{document\_metadata\}\\[0.8ex]
\# Top words of the topic\\
\{topic\_top\_words\}\\[0.8ex]
\# Representative documents of the topic\\
\{topic\_representative\_documents\}\\[0.8ex]
\# Output schema\\
\texttt{topic\_name}: string\\
\texttt{topic\_short\_description}: string\\[0.8ex]
\# Output format\\
\{%
\newline
\ \ "topic\_name": "Topic name",\\
\ \ "topic\_short\_description": "Short description of the topic"\\
\}%
}
&
\promptcell{%
\# タスク\\
トピックモデルによって作成された1つのトピックについて、\#トピックの上位単語および\#トピックの代表文章を参考に、適切なトピック名を決定してください。\\[0.8ex]
\# 要件\\
・トピック名（\texttt{topic\_name}）に加えて、トピック名についての短い説明（\texttt{topic\_short\_description}）を付与してください。\\
・JSON形式で出力してください。\\[0.8ex]
\# 命名規則の補足定義\\
・トピック名は名詞句としてください。\\
・トピック名は簡潔な表現とし、「AとB」「AおよびBに関する〜」のような冗長な表現は避け、単語数はできるだけ少なくしてください。\\
・トピック名は、\#トピックの代表文章の内容を網羅する表現としてください。\\
・トピック名は、可能な限り具体的な表現としてください。\\
・トピック名は、\#トピックモデリング対象の文章全体のメタ情報から読み取れる文脈やニュアンスに沿った表現としてください。\\
・トピック説明は1文程度とし、トピック名の補足説明となる内容としてください。\\[0.8ex]
\# トピックモデリング対象の文章全体のメタ情報\\
\{document\_metadata\}\\[0.8ex]
\# トピックの上位単語\\
\{topic\_top\_words\}\\[0.8ex]
\# トピックの代表文章\\
\{topic\_representative\_documents\}\\[0.8ex]
\# 出力型\\
\texttt{topic\_name}: str\\
\texttt{topic\_short\_description}: str\\[0.8ex]
\# 出力形式\\
\{%
\newline
\ \ "topic\_name": "トピック名",\\
\ \ "topic\_short\_description": "トピック名についての短い説明"\\
\}%
}
\\

\bottomrule
\end{tabularx}
\end{table*}

\begin{table*}[tbp]
\centering
\caption{Prompt used in Step~3 of Figure~\ref{fig:proposed_topic_model_workflow}}
\label{tab:topic-assignment-prompt}
\small
\setlength{\tabcolsep}{10pt}
\renewcommand{\arraystretch}{1.15}
\begin{tabularx}{\textwidth}{@{} X X @{}}
\toprule
\textbf{English Prompt (Translation)} & \textbf{Japanese Prompt (Original)} \\
\midrule

\promptcell{%
\# Task\\
Given the input text, select the topic or topics from the list of candidate topics to which the text corresponds.\\[0.8ex]
\# Requirements\\
- Judge whether the text corresponds to each topic by considering both the topic name and its description.\\
- If the text corresponds to multiple topics, select all applicable topics.\\
- For each selected topic, output the topic ID (\texttt{topic\_id}), topic name (\texttt{topic\_name}), and the reason for selection (\texttt{reason}).\\
- Output the results in JSON format.\\[0.8ex]
\# Supplementary guidelines for topic assignment\\
- Avoid judgments based on speculative or inferential reasoning.\\
- Select only topics that clearly apply to the text; apply a conservative judgment criterion.\\
- Make the judgment in accordance with the nuance inferred from the metadata of the input text.\\[0.8ex]
\# Notes on output\\
- If the text does not correspond to any topic, set the topic ID to \texttt{-1} and the topic name to \texttt{Other}.\\
- Except for \texttt{Other}, do not output topics that are not included in the candidate topic list.\\[0.8ex]
\# Metadata of the text\\
\{document\_metadata\}\\[0.8ex]
\# Candidate topics (legend: topic\_id, topic name (topic description))\\
\{topic\_definitions\}\\[0.8ex]
\# Input text\\
\{input\_text\}\\[0.8ex]
\# Output format\\
\{%
\newline
\ \ "topic\_list": [\\
\ \ \ \ \{\\
\ \ \ \ \ \ "topic\_id": int,\\
\ \ \ \ \ \ "topic\_name": str,\\
\ \ \ \ \ \ "reason": str\\
\ \ \ \ \},\\
\ \ \ \ \ldots\\
\ \ ]\\
\}%
}
&
\promptcell{%
\# タスク\\
与えられた文章が、候補となるトピックのうちどのトピックに該当するかを選出してください。\\[0.8ex]
\# 要件\\
・トピック名およびトピックの説明の両方を確認したうえで、文章がトピックに該当するかを判断してください。\\
・複数のトピックに該当する場合は、複数選択してください。\\
・選択したトピックについて、トピックID（\texttt{topic\_id}）、トピック名（\texttt{topic\_name}）、および選択理由（\texttt{reason}）を出力してください。\\
・JSON形式で出力してください。\\[0.8ex]
\# トピック判定における補足定義\\
・飛躍した推測による判断は避けてください。\\
・明確に該当すると判断できるトピックのみを選出してください（厳しめの判断基準としてください）。\\
・文章のメタ情報を踏まえたニュアンスに沿って判断してください。\\[0.8ex]
\# 出力に関する注意点\\
・どのトピックにも該当しない場合は、トピックIDは \texttt{-1}、トピック名は \texttt{その他} を選択してください。\\
・\texttt{その他} を除き、候補となるトピックに記載されていないトピックは出力しないでください。\\[0.8ex]
\# 文章のメタ情報\\
\{document\_metadata\}\\[0.8ex]
\# 候補となるトピック（凡例：トピックID，トピック名（トピック説明））\\
\{topic\_definitions\}\\[0.8ex]
\# 文章\\
\{input\_text\}\\[0.8ex]
\# 出力形式\\
\{%
\newline
\ \ "topic\_list": [\\
\ \ \ \ \{\\
\ \ \ \ \ \ "topic\_id": int,\\
\ \ \ \ \ \ "topic\_name": str,\\
\ \ \ \ \ \ "reason": str\\
\ \ \ \ \},\\
\ \ \ \ \ldots\\
\ \ ]\\
\}%
}
\\

\bottomrule
\end{tabularx}
\end{table*}

\begin{table*}[tbp]
\centering
\caption{Prompt used for splitting topics by polarity in Step~4 of Figure~\ref{fig:proposed_topic_model_workflow}}
\label{tab:opposing-stance-splitting-prompt}
\setlength{\tabcolsep}{10pt}
\renewcommand{\arraystretch}{1.15}
\begin{tabularx}{\textwidth}{@{} X X @{}}
\toprule
\textbf{English Prompt (Translation)} & \textbf{Japanese Prompt (Original)} \\
\midrule

\promptcell{%
\scriptsize
\# Task\\
For a single topic generated by the topic model, determine whether documents with opposing stances are mixed within the topic. If opposing stances are present, split the topic accordingly.\\[0.8ex]
\# Definition of terms\\
- ``Opposing stances'' refer to cases in which documents classified under the same topic convey conflicting meanings.\\
- Examples of opposing stances include contrasts such as ``present vs.\ absent,'' ``many vs.\ few,'' and ``strong vs.\ weak.''\\[0.8ex]
\# Requirements\\
- Output the judgment result indicating whether opposing stances are present (\texttt{contain\_opposing\_stance}).\\
- If topic splitting is required, output a list of child topics (\texttt{child\_topics}).\\
- Each element of \texttt{child\_topics} should be a dictionary containing the child topic name (\texttt{child\_topic\_name}), a short description (\texttt{child\_topic\_short\_description}), up to three example documents (\texttt{document\_examples}), and the reason for interpreting the stance as opposing (\texttt{opposing\_stance\_reason}).\\
- If splitting is not required, output an empty list for \texttt{child\_topics}.\\
- Output the results in JSON format.\\[0.8ex]
\# Supplementary guidelines for judgment\\
- Topic information is provided in the topic name, description, and the set of documents assigned to the topic.\\
- Judge opposing stances only when documents can be clearly interpreted as conveying conflicting stances.\\[0.8ex]
\# Supplementary guidelines for naming child topics\\
- Child topic names should be noun phrases.\\
- Child topic names should be concise; avoid redundant expressions such as ``A and B.''\\
- Child topic names should be as specific as possible.\\
- Child topic names should be interpretable on their own without reference to the parent topic.\\
- Child topic names should reflect the overall context inferred from the metadata of the topic modeling corpus.\\[0.8ex]
\# Topic name and description\\
Topic name: \{topic\_name\}\\
Topic description: \{topic\_short\_description\}\\[0.8ex]
\# Metadata of the topic modeling corpus\\
\{document\_metadata\}\\[0.8ex]
\# Documents assigned to the topic\\
\{topic\_documents\}\\[0.8ex]
\# Output format (if opposing stances are present)\\
\{%
\newline
\ \ "contain\_opposing\_stance": true,\\
\ \ "child\_topics": [\\
\ \ \ \ \{\\
\ \ \ \ \ \ "child\_topic\_name": "Child topic name 1",\\
\ \ \ \ \ \ "child\_topic\_short\_description": "Description of child topic 1",\\
\ \ \ \ \ \ "document\_examples": "Up to three example documents",\\
\ \ \ \ \ \ "opposing\_stance\_reason": "Reason for interpreting the stance as opposing"\\
\ \ \ \ \},\\
\ \ \ \ \{\\
\ \ \ \ \ \ "child\_topic\_name": "Child topic name 2",\\
\ \ \ \ \ \ "child\_topic\_short\_description": "Description of child topic 2",\\
\ \ \ \ \ \ "document\_examples": "Up to three example documents",\\
\ \ \ \ \ \ "opposing\_stance\_reason": "Reason for interpreting the stance as opposing"\\
\ \ \ \ \}\\
\ \ ]\\
\}%
\\[0.8ex]
\# Output format (if no opposing stances are present)\\
\{%
\newline
\ \ "contain\_opposing\_stance": false,\\
\ \ "child\_topics": []\\
\}%
}
&
\promptcell{%
\scriptsize
\# タスク\\
トピックモデルによって作成された1つのトピックに、スタンスが対立する文章が混在しているかどうかを判定してください。混在している場合は、そのトピックを分割してください。\\[0.8ex]
\# 用語の定義\\
・スタンスが対立するとは、同一トピックに分類されるものの、対立する意味合いを持つ文章が含まれている場合を指します。\\
・スタンスが対立する例として、「有る vs 無い」「多い vs 少ない」「強い vs 弱い」などがあります。\\[0.8ex]
\# 要件\\
・スタンスが対立する文章が混在しているかの判定結果（\texttt{contain\_opposing\_stance}）を出力してください。\\
・トピックの分割が必要な場合は、分割された子トピック（\texttt{child\_topics}）のリストを出力してください。\\
・\texttt{child\_topics} の各要素には、子トピック名（\texttt{child\_topic\_name}）、子トピックの説明（\texttt{child\_topic\_short\_description}）、該当する文章例（最大3件、\texttt{document\_examples}）、およびスタンスが対立すると解釈した理由（\texttt{opposing\_stance\_reason}）を含めてください。\\
・分割が不要な場合は、\texttt{child\_topics} は空リストとしてください。\\
・JSON形式で出力してください。\\[0.8ex]
\# 判定における補足定義\\
・該当トピックの情報は、トピック名、トピック説明、およびトピックに該当する文章群に記載されています。\\
・明らかにスタンスが対立していると解釈できる文章が混在している場合のみ、スタンスが対立すると判定してください。\\[0.8ex]
\# 子トピック命名規則の補足定義\\
・名詞句としてください。\\
・簡潔な表現とし、「AとB」「AおよびBに関する〜」のような冗長な表現は避け、単語数はできるだけ少なくしてください。\\
・可能な限り具体的な表現としてください。\\
・親トピック名がなくても、子トピック名のみで意味が明確に解釈できる表現としてください。\\
・トピックモデリング対象の文章全体のメタ情報を考慮した表現としてください。\\[0.8ex]
\# トピック名と説明\\
トピック名: \{topic\_name\}\\
トピック説明: \{topic\_short\_description\}\\[0.8ex]
\# トピックモデリング対象の文章全体のメタ情報\\
\{document\_metadata\}\\[0.8ex]
\# トピックに該当する文章群\\
\{topic\_documents\}\\[0.8ex]
\# 出力形式（スタンスが対立する文章が含まれている場合）\\
\{%
\newline
\ \ "contain\_opposing\_stance": true,\\
\ \ "child\_topics": [\\
\ \ \ \ \{\\
\ \ \ \ \ \ "child\_topic\_name": "子トピック名1",\\
\ \ \ \ \ \ "child\_topic\_short\_description": "子トピック名1の説明",\\
\ \ \ \ \ \ "document\_examples": "子トピック名1の文章例（最大3件）",\\
\ \ \ \ \ \ "opposing\_stance\_reason": "子トピック2とスタンスが対立すると解釈した理由"\\
\ \ \ \ \},\\
\ \ \ \ \{\\
\ \ \ \ \ \ "child\_topic\_name": "子トピック名2",\\
\ \ \ \ \ \ "child\_topic\_short\_description": "子トピック名2の説明",\\
\ \ \ \ \ \ "document\_examples": "子トピック名2の文章例（最大3件）",\\
\ \ \ \ \ \ "opposing\_stance\_reason": "子トピック1とスタンスが対立すると解釈した理由"\\
\ \ \ \ \}\\
\ \ ]\\
\}%
\\[0.8ex]
\# 出力形式（スタンスが対立する文章が含まれていない場合）\\
\{%
\newline
\ \ "contain\_opposing\_stance": false,\\
\ \ "child\_topics": []\\
\}%
}
\\

\bottomrule
\end{tabularx}
\end{table*}

\begin{table*}[tbp]
\centering
\caption{Prompt used for assigning documents to polarity-specific topics in Step~4 of Figure~\ref{fig:proposed_topic_model_workflow}}
\label{tab:split-topic-assignment-prompt}
\small
\setlength{\tabcolsep}{10pt}
\renewcommand{\arraystretch}{1.15}
\begin{tabularx}{\textwidth}{@{} X X @{}}
\toprule
\textbf{English Prompt (Translation)} & \textbf{Japanese Prompt (Original)} \\
\midrule

\promptcell{%
\# Task\\
Determine which of the candidate child topics best matches the input text that has been classified under the parent topic.\\[0.8ex]
\# Requirements\\
- Always output the topic ID (\texttt{topic\_id}), topic name (\texttt{topic\_name}), and the reason for the decision (\texttt{reason}).\\
- If the text does not match any of the candidate child topics, select \texttt{Other}.\\
- Output the result in JSON format.\\[0.8ex]
\# Supplementary guidelines for judgment\\
- Make the judgment by taking into account the metadata of the input text.\\[0.8ex]
\# Parent topic\\
\{parent\_topic\}\\[0.8ex]
\# Input text\\
\{input\_text\}\\[0.8ex]
\# Metadata of the text\\
\{document\_metadata\}\\[0.8ex]
\# Candidate child topics (legend: topic\_id, topic name (topic description))\\
\{child\_topic\_definition\}\\[0.8ex]
\# Output format\\
\{%
\newline
\ \ "topic\_id": int,\\
\ \ "topic\_name": str,\\
\ \ "reason": str\\
\}%
}
&
\promptcell{%
\# タスク\\
親トピックに分類されている文章が、子トピック候補のいずれに一致するかを判断してください。\\[0.8ex]
\# 要件\\
・トピックID（\texttt{topic\_id}）、トピック名（\texttt{topic\_name}）、および判断理由（\texttt{reason}）を必ず出力してください。\\
・いずれの子トピック候補とも一致しない場合は、\texttt{その他} を選択してください。\\
・JSON形式で出力してください。\\[0.8ex]
\# 判定の際の補足定義\\
・文章のメタ情報を考慮して判断してください。\\[0.8ex]
\# 親トピック\\
\{parent\_topic\}\\[0.8ex]
\# 文章\\
\{input\_text\}\\[0.8ex]
\# 文章のメタ情報\\
\{document\_metadata\}\\[0.8ex]
\# トピック候補（凡例：トピックID，トピック名（トピック説明））\\
\{child\_topic\_definition\}\\[0.8ex]
\# 出力形式\\
\{%
\newline
\ \ "topic\_id": int,\\
\ \ "topic\_name": str,\\
\ \ "reason": str\\
\}%
}
\\

\bottomrule
\end{tabularx}
\end{table*}

\begin{table*}[tbp]
\centering
\caption{Prompt used for evaluating polarity stance similarity between all pairs of topics (described in Section~\ref{topic_integration})}
\label{tab:geval-topic-stance-diversity}
\small
\setlength{\tabcolsep}{10pt}
\renewcommand{\arraystretch}{1.15}
\begin{tabularx}{\textwidth}{@{} X @{}}
\toprule
\promptcell{%
\# Criteria name\\
\texttt{topic\_stance\_similarity}\\[0.8ex]
\# Evaluation steps\\
1.\ Read the topic labels and descriptions of the two topics carefully.\\
2.\ Compare the main themes, concepts, and ideas expressed in both topics.\\
3.\ Determine whether the topics are clearly distinct in stances.\\[0.8ex]
\# Criteria (prompt)\\
Are the two topics clearly distinct in stance, describing opposing or mutually exclusive positions on a theme or idea?\\[0.8ex]
\# Rubric (score interpretation)\\
0--2:\ The two topics have almost the same stance (very low stance diversity).\\
3--5:\ The topics are somewhat distinct in stance (low stance diversity).\\
6--8:\ The topics are mostly different in stance (moderate stance diversity).\\
9--10:\ The topics are clearly distinct, expressing opposing or mutually exclusive positions on a theme or idea (high stance diversity).%
}
\\
\bottomrule
\end{tabularx}
\vspace{1mm}
\begin{minipage}{\textwidth}
\footnotesize
\textit{Notes:}The evaluator is configured by defining the evaluation steps and criterion question through a \textit{Criteria} function, while qualitative score meanings are provided independently via a \textit{Rubric} list that specifies expected outcomes for each score range.
\end{minipage}
\end{table*}

\begin{table*}[tbp]
\centering
\caption{Prompt used for evaluatiing Topic Label Alignment (described in Section~\ref{topic_label_alignment})}
\label{tab:geval-topic-label-alignment}
\small
\setlength{\tabcolsep}{10pt}
\renewcommand{\arraystretch}{1.15}
\begin{tabularx}{\textwidth}{@{} X @{}}
\toprule
\promptcell{%
\# Criteria name\\
\texttt{Topic Label Alignment}\\[0.8ex]
\# Evaluation steps\\
1.\ Read the topic label and topic description carefully.\\
2.\ Read the given document associated with the topic.\\
3.\ For the given document, strictly judge whether its main meaning, theme, and details are fully and semantically captured by the topic label and description, and vice versa.\\
4.\ If any meaning-level mismatch, omission, or extraneous concept is found between the document and the label and description, even if minor, count the document as misaligned.\\[0.8ex]
\# Criteria (prompt)\\
For the document, do the topic label and description align completely and semantically with its content?\\[0.8ex]
\# Rubric (score interpretation)\\
0--2:\ The document is largely misaligned with the topic label and description; its main meaning, theme, or details differ substantially, and the label fails to capture the document's semantic core.\\
3--5:\ The document shows partial alignment, but key meanings or important details are missing or incorrectly represented.\\
6--8:\ The document is mostly aligned; minor omissions or slight semantic mismatches are present, but the overall meaning is adequately captured.\\
9--10:\ The document is fully and semantically aligned; its central meaning, theme, and key details are precisely and completely represented.%
}
\\
\bottomrule
\end{tabularx}
\vspace{1mm}
\begin{minipage}{\textwidth}
\footnotesize
\textit{Notes:}In the G-Eval implementation, the evaluation steps and criterion question are specified through a \textit{Criteria} function, while score meanings are defined separately via a list of \textit{Rubric} objects that map score ranges (0--10) to qualitative levels of topic label alignment.
\end{minipage}
\end{table*}

\begin{table*}[tbp]
\centering
\caption{Prompt used for evaluating Semantic-based Topic Diversity (described in Section~\ref{semantic_based_topic_diversity})}
\label{tab:semantic-topic-diversity}
\small
\setlength{\tabcolsep}{10pt}
\renewcommand{\arraystretch}{1.15}
\begin{tabularx}{\textwidth}{@{} X X @{}}
\toprule
\textbf{English Prompt (Translation)} & \textbf{Japanese Prompt (Original)} \\
\midrule

\promptcell{%
\# Task\\
To evaluate the diversity of topic modeling results, judge the semantic similarity between two topics.\\[0.8ex]
\# Requirements\\
- Based on the topic names and descriptions, evaluate similarity using the criteria below and assign a score on a 10-point scale.\\
- Output the reason for the assigned score.\\
- Output the results in JSON format.\\[0.8ex]
\# Evaluation criteria\\
- Whether the topic names describe similar content.\\
- Whether the topic descriptions describe similar content.\\[0.8ex]
\# Definition of similarity\\
- The two topics are described at a comparable level of granularity.\\
- The two topics share similar evaluative or affective nuances (e.g., positive vs.\ negative).\\[0.8ex]
\# Topic 1\\
Topic name: \{topic\_name\_1\}\\
Topic description: \{topic\_short\_description\_1\}\\[0.8ex]
\# Topic 2\\
Topic name: \{topic\_name\_2\}\\
Topic description: \{topic\_short\_description\_2\}\\[0.8ex]
\# Examples\\
\textbf{Score 0--2: Completely different content}\\
Topic 1: ``Lack of teamwork'' (inability or unwillingness to cooperate with team members).\\
Topic 2: ``One-on-one meetings'' (regular one-on-one meetings between supervisors and subordinates).\\[0.6ex]
\textbf{Score 3--5: Partially related, but different in granularity or nuance}\\
Topic 1: ``Lack of teamwork'' (attitudes or behaviors reflecting inability or unwillingness to cooperate).\\
Topic 2: ``Teamwork culture'' (organizational culture regarding collaboration and cooperation).\\[0.6ex]
\textbf{Score 6--8: Semantically similar despite lexical differences}\\
Topic 1: ``Lack of teamwork'' (difficulty or reluctance to collaborate).\\
Topic 2: ``Passive teamwork'' (collaboration characterized by passive attitudes).\\[0.6ex]
\textbf{Score 10: Identical wording and content}\\
Topic names and descriptions are fully identical.\\[0.8ex]
\# Output format\\
\{%
\newline
\ \ "score": int,\\
\ \ "reason": str\\
\}%
}
&
\promptcell{%
\# メトリクス名\\
\texttt{Semantic-based Topic Diversity}\\[0.8ex]
\# タスク\\
トピックモデリング結果の多様性を評価するために、2つのトピック内容の類似性を判断してください。\\[0.8ex]
\# 要件\\
・トピック名およびトピック説明をもとに、以下の判定基準を参考に10段階評価でスコアリングしてください。\\
・そのように判断した理由を出力してください。\\
・JSON形式で出力してください。\\[0.8ex]
\# 判定基準\\
・2つのトピック名が似た内容であるか。\\
・2つのトピック説明が似た内容であるか。\\[0.8ex]
\# 「似ている」の定義\\
・内容の粒度が同程度であること。\\
・ポジティブ／ネガティブなどのニュアンスが一致していること。\\[0.8ex]
\# Topic 1\\
トピック名: \{topic\_name\_1\}\\
トピック説明: \{topic\_short\_description\_1\}\\[0.8ex]
\# Topic 2\\
トピック名: \{topic\_name\_2\}\\
トピック説明: \{topic\_short\_description\_2\}\\[0.8ex]
\# 例\\
\textbf{評価スコア0--2：全く異なる内容}\\
Topic 1：チームワークの欠如（協力できない、または協力しようとしない態度や行動）。\\
Topic 2：1on1面談の実施（上司と部下が定期的に面談を行うこと）。\\[0.6ex]
\textbf{評価スコア3--5：一部関連性はあるが粒度やニュアンスが異なる}\\
Topic 1：チームワークの欠如（協力できない態度や行動）。\\
Topic 2：チームワークの風土（協力姿勢や能力に関する文化）。\\[0.6ex]
\textbf{評価スコア6--8：用語は異なるが内容は類似}\\
Topic 1：チームワーク不足（協力できない状況）。\\
Topic 2：消極的チームワーク（協力姿勢が消極的な状態）。\\[0.6ex]
\textbf{評価スコア10：完全に同一}\\
トピック名および説明が完全に一致している場合。\\[0.8ex]
\# 出力形式\\
\{%
\newline
\ \ "score": int,\\
\ \ "reason": str\\
\}%
}
\\

\bottomrule
\end{tabularx}
\vspace{1mm}
\begin{minipage}{\textwidth}
\footnotesize
\textit{Notes:} The illustrative examples are provided in Japanese and may be adapted or replaced depending on the specific task or evaluation context.
\end{minipage}
\end{table*}

\begin{table*}[tbp]
\centering
\caption{Prompt used for evaluating Specificity (described in Section~\ref{specificity})}
\label{tab:geval-topic-specificity}
\small
\setlength{\tabcolsep}{10pt}
\renewcommand{\arraystretch}{1.15}
\begin{tabularx}{\textwidth}{@{} X @{}}
\toprule
\promptcell{%
\# Criteria name\\
\texttt{Specificity}\\[0.8ex]
\# Evaluation steps\\
1.\ Read the topic label and its description carefully.\\
2.\ When it becomes clear that the topic has a positive or negative impact on business performance or employee engagement, evaluate whether the leader --- the subject of the topic --- can easily form an actionable mental image of the behavioral changes they should implement.\\
3.\ Evaluate whether the topic refers to a narrowly defined situation rather than a broad or generalized category of issues.\\
4.\ If the topic relies on overly broad themes or spans multiple unrelated aspects, treat it as low in specificity.\\[0.8ex]
\# Criteria (prompt)\\
This criterion evaluates the topic along two axes: (i) imaginability --- whether a concrete and actionable mental image can be formed; and (ii) specificity --- whether the described situation is narrow and well-defined rather than overly broad or semantically dispersed. Is the topic imaginable and specific enough for the leader?\\[0.8ex]
\# Rubric (score interpretation)\\
0--2:\ Extremely low specificity and imaginability. The topic is abstract, overly broad, or mixes multiple unrelated aspects, preventing a coherent mental image. The leader cannot visualize who is acting, what is happening, or in what situation. Example: 組織迷走と多問題化（経営層の方向性が不明確なまま、複数の問題が同時に生じている状況）。\\
3--5:\ Low specificity and imaginability. Some concrete elements are present, but the topic remains broad or semantically dispersed, making it difficult to form a single actionable scenario. The leader can grasp the general idea but not a coherent behavioral change. Example: 新規事業推進の負荷増大（意思決定遅延と情報共有不足により現場負荷が増加している状態）。\\
6--8:\ Moderate to high specificity and imaginability. The topic is reasonably focused with identifiable actors and actions, allowing a mostly coherent mental image, though some details may remain generalized. Example: 承認停滞を生む業務放置（管理職によるレビュー遅延で業務進行が滞る状況）。\\
9--10:\ Very high specificity and imaginability. The topic is narrow, concrete, and semantically unified, with clear actors, actions, and context. The leader can immediately visualize a vivid and actionable scene. Example: 会議発言遮断による停滞（週次会議で部長が部下の発言を遮る場面）。%
}
\\
\bottomrule
\end{tabularx}
\vspace{1mm}
\begin{minipage}{\textwidth}
\footnotesize
\textit{Notes:}In the G-Eval implementation, the evaluation procedure and criterion question are defined via a \textit{Criteria} function, while score meanings and example-based interpretations are specified independently through a \textit{Rubric} list mapping score ranges (0--10) to qualitative levels of topic specificity. The illustrative rubric examples are provided in Japanese and may be adapted depending on the task.
\end{minipage}
\end{table*}

\begin{table*}[tbp]
\centering
\caption{Prompt used for evaluating Polarity Stance Consistency (described in Section~\ref{polarity_stance_consisitency})}
\label{tab:geval-polarity-stance-consistency}
\small
\setlength{\tabcolsep}{10pt}
\renewcommand{\arraystretch}{1.15}
\begin{tabularx}{\textwidth}{@{} X @{}}
\toprule
\promptcell{%
\# Criteria name\\
\texttt{Polarity Stance Consistency}\\[0.8ex]
\# Evaluation steps\\
1.\ Read the topic label and description carefully.\\
2.\ Paraphrase the main phenomenon, condition, or state described, without considering emotional or evaluative direction.\\
3.\ Consider whether the topic could plausibly be interpreted as describing more than one mutually exclusive or opposite state, such as presence vs.\ absence, strong vs.\ weak, positive vs.\ negative, or increase vs.\ decrease. For example, topics like ``manager influence,'' ``job satisfaction,'' or ``work--life balance'' may refer to either high or low levels, presence or absence, or improvement or decline.\\
4.\ List the main plausible interpretations regarding the presence, absence, or degree of the phenomenon. If any pair of interpretations are mutually exclusive or opposites, mark the topic as inconsistent. If only a single meaning or state is reasonably plausible, mark it as consistent.\\[0.8ex]
\# Criteria (prompt)\\
Do the topic label and description allow for mutually exclusive or opposite meanings (e.g., presence vs.\ absence, high vs.\ low, increase vs.\ decrease)? If any pair of plausible interpretations are opposites or mutually exclusive, the topic is inconsistent, regardless of evaluative direction. If only one meaning or state is reasonably plausible, the topic is consistent.\\[0.8ex]
\# Rubric (score interpretation)\\
0--2:\ The topic is clearly contradictory or contains explicitly opposing stances, making it impossible to assign a single position. Example: ``Manager's management of subordinates'' (describes various and opposing behaviors and attitudes without indicating a clear stance).\\
3--5:\ The topic somewhat includes opposing or conflicting stances. Both positive and negative interpretations are possible, but one may be slightly more dominant.\\
6--8:\ The topic is generally consistent in stance, though minor ambiguity or alternative interpretations are possible.\\
9--10:\ The topic is clearly consistent, expressing a single and unambiguous stance. Example: ``Supportive management practices'' (clearly indicates a positive stance).%
}
\\
\bottomrule
\end{tabularx}
\vspace{1mm}
\begin{minipage}{\textwidth}
\footnotesize
\textit{Notes:}In the G-Eval implementation, the evaluation steps and criterion question are defined via a \textit{Criteria} function, while score meanings and example-based interpretations are specified independently through a \textit{Rubric} list mapping score ranges (0--10) to qualitative levels of polarity stance consistency.
\end{minipage}
\end{table*}

\FloatBarrier
\section{Full Results on Topic and Outcomes Relationship}

\begin{table*}[t]
\centering
\scriptsize
\setlength{\tabcolsep}{2pt}
\renewcommand{\arraystretch}{0.92}
\caption{Relationship between leadership topics and firm outcomes}
\label{tab:relationship_of_leadership_topic_and_outcomes}
\begin{adjustbox}{max width=\textwidth}
\begin{tabular}{>{\raggedright\arraybackslash}p{0.07\textwidth}>{\raggedright\arraybackslash}p{0.08\textwidth}>{\raggedright\arraybackslash}p{0.18\textwidth}>{\raggedright\arraybackslash}p{0.30\textwidth}>{\centering\arraybackslash}p{0.06\textwidth}>{\centering\arraybackslash}p{0.06\textwidth}>{\centering\arraybackslash}p{0.06\textwidth}>{\centering\arraybackslash}p{0.06\textwidth}>{\centering\arraybackslash}p{0.06\textwidth}>{\centering\arraybackslash}p{0.06\textwidth}>{\centering\arraybackslash}p{0.05\textwidth}}
\toprule
Type. & Char. & Translated Topic Name (Original Topic Name) & Translated Topic Description (Original Topic Description) & \multicolumn{3}{c}{ROA} & \multicolumn{3}{c}{Employees Morale} & N \\
\cmidrule(lr){5-7}\cmidrule(lr){8-10}
 &  &  &  & $t=5$ & $t=10$ & $t=15$ & $t=5$ & $t=10$ & $t=15$ &  \\
\midrule
Non-top & Ability & Advanced Management Competence (高いマネジメント能力) & A collection of texts demonstrating that leaders possess excellent leadership, coordination skills, and rational organizational management abilities, reflecting high-level management expertise. (リーダーが優れた統率力や調整力、合理的な組織運営能力を持ち、マネジメントスキルが高いことを示す文章群。) & -4.44 (6.78) & -5.13 (11.71) & -16.74 (15.70) & -0.00 (0.40) & 1.17 (0.62)* & 1.53 (0.81)* & 608 \\
\addlinespace[0.3em]
\hdashline
\addlinespace[0.3em]
Non-top & Ability & Advanced Technical and Managerial Competence (高い技術・管理能力) & An evaluation recognizing the superior specialized technical skills and management capabilities of project leaders and managers. (プロジェクトリーダーやマネージャーの専門技術やマネジメント能力が優れている評価) & 45.32 (23.16)* & 66.83 (36.52)* & 93.62 (47.51)** & 4.28 (1.37)*** & 4.11 (1.94)** & 4.81 (2.43)** & 95 \\
\addlinespace[0.3em]
\hdashline
\addlinespace[0.3em]
Non-top & Ability & Competencies of Outstanding Store Managers (優秀な店長の能力) & An evaluation of store managers who demonstrate high-level skills and achievements in store operations and management. (店舗運営やマネジメントにおいて高いスキルや実績を持つ店長に関する評価) & -0.60 (6.51) & -5.03 (14.41) & -12.25 (19.39) & -0.45 (0.38) & 0.98 (0.76) & -0.08 (0.99) & 277 \\
\addlinespace[0.3em]
\hdashline
\addlinespace[0.3em]
Non-top & Ability & Competency Evaluation of Female Managers (女性管理職の能力評価) & Positive and negative assessments regarding the abilities and skills of female managers (女性管理職の実力やスキルに関する肯定的・否定的な評価) & 8.39 (12.10) & 35.18 (24.08) & 50.06 (32.09) & 0.01 (0.72) & 3.51 (1.28)*** & 3.45 (1.65)** & 185 \\
\addlinespace[0.3em]
\hdashline
\addlinespace[0.3em]
Non-top & Ability & Deficiencies in Management Competence (マネジメント能力の低さ) & Opinions evaluating the management abilities of middle managers as weak or insufficient. (中間管理職のマネジメント能力が弱い、低いと評価される意見) & 186.74 (242.34) & 178.69 (244.46) & 158.34 (249.37) & -4.79 (14.33) & -10.42 (13.00) & -12.56 (12.82) & 5 \\
\addlinespace[0.3em]
\hdashline
\addlinespace[0.3em]
Non-top & Ability & Evaluation of Incompetent Supervisors (優秀でない上司の評価) & Reviews assessing direct supervisors or managers as incompetent or exhibiting problematic personal characteristics. (直属の上司や管理職に対して優秀でない、または人格面で問題があると評価されている口コミ) & 68.19 (84.99) & 52.47 (85.76) & -34.82 (132.23) & 0.88 (5.02) & -0.26 (4.56) & 0.97 (6.80) & 5 \\
\addlinespace[0.3em]
\hdashline
\addlinespace[0.3em]
Non-top & Ability & Evaluation of Outstanding Supervisors (優秀な上司の評価) & Reviews assessing the direct supervisors and managerial staff as highly competent. (直属の上司や管理職が優秀であると評価されている口コミ) & -1.21 (8.26) & 21.04 (13.50) & 26.23 (17.42) & 0.54 (0.49) & 1.90 (0.71)*** & 1.20 (0.89) & 509 \\
\addlinespace[0.3em]
\hdashline
\addlinespace[0.3em]
Non-top & Ability & Excellence in Managerial Competence (管理能力の優秀さ) & This topic evaluates the managerial layer as highly capable, with strong management abilities and advanced leadership skills. (マネージャー層が優秀であり、管理能力やマネジメントスキルが高いと評価されている内容) & 12.24 (12.12) & 31.56 (20.43) & 36.52 (25.68) & 2.33 (0.71)*** & 0.70 (1.08) & 1.15 (1.32) & 186 \\
\addlinespace[0.3em]
\hdashline
\addlinespace[0.3em]
Non-top & Ability & Executive Expertise and Practical Competence (役員の高い専門能力と実践力) & Content concerning the outstanding specialized knowledge and high practical abilities possessed by executives and senior management. (役員や幹部が持つ優れた専門知識や高い実践能力に関する内容) & 0.05 (18.86) & -38.49 (42.13) & -73.77 (53.96) & 2.54 (1.11)** & -2.58 (2.20) & -2.61 (2.70) & 63 \\
\addlinespace[0.3em]
\hdashline
\addlinespace[0.3em]
Non-top & Ability & High Competency Evaluation of Managers (管理職の高い能力評価) & An assessment highlighting the high regard for managers' expertise, skills, and achievements. (管理職の専門知識やスキル、実績が高く評価されている内容) & 15.11 (5.87)** & 23.00 (10.30)** & 31.01 (14.81)** & -0.06 (0.35) & 0.71 (0.55) & 0.02 (0.76) & 701 \\
\addlinespace[0.3em]
\hdashline
\addlinespace[0.3em]
Non-top & Ability & High Proficiency in Task Execution (業務遂行能力の高さ) & Opinions evaluating managerial capabilities such as sales skills, coordination abilities, and management competence as demonstrating a high level of task execution proficiency. (管理職の営業力や調整力、マネジメント力など業務遂行能力が高いと評価される意見) & -11.27 (13.91) & 5.38 (22.68) & 14.27 (34.89) & -0.67 (0.82) & 0.15 (1.21) & -0.58 (1.79) & 131 \\
\addlinespace[0.3em]
\hdashline
\addlinespace[0.3em]
Non-top & Ability & Inadequacies and Opacity in Managerial Competence (管理職の能力不足や不透明さ) & This content highlights concerns regarding the competence of managerial staff and the lack of transparency in the criteria for their promotion. (管理職の能力や昇格理由に疑問や不透明さを指摘している内容) & 21.30 (15.73) & 15.97 (35.74) & 3.97 (36.66) & -1.18 (0.92) & -0.41 (1.90) & -1.67 (1.88) & 93 \\
\addlinespace[0.3em]
\hdashline
\addlinespace[0.3em]
Non-top & Ability & Inadequate Competence of Executives (役員の能力不足) & An evaluation concerning the insufficient abilities of executives and managerial personnel in decision-making and management judgment. (役員や管理職の意思決定や経営判断に関する能力が低いことに関する評価) & 28.60 (10.93)*** & 32.98 (21.57) & 45.38 (34.32) & -0.24 (0.65) & -1.03 (1.15) & 0.56 (1.76) & 181 \\
\addlinespace[0.3em]
\hdashline
\addlinespace[0.3em]
Non-top & Ability & Inadequate Competence of Supervisors (上司の能力不足) & Reviews concerning the lack of skills and leadership abilities of supervisors, which negatively impact the growth and evaluation of subordinates. (上司のスキルや指導力が不足しており、部下の成長や評価に悪影響を与える能力に関する口コミ) & 10.23 (12.15) & 19.17 (18.03) & 36.80 (24.36) & 0.15 (0.72) & 1.28 (0.96) & 1.13 (1.25) & 263 \\
\addlinespace[0.3em]
\hdashline
\addlinespace[0.3em]
Non-top & Ability & Insufficient Expertise and Practical Competence among Executives (役員の専門能力不足と実践力の低さ) & This topic addresses the lack of specialized knowledge and low practical skills among executives and senior management. (役員や幹部の専門知識不足や実践能力の低さに関する内容) & -3.26 (35.19) & 10.62 (44.02) & -31.75 (61.43) & 1.29 (2.08) & 1.80 (2.34) & 4.42 (3.16) & 39 \\
\addlinespace[0.3em]
\hdashline
\addlinespace[0.3em]
Non-top & Ability & Insufficient Management Competencies among Middle Managers (中間管理職のマネジメント能力不足) & The issue of inadequate management skills and leadership abilities within the middle management tier, which adversely affects organizational operations. (中間管理職層におけるマネジメントスキルや指導力の不足が組織運営に影響している課題。) & 37.75 (33.13) & 9.30 (40.37) & 11.27 (51.76) & 1.49 (1.96) & 0.91 (2.15) & 2.79 (2.66) & 40 \\
\addlinespace[0.3em]
\hdashline
\addlinespace[0.3em]
Non-top & Ability & Insufficient Operational Competence (業務遂行能力の不足) & Opinions regarding the low operational competence of managerial staff, characterized by deficiencies in problem-solving and coordination skills. (管理職の業務遂行能力が低い、問題解決や調整力が不足している状況に関する意見) & 47.83 (24.95)* & -6.52 (63.53) & -22.11 (64.89) & -0.92 (1.46) & 0.42 (3.35) & -1.36 (3.31) & 49 \\
\addlinespace[0.3em]
\hdashline
\addlinespace[0.3em]
Non-top & Ability & Lack of Competence Among Managerial Staff (管理職の能力不足) & A topic describing the situation in which deficiencies in skills and management capabilities are identified within the managerial ranks. (管理職層における実力やマネジメント能力の不足が指摘されている状況を表すトピック。) & 4.30 (5.37) & 2.43 (9.26) & 21.05 (13.57) & 0.04 (0.32) & -0.70 (0.49) & -0.70 (0.69) & 954 \\
\bottomrule
\end{tabular}
\end{adjustbox}
\end{table*}


\begin{table*}[t]
\centering
\scriptsize
\setlength{\tabcolsep}{2pt}
\renewcommand{\arraystretch}{0.92}
\ContinuedFloat
\caption{Relationship between leadership topics and firm outcomes (continued)}
\begin{adjustbox}{max width=\textwidth}
\begin{tabular}{>{\raggedright\arraybackslash}p{0.07\textwidth}>{\raggedright\arraybackslash}p{0.08\textwidth}>{\raggedright\arraybackslash}p{0.18\textwidth}>{\raggedright\arraybackslash}p{0.30\textwidth}>{\centering\arraybackslash}p{0.06\textwidth}>{\centering\arraybackslash}p{0.06\textwidth}>{\centering\arraybackslash}p{0.06\textwidth}>{\centering\arraybackslash}p{0.06\textwidth}>{\centering\arraybackslash}p{0.06\textwidth}>{\centering\arraybackslash}p{0.06\textwidth}>{\centering\arraybackslash}p{0.05\textwidth}}
\toprule
Type. & Char. & Translated Topic Name (Original Topic Name) & Translated Topic Description (Original Topic Description) & \multicolumn{3}{c}{ROA} & \multicolumn{3}{c}{Employees Morale} & N \\
\cmidrule(lr){5-7}\cmidrule(lr){8-10}
 &  &  &  & $t=5$ & $t=10$ & $t=15$ & $t=5$ & $t=10$ & $t=15$ &  \\
\midrule
Non-top & Ability & Lack of Competence Among Managers (管理職の能力不足) & This refers to the situation in which managers lack specialized knowledge, practical skills, and management capabilities. (管理職の専門知識や実践スキル、マネジメント能力が不足している状況を示す。) & -0.83 (3.52) & -3.25 (5.80) & 1.18 (8.38) & -0.52 (0.21)** & -0.84 (0.31)*** & -0.93 (0.43)** & 2238 \\
\addlinespace[0.3em]
\hdashline
\addlinespace[0.3em]
Non-top & Ability & Lack of Competence Among Middle Management (中間管理職の能力不足) & A situation in which deficiencies in specialized knowledge and management skills have been identified among leaders at the department head and section manager levels. (部長や課長クラスのリーダー層における専門知識やマネジメントスキルの不足が指摘されている状況。) & -3.34 (8.67) & -14.31 (14.84) & -49.55 (21.58)** & -0.40 (0.51) & -0.58 (0.79) & -0.37 (1.11) & 395 \\
\addlinespace[0.3em]
\hdashline
\addlinespace[0.3em]
Non-top & Ability & Lack of Competence and Weak Management Skills (能力不足・管理力の弱さ) & An evaluation indicating insufficient technical expertise and managerial capabilities of project leaders and managers. (プロジェクトリーダーやマネージャーの技術力やマネジメント能力が不足している評価) & 79.87 (44.93)* & 168.28 (59.52)*** & 158.83 (60.73)*** & -0.50 (2.65) & 2.39 (3.15) & 0.42 (3.11) & 24 \\
\addlinespace[0.3em]
\hdashline
\addlinespace[0.3em]
Non-top & Ability & Lack of Knowledge and Experience among Managers (管理職の知識・経験不足) & This issue highlights the deficiencies of managers in understanding operations, possessing specialized knowledge, and sharing a common vision. (管理職が業務理解や専門知識、ビジョンの共有において不足している課題を示す。) & -1.63 (6.24) & -2.89 (10.85) & 0.85 (16.45) & -0.84 (0.37)** & -1.97 (0.57)*** & -1.93 (0.84)** & 719 \\
\addlinespace[0.3em]
\hdashline
\addlinespace[0.3em]
Non-top & Ability & Lack of Leadership Skills (リーダーのスキル不足) & This issue highlights the generally low levels of management, technical expertise, and operational execution abilities among managerial leaders. (管理層リーダーのマネジメントや技術力、業務遂行能力が全般的に低い課題を示す。) & 12.13 (7.86) & 10.45 (12.50) & 11.37 (17.73) & -0.07 (0.46) & 0.15 (0.66) & 0.21 (0.91) & 555 \\
\addlinespace[0.3em]
\hdashline
\addlinespace[0.3em]
Non-top & Ability & Lack of Management Competence (マネジメント能力の不足) & This topic highlights the current situation in which managerial personnel lack specialized knowledge and management skills. (管理職層の専門知識やマネジメントスキルが不足している現状を示すトピックです。) & 9.48 (5.63)* & 5.67 (9.27) & 12.37 (13.81) & -0.37 (0.33) & -1.05 (0.49)** & -0.55 (0.71) & 899 \\
\addlinespace[0.3em]
\hdashline
\addlinespace[0.3em]
Non-top & Ability & Lack of Managerial Competence (管理能力の不足) & The managerial abilities and operational management skills of the managerial level are insufficient or have been evaluated as low. (マネージャー層の管理能力や業務運営スキルが不足している、または低いと評価されている内容) & 38.74 (10.68)*** & 68.21 (18.12)*** & 120.71 (26.49)*** & 2.42 (0.62)*** & 2.97 (0.95)*** & 3.05 (1.36)** & 249 \\
\addlinespace[0.3em]
\hdashline
\addlinespace[0.3em]
Non-top & Ability & Low Management Competence (低いマネジメント能力) & A collection of texts pointing out that a leader's management skills are immature or insufficient. (リーダーのマネジメント能力が未熟であったり、能力不足であることを指摘する文章群。) & 15.45 (9.18)* & 47.55 (18.05)*** & 62.08 (23.21)*** & -0.71 (0.54) & 0.96 (0.96) & 1.22 (1.19) & 305 \\
\addlinespace[0.3em]
\hdashline
\addlinespace[0.3em]
Non-top & Ability & Management Challenges of Female Managers (女性管理職のマネジメント課題) & Issues and challenges related to the management methods and approaches of female managers and supervisors. (女性管理職や上司のマネジメント方法や対応に関する課題や問題点) & -4.64 (19.38) & -23.37 (33.22) & -17.19 (51.87) & -1.04 (1.14) & -0.78 (1.76) & 3.03 (2.66) & 46 \\
\addlinespace[0.3em]
\hdashline
\addlinespace[0.3em]
Non-top & Ability & Managerial Incompetence of Store Managers (店長の能力不足) & Negative evaluations and criticisms regarding store managers' management skills and lack of specialized knowledge. (店長のマネジメント能力や専門知識に対する否定的な評価や能力不足の指摘) & -17.20 (13.61) & 28.23 (26.17) & 21.71 (32.44) & -0.19 (0.80) & -1.47 (1.39) & -0.03 (1.67) & 116 \\
\addlinespace[0.3em]
\hdashline
\addlinespace[0.3em]
Non-top & Ability & Outstanding Middle Management (優秀な中間管理職) & Content concerning middle managers who are highly skilled in practical abilities and management, and are recognized as outstanding. (実務能力やマネジメントスキルが高く、優秀であると評価される中間管理職に関する内容) & 9.58 (6.92) & 9.19 (10.82) & -1.45 (14.68) & 0.51 (0.41) & 1.30 (0.57)** & 0.76 (0.75) & 700 \\
\addlinespace[0.3em]
\hdashline
\addlinespace[0.3em]
Non-top & Ability & Superior Competence of Supervisors (上司の高い能力) & Reviews concerning supervisors' exceptional expertise and leadership skills, and their positive impact on the growth and evaluation of subordinates. (上司の専門知識や指導力が優れており、部下の成長や評価に良い影響を与える能力に関する口コミ) & -1.17 (6.89) & 6.12 (11.93) & 20.77 (16.91) & 0.62 (0.41) & 1.01 (0.63) & 1.75 (0.86)** & 586 \\
\addlinespace[0.3em]
\hdashline
\addlinespace[0.3em]
Non-top & Ability & Variability in Management Competence (マネジメント能力のばらつき) & An evaluation of individual differences and disparities in management competence among middle managers. (中間管理職のマネジメント能力における個人差や能力の差異に関する評価) & -41.26 (28.15) & 44.57 (47.44) & 101.31 (83.98) & -0.34 (1.66) & -1.01 (2.52) & -2.27 (4.32) & 36 \\
\addlinespace[0.3em]
\hdashline
\addlinespace[0.3em]
Non-top & Ability & Vulnerabilities in the Competencies of Middle Management (能力が脆弱な中間管理職) & Issues highlighting the insufficiency or specific weaknesses in the competencies of middle management. (中間管理職の能力が不足している、または具体的な能力が弱いと指摘される内容) & -5.34 (21.84) & -4.10 (41.26) & -37.92 (57.93) & -0.22 (1.29) & 0.86 (2.19) & 0.87 (2.98) & 47 \\
\addlinespace[0.3em]
\hdashline
\addlinespace[0.3em]
Non-top & Attitude & Attitudes of Conservative Managers (保守的な管理職の態度) & The tendency of managers to adhere to traditional mindsets, seniority-based systems, and outdated organizational cultures, thereby exhibiting resistance to change. (管理職が昔ながらの考え方や年功序列、古い体質に固執し、変化を拒む態度を示す様子) & 6.11 (6.08) & 12.35 (9.95) & 11.89 (13.22) & -0.64 (0.36)* & -0.95 (0.53)* & -1.13 (0.68)* & 820 \\
\addlinespace[0.3em]
\hdashline
\addlinespace[0.3em]
Non-top & Attitude & Attitudes of Managers Adapting to Changing Times (時代に適応する管理職の態度) & The flexible mindset of managers who either strive to respond to evolving trends and changes or consider the current approach to be adequate. (管理職が時代の流れや変化に対応しようとする、または現状の考え方で良いと考える柔軟な姿勢) & -37.88 (37.84) & -41.22 (56.03) & -89.25 (84.28) & -0.35 (2.24) & 0.90 (2.98) & -0.21 (4.33) & 12 \\
\addlinespace[0.3em]
\hdashline
\addlinespace[0.3em]
Non-top & Attitude & Avoidance of Responsibility and Psychological Burden (責任回避と精神的負担) & An attitude reflecting the internal negative aspects wherein managerial staff exhibit a reluctance to assume responsibility and experience psychological exhaustion, being overwhelmed by pressure. (管理職が責任を取らない態度や精神的に疲弊し、プレッシャーに押し潰される内面的な負の側面を示す態度。) & 21.83 (13.02)* & 35.49 (19.48)* & 38.34 (24.11) & -0.29 (0.77) & -1.77 (1.03)* & -3.14 (1.24)** & 273 \\
\addlinespace[0.3em]
\hdashline
\addlinespace[0.3em]
Non-top & Attitude & Career Ambition and Aspirations for Advancement (出世意欲・上昇志向) & Attitudes toward the desire for promotion and career advancement among middle managers. (中間管理職の出世や昇進に対する意欲や上昇志向に関する態度) & -6.09 (17.44) & -4.46 (24.98) & -5.59 (41.40) & -2.94 (0.97)*** & -3.38 (1.33)** & -2.45 (2.13) & 101 \\
\bottomrule
\end{tabular}
\end{adjustbox}
\end{table*}


\begin{table*}[t]
\centering
\scriptsize
\setlength{\tabcolsep}{2pt}
\renewcommand{\arraystretch}{0.92}
\ContinuedFloat
\caption{Relationship between leadership topics and firm outcomes (continued)}
\begin{adjustbox}{max width=\textwidth}
\begin{tabular}{>{\raggedright\arraybackslash}p{0.07\textwidth}>{\raggedright\arraybackslash}p{0.08\textwidth}>{\raggedright\arraybackslash}p{0.18\textwidth}>{\raggedright\arraybackslash}p{0.30\textwidth}>{\centering\arraybackslash}p{0.06\textwidth}>{\centering\arraybackslash}p{0.06\textwidth}>{\centering\arraybackslash}p{0.06\textwidth}>{\centering\arraybackslash}p{0.06\textwidth}>{\centering\arraybackslash}p{0.06\textwidth}>{\centering\arraybackslash}p{0.06\textwidth}>{\centering\arraybackslash}p{0.05\textwidth}}
\toprule
Type. & Char. & Translated Topic Name (Original Topic Name) & Translated Topic Description (Original Topic Description) & \multicolumn{3}{c}{ROA} & \multicolumn{3}{c}{Employees Morale} & N \\
\cmidrule(lr){5-7}\cmidrule(lr){8-10}
 &  &  &  & $t=5$ & $t=10$ & $t=15$ & $t=5$ & $t=10$ & $t=15$ &  \\
\midrule
Non-top & Attitude & Decline in Motivation and Ambition (意欲・向上心の低下) & Issues related to the lack of motivation, ambition, and a diminished sense of responsibility observed among middle managers. (中間管理職に見られるやる気や向上心の欠如、責任感の薄さに関する問題) & -10.53 (7.93) & -13.03 (12.93) & -25.78 (19.24) & -0.78 (0.47)* & -1.00 (0.68) & -2.07 (0.98)** & 447 \\
\addlinespace[0.3em]
\hdashline
\addlinespace[0.3em]
Non-top & Attitude & Decline in Motivation and Willingness (意欲・やる気の低下) & This topic addresses the lack of motivation and willingness observed among managerial personnel, as well as their passive attitudes toward work. (管理職層に見られるやる気や意欲の欠如、仕事に対する消極的な態度に関する内容。) & -16.45 (7.03)** & -31.81 (11.70)*** & -39.08 (16.82)** & -0.74 (0.41)* & -1.62 (0.62)*** & -2.37 (0.86)*** & 545 \\
\addlinespace[0.3em]
\hdashline
\addlinespace[0.3em]
Non-top & Attitude & Denial of Paid Leave Acquisition (有給休暇取得否定) & Reviews concerning managerial attitudes that are negative or reluctant toward the use of paid leave, reflecting a stance that does not endorse taking paid leave. (管理職が有給休暇取得に否定的、または消極的な態度を示し、取得を良しとしない姿勢に関する口コミ。) & 3.65 (16.13) & -18.12 (28.62) & -50.81 (44.57) & -1.68 (0.95)* & -0.70 (1.52) & -0.41 (2.28) & 89 \\
\addlinespace[0.3em]
\hdashline
\addlinespace[0.3em]
Non-top & Attitude & Favorable Personality and Influence (好意的な人柄と影響力) & The manner in which the branch manager's personality and attitude positively affect the workplace atmosphere and evaluations. (支店長の人柄や態度が職場の雰囲気や評価に良い影響を与えている様子) & -16.76 (13.26) & -9.58 (24.36) & -5.89 (39.41) & -0.54 (0.78) & -1.53 (1.29) & -2.35 (2.02) & 94 \\
\addlinespace[0.3em]
\hdashline
\addlinespace[0.3em]
Non-top & Attitude & Indifference to Managerial Development and Lack of Growth Mindset (管理職の育成無関心と成長意識の欠如) & This refers to a situation in which managers show a lack of interest in the development and career advancement of their subordinates, coupled with a low awareness of growth and improvement. (管理職が部下の育成やキャリア開発に関心を持たず、成長や改善への意識が低い態度を示す状況を表しています。) & -2.04 (8.48) & -5.51 (15.08) & -20.90 (20.45) & -0.22 (0.50) & -1.01 (0.80) & -1.84 (1.05)* & 382 \\
\addlinespace[0.3em]
\hdashline
\addlinespace[0.3em]
Non-top & Attitude & Lack of Consideration and Compassion (配慮・優しさの欠如) & An internal attitude wherein leaders or supervisors fail to demonstrate consideration and compassion toward their subordinates or workplace, thereby lacking a sense of responsibility and trustworthiness. (リーダーや上司が部下や現場に対して配慮や優しさを欠き、責任感や信頼を示さない内面的姿勢。) & -9.32 (6.49) & -5.06 (10.85) & 1.19 (15.47) & -0.93 (0.38)** & -0.83 (0.58) & -1.10 (0.79) & 553 \\
\addlinespace[0.3em]
\hdashline
\addlinespace[0.3em]
Non-top & Attitude & Leaders' Internal Attitudes and Character (リーダーの内面的姿勢と人柄) & This topic describes how the internal attitudes and personal qualities of managers—such as their spirit, cooperativeness, and integrity—influence the team climate and operational performance. (管理職の心意気や協調性、誠実さなどの内面的な態度や人柄がチームの風土や業務に影響を与える様子を表す。) & -2.32 (4.26) & 8.55 (7.22) & 10.54 (9.94) & 0.30 (0.25) & 0.53 (0.38) & 1.41 (0.51)*** & 1528 \\
\addlinespace[0.3em]
\hdashline
\addlinespace[0.3em]
Non-top & Attitude & Leniency in Subjective Evaluations (主観的評価の甘さ) & Reviews concerning the tendency of supervisors to provide lenient evaluations or preferential treatment based on personal likes, dislikes, or emotions. (上司が好き嫌いや個人的感情で評価を甘くしたり優遇したりする傾向に関する口コミ。) & 10.70 (9.19) & 43.00 (18.86)** & 34.26 (24.68) & -0.66 (0.54) & -0.07 (1.00) & 1.32 (1.27) & 222 \\
\addlinespace[0.3em]
\hdashline
\addlinespace[0.3em]
Non-top & Attitude & Managers Demonstrating Understanding of Parental Leave (育児休業に理解を示す管理職) & Reviews of managers who exhibit understanding and supportive attitudes toward parental leave and child-rearing. (育児休業や子育てに対して理解や協力的な態度を持つ管理職の口コミ。) & -37.91 (23.45) & 26.27 (34.00) & 32.90 (40.98) & -1.50 (1.39) & 0.74 (1.81) & 0.34 (2.10) & 126 \\
\addlinespace[0.3em]
\hdashline
\addlinespace[0.3em]
Non-top & Attitude & Managers with Negative Attitudes Toward Parental Leave (育児休業に否定的な管理職) & Reviews from managers who demonstrate limited understanding of parental leave and child-rearing, and who hold negative or unfavorable attitudes toward these matters. (育児休業や子育てに対して理解が乏しく、否定的または好ましく思わない態度を持つ管理職の口コミ。) & -24.52 (21.65) & -35.98 (28.36) & -37.87 (39.81) & -0.79 (1.28) & -1.89 (1.50) & 0.21 (2.04) & 85 \\
\addlinespace[0.3em]
\hdashline
\addlinespace[0.3em]
Non-top & Attitude & Negative Attitudes Toward Female Managers (女性管理職への否定的態度) & Reviews reflecting negative perceptions, discriminatory attitudes, and a lack of understanding toward female managers. (女性管理職に対する否定的な見方や差別的態度、理解不足を示す口コミ。) & -5.56 (9.50) & -2.00 (13.42) & 5.24 (20.06) & 0.40 (0.56) & 0.99 (0.71) & 1.00 (1.03) & 373 \\
\addlinespace[0.3em]
\hdashline
\addlinespace[0.3em]
Non-top & Attitude & Negative Attitudes of Supervisors (否定的な上長の態度) & Descriptions concerning supervisors' negative, critical, or problematic attitudes and personalities. (上長の否定的、批判的、または問題のある態度や性格に関する記述) & 0.79 (5.33) & 5.00 (8.48) & 0.69 (12.32) & 0.03 (0.31) & 0.20 (0.45) & 0.26 (0.63) & 986 \\
\addlinespace[0.3em]
\hdashline
\addlinespace[0.3em]
Non-top & Attitude & Negative Evaluation of Long Working Hours (長時間労働の否定的評価) & The perspective held by managerial personnel that long working hours and overtime are detrimental or problematic, leading to their negative assessment. (管理職層が長時間労働や残業を悪や問題と捉え、否定的に評価する価値観。) & -25.61 (21.92) & -13.51 (45.38) & -8.66 (56.00) & 0.41 (1.29) & -0.99 (2.38) & -2.74 (2.82) & 61 \\
\addlinespace[0.3em]
\hdashline
\addlinespace[0.3em]
Non-top & Attitude & Negative Personality Traits and Their Impact on Influence (否定的な人柄と影響力) & The manner in which a branch manager's personality and attitude adversely affect the workplace atmosphere and evaluations. (支店長の人柄や態度が職場の雰囲気や評価に悪い影響を与えている様子) & -22.63 (15.30) & -33.02 (23.52) & -35.46 (28.07) & -2.44 (0.90)*** & -1.94 (1.25) & -3.16 (1.44)** & 99 \\
\addlinespace[0.3em]
\hdashline
\addlinespace[0.3em]
Non-top & Attitude & Negative and Strict Attitudes (否定的・厳しい態度) & An internal disposition in which leaders or supervisors adopt a negative and strict attitude toward subordinates, characterized by a lack of trust and consideration. (リーダーや上司が部下に対して否定的、厳しい態度をとり、信頼や配慮が欠如している内面的姿勢。) & -9.09 (12.27) & -11.08 (18.66) & -52.47 (28.14)* & -0.81 (0.73) & -1.11 (0.99) & -1.67 (1.45) & 221 \\
\addlinespace[0.3em]
\hdashline
\addlinespace[0.3em]
Non-top & Attitude & Outdated Values Among Managerial Staff (管理職の旧態依然な価値観) & This refers to the tendency of managerial personnel to cling to outdated ideas and values from the Showa era, resulting in an inability to adapt to contemporary changes. (管理職層が昭和時代の古い考え方や価値観に固執し、現代の変化に対応できていない姿勢を指す。) & -1.74 (5.16) & -1.14 (8.37) & -6.62 (11.28) & -0.27 (0.30) & -0.31 (0.44) & -0.57 (0.58) & 1136 \\
\addlinespace[0.3em]
\hdashline
\addlinespace[0.3em]
Non-top & Attitude & Passive Attitudes (消極的態度) & Internal passive attitudes observed among middle management, including a tendency to maintain the status quo, avoidance of responsibility, and decreased motivation. (中間管理職層に見られる現状維持志向や責任回避、モチベーション低下などの内面的な消極的態度) & 6.15 (6.10) & 1.61 (10.16) & 13.98 (13.08) & -1.36 (0.36)*** & -1.43 (0.54)*** & -2.69 (0.67)*** & 834 \\
\addlinespace[0.3em]
\hdashline
\addlinespace[0.3em]
Non-top & Attitude & Passive Attitudes and Behaviors (消極的な意識と行動) & This topic addresses passive and negative attitudes exhibited by managers, characterized by a lack of motivation and sense of responsibility, limited understanding and expectations, and a tendency to neglect management duties. (マネージャーがやる気や責任感に欠け、理解や期待が薄い、マネジメントを軽視するなど消極的・否定的な態度に関する内容。) & 39.30 (16.54)** & 80.08 (27.64)*** & 122.98 (35.44)*** & 0.86 (0.98) & 1.28 (1.46) & 0.37 (1.80) & 137 \\
\bottomrule
\end{tabular}
\end{adjustbox}
\end{table*}


\begin{table*}[t]
\centering
\scriptsize
\setlength{\tabcolsep}{2pt}
\renewcommand{\arraystretch}{0.92}
\ContinuedFloat
\caption{Relationship between leadership topics and firm outcomes (continued)}
\begin{adjustbox}{max width=\textwidth}
\begin{tabular}{>{\raggedright\arraybackslash}p{0.07\textwidth}>{\raggedright\arraybackslash}p{0.08\textwidth}>{\raggedright\arraybackslash}p{0.18\textwidth}>{\raggedright\arraybackslash}p{0.30\textwidth}>{\centering\arraybackslash}p{0.06\textwidth}>{\centering\arraybackslash}p{0.06\textwidth}>{\centering\arraybackslash}p{0.06\textwidth}>{\centering\arraybackslash}p{0.06\textwidth}>{\centering\arraybackslash}p{0.06\textwidth}>{\centering\arraybackslash}p{0.06\textwidth}>{\centering\arraybackslash}p{0.05\textwidth}}
\toprule
Type. & Char. & Translated Topic Name (Original Topic Name) & Translated Topic Description (Original Topic Description) & \multicolumn{3}{c}{ROA} & \multicolumn{3}{c}{Employees Morale} & N \\
\cmidrule(lr){5-7}\cmidrule(lr){8-10}
 &  &  &  & $t=5$ & $t=10$ & $t=15$ & $t=5$ & $t=10$ & $t=15$ &  \\
\midrule
Non-top & Attitude & Passive and Laissez-Faire Development Approach (消極的・放任的な育成姿勢) & An approach characterized by a desire for subordinates' growth while exhibiting risk-averse or laissez-faire attitudes, resulting in restrained active involvement. (部下の成長を望みつつもリスク回避や放任的な態度を持ち、積極的な関与を控える姿勢。) & -2.68 (43.37) & -43.52 (55.12) & -52.96 (56.27) & 3.21 (2.56) & 2.97 (2.93) & 1.99 (2.89) & 22 \\
\addlinespace[0.3em]
\hdashline
\addlinespace[0.3em]
Non-top & Attitude & Passive and Negative Attitudes (消極的・否定的な態度) & Customer reviews highlighting store managers' lack of motivation, negative attitudes, and the issues arising from these behaviors. (店舗店長のやる気のなさや否定的な態度、問題点を指摘する口コミ。) & 3.21 (12.03) & 7.62 (22.45) & 6.16 (29.05) & 0.11 (0.69) & 1.01 (1.19) & 3.08 (1.49)** & 131 \\
\addlinespace[0.3em]
\hdashline
\addlinespace[0.3em]
Non-top & Attitude & Passive and Negative Leadership Attitudes (消極的・否定的なリーダーシップ態度) & Employee feedback indicating that managers exhibit a lack of transformative awareness, demonstrate loyalty to the organization by adopting a defensive stance, and show a deficiency in flexibility. (管理職が変革意識に乏しく、組織に忠実で守りに入る態度や柔軟性の欠如を示す口コミ。) & 17.42 (20.90) & -3.56 (28.21) & 10.22 (36.11) & 2.36 (1.23)* & 1.22 (1.50) & 0.84 (1.86) & 117 \\
\addlinespace[0.3em]
\hdashline
\addlinespace[0.3em]
Non-top & Attitude & Positive Attitudes of Supervisors (肯定的な上長の態度) & Descriptions concerning supervisors' commendable personal qualities, proactive attitudes, and trustworthy behaviors. (上長の良い人間性や前向きな姿勢、信頼できる態度に関する記述) & 6.53 (5.48) & 11.65 (9.84) & 19.57 (13.16) & 0.43 (0.32) & 1.60 (0.52)*** & 2.11 (0.67)*** & 890 \\
\addlinespace[0.3em]
\hdashline
\addlinespace[0.3em]
Non-top & Attitude & Positive Evaluation of Female Managers (女性管理職の肯定的評価) & Word-of-mouth that positively assesses the abilities and attitudes of female managers, demonstrating understanding and support. (女性管理職の能力や姿勢を肯定的に評価し、理解や支援を示す口コミ。) & -19.07 (13.50) & -9.91 (20.86) & -28.37 (30.00) & -0.36 (0.80) & 0.65 (1.11) & 0.09 (1.54) & 179 \\
\addlinespace[0.3em]
\hdashline
\addlinespace[0.3em]
Non-top & Attitude & Positive Evaluation of Long Working Hours (長時間労働の肯定的評価) & The perspective among managerial personnel that long working hours and overtime are viewed positively as symbols of effort and responsibility, regarded as virtues. (管理職層が長時間労働や残業を努力や責任感の象徴として肯定的に捉え、美徳と考える価値観。) & 3.75 (7.87) & 4.92 (13.04) & 6.79 (20.87) & -1.01 (0.46)** & -1.98 (0.69)*** & -0.77 (1.07) & 354 \\
\addlinespace[0.3em]
\hdashline
\addlinespace[0.3em]
Non-top & Attitude & Positive Personality Traits (肯定的な人柄) & Internal characteristics that are positively evaluated, such as executives and officers being cheerful and having distinctive personalities. (役員や役職者が明るく個性的であるなど、肯定的に評価される内面的特徴。) & 39.23 (21.60)* & 49.19 (32.31) & 9.58 (52.81) & 0.58 (1.28) & -0.57 (1.72) & 2.63 (2.71) & 49 \\
\addlinespace[0.3em]
\hdashline
\addlinespace[0.3em]
Non-top & Attitude & Proactive Desire for Transformation (積極的変革意欲) & An active attitude and ambitious disposition among middle management toward seeking change and improvement. (中間管理職層の中で変革や改善を望む積極的な姿勢や意欲的な態度) & 6.33 (62.85) & -10.53 (63.45) & -35.98 (116.59) & -1.46 (3.72) & -3.17 (3.37) & 1.90 (5.99) & 10 \\
\addlinespace[0.3em]
\hdashline
\addlinespace[0.3em]
Non-top & Attitude & Proactive Development Approach (積極的な育成姿勢) & An attitude and behavior characterized by enthusiasm for the growth and career development of subordinates, actively engaging in their development. (部下の成長やキャリア開発に熱意を持ち、積極的に育成に取り組む態度や行動。) & 8.00 (11.86) & 9.22 (18.51) & 33.73 (25.47) & 0.12 (0.70) & 1.02 (0.98) & 1.55 (1.30) & 249 \\
\addlinespace[0.3em]
\hdashline
\addlinespace[0.3em]
Non-top & Attitude & Proactive Leadership (積極的なリーダーシップ) & Reviews concerning store managers' motivation, sense of responsibility, proactive attitudes, and positive internal dispositions. (店舗店長のやる気や責任感、積極的な態度や前向きな内面的姿勢に関する口コミ。) & -19.00 (10.51)* & 3.54 (19.34) & 7.18 (24.27) & 0.16 (0.62) & 1.29 (1.02) & 1.47 (1.24) & 149 \\
\addlinespace[0.3em]
\hdashline
\addlinespace[0.3em]
Non-top & Attitude & Proactive Leadership Attitude (積極的なリーダーシップ態度) & Reviews reflecting a proactive internal stance in which managers demonstrate a sense of responsibility and awareness of change, actively promoting challenges and improvements. (管理職が責任感や変革意識を持ち、挑戦や改善を推進する積極的な内面的姿勢を示す口コミ。) & -4.02 (11.62) & 3.52 (21.10) & -13.31 (27.02) & 0.69 (0.69) & -0.52 (1.11) & 0.50 (1.38) & 209 \\
\addlinespace[0.3em]
\hdashline
\addlinespace[0.3em]
Non-top & Attitude & Proactive Mindset and Behavior (積極的な意識と行動) & This topic addresses managers' attitudes characterized by high motivation, a strong sense of responsibility, and a desire for growth, focusing on their proactive engagement in work tasks and personnel development. (マネージャーが高いモチベーションや責任感、成長意欲を持ち、積極的に業務や育成に取り組む姿勢に関する内容。) & 21.62 (10.10)** & 31.95 (15.17)** & 93.76 (22.65)*** & 1.91 (0.60)*** & 2.67 (0.80)*** & 5.75 (1.16)*** & 325 \\
\addlinespace[0.3em]
\hdashline
\addlinespace[0.3em]
Non-top & Attitude & Proactive and Positive Attitude (積極的・前向きな姿勢) & This topic addresses the proactiveness, spirit of challenge, and willingness to work for the benefit of the company and society, as observed among managerial personnel. (管理職層に見られる仕事に対する積極性やチャレンジ精神、会社や社会のために働く意欲に関する内容。) & 40.53 (109.85) & 29.30 (110.82) & 52.17 (231.32) & 2.62 (6.49) & 1.23 (5.89) & 4.86 (11.89) & 4 \\
\addlinespace[0.3em]
\hdashline
\addlinespace[0.3em]
Non-top & Attitude & Promotion of Paid Leave Utilization (有給休暇取得推進) & Reviews concerning the attitudes and behaviors of managerial staff actively encouraging their subordinates to take paid leave, demonstrating understanding and consideration. (管理職が部下の有給休暇取得を積極的に促進し、理解や配慮を示す態度や姿勢に関する口コミ。) & -10.96 (18.83) & 1.66 (28.91) & -2.48 (39.06) & 1.21 (1.11) & 3.00 (1.54)* & 1.33 (2.01) & 95 \\
\addlinespace[0.3em]
\hdashline
\addlinespace[0.3em]
Non-top & Attitude & Self-Centered Attitude (自己中心的態度) & An internal disposition wherein executives and officers exhibit self-preserving and self-prioritizing behaviors, disregarding others and the organization as a whole. (役員や役職者が自己保身や自己優先的な態度を持ち、他者や組織全体を軽視する内面的姿勢。) & -23.72 (7.67)*** & -1.16 (13.37) & 0.17 (19.55) & -1.34 (0.45)*** & -1.77 (0.71)** & -1.75 (0.99)* & 418 \\
\addlinespace[0.3em]
\hdashline
\addlinespace[0.3em]
Non-top & Attitude & Stringency of Evaluations and Perceptions of Unfairness (評価の厳格さと不公平感) & Reviews concerning supervisors who impose strict evaluations based on personal preferences, as well as behaviors that create a sense of opacity and unfairness. (上司が好き嫌いで評価を厳しくしたり不透明感や不公平さを感じさせる態度に関する口コミ。) & -17.44 (20.97) & -41.10 (28.83) & -37.42 (36.96) & -2.16 (1.24)* & -2.64 (1.53)* & -2.30 (1.90) & 106 \\
\addlinespace[0.3em]
\hdashline
\addlinespace[0.3em]
Non-top & Attitude & Strong Sense of Responsibility and Dedicated Attitude (高い責任感と献身的態度) & An intrinsic attitude demonstrated by managers who possess a strong sense of responsibility and approach their work with dedication and seriousness. (管理職が強い責任感を持ち、仕事に対して献身的かつ真面目に取り組む姿勢を示す内面的態度。) & 4.57 (7.21) & 6.05 (12.13) & 28.07 (17.22) & 0.51 (0.43) & 0.19 (0.64) & 0.53 (0.88) & 560 \\
\addlinespace[0.3em]
\hdashline
\addlinespace[0.3em]
Non-top & Attitude & The Attitude of a Dedicated and Empathetic Supervisor (部下思いの熱意ある上司の姿勢) & A topic that describes the internal attitudes and dispositions of supervisors who enhance their subordinates' enthusiasm and motivation by engaging with them in a caring and warm manner. (部下のやる気やモチベーションを高め、親身で温かく接する上司の内面的な態度や姿勢を表すトピック。) & -9.98 (11.43) & 13.22 (17.80) & 10.39 (24.05) & 0.78 (0.67) & 2.24 (0.94)** & 3.76 (1.23)*** & 240 \\
\bottomrule
\end{tabular}
\end{adjustbox}
\end{table*}


\begin{table*}[t]
\centering
\scriptsize
\setlength{\tabcolsep}{2pt}
\renewcommand{\arraystretch}{0.92}
\ContinuedFloat
\caption{Relationship between leadership topics and firm outcomes (continued)}
\begin{adjustbox}{max width=\textwidth}
\begin{tabular}{>{\raggedright\arraybackslash}p{0.07\textwidth}>{\raggedright\arraybackslash}p{0.08\textwidth}>{\raggedright\arraybackslash}p{0.18\textwidth}>{\raggedright\arraybackslash}p{0.30\textwidth}>{\centering\arraybackslash}p{0.06\textwidth}>{\centering\arraybackslash}p{0.06\textwidth}>{\centering\arraybackslash}p{0.06\textwidth}>{\centering\arraybackslash}p{0.06\textwidth}>{\centering\arraybackslash}p{0.06\textwidth}>{\centering\arraybackslash}p{0.06\textwidth}>{\centering\arraybackslash}p{0.05\textwidth}}
\toprule
Type. & Char. & Translated Topic Name (Original Topic Name) & Translated Topic Description (Original Topic Description) & \multicolumn{3}{c}{ROA} & \multicolumn{3}{c}{Employees Morale} & N \\
\cmidrule(lr){5-7}\cmidrule(lr){8-10}
 &  &  &  & $t=5$ & $t=10$ & $t=15$ & $t=5$ & $t=10$ & $t=15$ &  \\
\midrule
Non-top & Attitude & The Authoritative and Distinctive Attitude of a Supervisor (高圧的で個性的な上司の態度) & This describes the internal disposition of a supervisor who exhibits strong demands and an intimidating demeanor toward subordinates, characterized by a unique personality and particular idiosyncrasies. (部下に対して強い要求や威圧的な態度を示し、独特な性格やこだわりを持つ上司の内面的姿勢を表す。) & -0.92 (9.10) & -10.32 (15.69) & -61.76 (23.23)*** & 0.44 (0.54) & 0.01 (0.83) & -0.52 (1.20) & 319 \\
\addlinespace[0.3em]
\hdashline
\addlinespace[0.3em]
Non-top & Attitude & The Internal Attitudes of Sports-Oriented Leaders (体育会系リーダーの内面的姿勢) & The internal attitudes and dispositions observed among managerial and leadership personnel that emphasize perseverance and enthusiasm, grounded in the traditional values of sports-oriented organizational culture. (管理職やリーダー層に見られる、根性やノリを重視し伝統的な体育会系の価値観に基づく内面的な態度や気質。) & 5.73 (9.01) & -8.63 (15.55) & 3.36 (20.82) & -0.15 (0.53) & -0.48 (0.83) & -1.75 (1.07) & 307 \\
\addlinespace[0.3em]
\hdashline
\addlinespace[0.3em]
Non-top & Behavior & Abdication and Delegation of Discretionary Authority (裁量権の放棄・丸投げ) & Opinions regarding store managers who abdicate their discretionary authority in store operations or delegate responsibilities entirely to others. (店長が店舗運営の裁量を放棄したり、丸投げすることに関する意見) & -190.61 (51.67)*** & -60.72 (71.62) & -82.34 (110.80) & -3.49 (3.06) & -1.36 (3.81) & 1.67 (5.70) & 11 \\
\addlinespace[0.3em]
\hdashline
\addlinespace[0.3em]
Non-top & Behavior & Acceptance of Paid Leave Acquisition (有給休暇取得の容認) & Reviews concerning supervisors' and leaders' attitudes that demonstrate a certain level of understanding and approval toward subordinates taking paid leave. (上司やリーダーが部下の有給休暇取得に対して一定の理解や許可を示す態度に関する口コミ。) & 45.75 (39.43) & 22.29 (43.14) & 31.95 (66.26) & 1.28 (2.33) & 0.28 (2.29) & 2.74 (3.41) & 27 \\
\addlinespace[0.3em]
\hdashline
\addlinespace[0.3em]
Non-top & Behavior & Active Acceptance of Opinions (積極的な意見受容) & A behavioral pattern in which leaders proactively listen to their subordinates' opinions and proposals, and engage in reflecting and improving based on them. (リーダーが部下の意見や提案を積極的に聞き入れ、反映や改善に取り組む行動パターン) & 29.36 (18.98) & 27.69 (33.33) & 39.87 (42.16) & -0.51 (1.12) & 1.70 (1.76) & 1.50 (2.14) & 104 \\
\addlinespace[0.3em]
\hdashline
\addlinespace[0.3em]
Non-top & Behavior & Actual Conditions of Overtime Approval and Management (残業容認・管理の実態) & A description concerning the reality in which managers oversee overtime hours but either tolerate excessive overtime or inadequately manage it. (管理職が残業時間を管理しているが、残業が多い状況を容認または管理が不十分な実態に関する記述) & -4.04 (15.60) & -19.62 (24.91) & -26.90 (37.15) & -0.81 (0.92) & -0.80 (1.32) & -2.29 (1.91) & 118 \\
\addlinespace[0.3em]
\hdashline
\addlinespace[0.3em]
Non-top & Behavior & An Unwillingness to Accept Opinions (意見を受け入れない態度) & A behavioral pattern exhibited by supervisors who reject subordinates' opinions and proposals, refuse to listen, and insist on imposing their own ideas. (部下の意見や提案を否定し、聞き入れず自分の考えを押し通す上司の行動パターン。) & -8.28 (6.72) & -12.99 (12.01) & -18.90 (16.39) & -0.35 (0.40) & -1.61 (0.63)** & -1.77 (0.83)** & 588 \\
\addlinespace[0.3em]
\hdashline
\addlinespace[0.3em]
Non-top & Behavior & Application of Fair Evaluation Criteria (正当な評価基準の適用) & Cases in which leaders appropriately assess subordinates based on their performance and behavior. (リーダーが部下の実績や行動に基づき、適切に評価を行う事例) & 40.61 (33.70) & 108.80 (76.32) & 188.81 (94.92)** & -3.12 (1.99) & 0.61 (4.06) & 4.51 (4.88) & 20 \\
\addlinespace[0.3em]
\hdashline
\addlinespace[0.3em]
Non-top & Behavior & Appropriate Evaluation (適正な評価) & Reviews concerning store managers conducting fair evaluations and salary increases based on performance and effort. (店長が実績や努力に基づいて公正に評価や昇給を行うことに関する口コミ) & -9.64 (17.95) & -8.72 (25.79) & -37.61 (39.63) & 0.50 (1.06) & 0.70 (1.37) & 3.35 (2.04)* & 74 \\
\addlinespace[0.3em]
\hdashline
\addlinespace[0.3em]
Non-top & Behavior & Appropriate Task Allocation (適切なタスク配分) & The behavior in which a leader appropriately assigns tasks based on subordinates' abilities and motivation, while coordinating work through communication. (リーダーが部下の能力や意欲に応じて仕事を適切に割り振り、コミュニケーションを取りながら業務を調整する行動) & 18.86 (8.25)** & 13.19 (12.43) & 9.36 (16.98) & -0.01 (0.49) & -0.58 (0.66) & 0.91 (0.87) & 528 \\
\addlinespace[0.3em]
\hdashline
\addlinespace[0.3em]
Non-top & Behavior & Approval and Consideration of Transfer Requests (異動希望の承認と配慮) & The process by which supervisors take into account employees' transfer requests, engage in consultations, and respect their preferences when implementing personnel reassignments. (上司が従業員の異動希望を考慮し、相談に応じたり希望を尊重して配置転換を行う対応。) & 27.74 (11.64)** & 16.98 (20.66) & -10.32 (27.01) & 0.92 (0.69) & 2.01 (1.10)* & 1.64 (1.39) & 262 \\
\addlinespace[0.3em]
\hdashline
\addlinespace[0.3em]
Non-top & Behavior & Attitudes Toward Accepting Opinions (意見を聞き入れる態度) & Behavioral patterns of supervisors who, depending on the situation, may accept and respect subordinates' opinions but exhibit inconsistency in their responses. (部下の意見を場合によっては聞き入れたり尊重するが、対応に差がある上司の行動パターン。) & -0.37 (63.94) & -22.14 (70.60) & -60.13 (83.43) & -4.51 (3.78) & -8.76 (3.75)** & -8.57 (4.28)** & 16 \\
\addlinespace[0.3em]
\hdashline
\addlinespace[0.3em]
Non-top & Behavior & Attitudes Toward Accepting Subordinates' Opinions (部下の意見受容態度) & Reviews concerning supervisors' behaviors and attitudes that actively listen to, respect, and seek to incorporate the opinions of their subordinates. (上司が部下の意見を積極的に聞き入れ、尊重し反映しようとする行動や態度に関する口コミ。) & -0.83 (12.09) & 3.55 (19.09) & 16.32 (27.59) & 1.84 (0.71)** & 1.42 (1.01) & 2.27 (1.42) & 261 \\
\addlinespace[0.3em]
\hdashline
\addlinespace[0.3em]
Non-top & Behavior & Authoritarian Top-Down Decision Making (威圧的なトップダウン意思決定) & The process of decision making in which executives issue directives with a coercive and intimidating attitude, resulting in pressure and confusion at the operational level. (役員層が強権的・威圧的な態度で指示を出し、現場に圧力や混乱をもたらす意思決定の様子) & -8.68 (10.06) & -7.31 (15.28) & -30.22 (21.53) & -1.92 (0.59)*** & -2.83 (0.81)*** & -2.12 (1.10)* & 338 \\
\addlinespace[0.3em]
\hdashline
\addlinespace[0.3em]
Non-top & Behavior & Authoritarian Top-Down Directives (強圧的トップダウン指示) & A behavioral pattern in which supervisors or leaders possess absolute authority and enforce orders in a forceful and coercive manner. (上司やリーダーが絶対的な権力を持ち、強引かつ高圧的に命令を押し通す行動様式) & 1.31 (5.54) & -4.37 (9.04) & -17.52 (12.41) & -0.30 (0.33) & -1.06 (0.48)** & -1.09 (0.64)* & 892 \\
\addlinespace[0.3em]
\hdashline
\addlinespace[0.3em]
Non-top & Behavior & Avoidance and Mitigation of Micromanagement (マイクロマネジメントの回避・緩和) & Behavioral patterns in which middle management and senior leadership refrain from micromanagement and avoid excessive interference with subordinates. (中間管理職や上層部がマイクロマネジメントを控えたり、部下に対して過度な干渉をしない行動パターン) & 77.19 (41.46)* & 194.95 (123.79) & 616.83 (237.80)*** & 3.90 (2.45) & 10.16 (6.58) & 18.96 (12.24) & 5 \\
\addlinespace[0.3em]
\hdashline
\addlinespace[0.3em]
Non-top & Behavior & Awareness of Compliance with Working Hours (労働時間のコンプライアンス意識) & The situation in which store managers are conscious of compliance regarding working hours and overtime, and appropriately manage work schedules. (店舗管理職が労働時間や残業に関してコンプライアンスを意識し、適正な勤務管理を行っている状況) & 14.12 (17.45) & 24.97 (37.51) & 40.39 (48.03) & -0.30 (0.98) & 2.24 (1.99) & 2.16 (2.47) & 55 \\
\addlinespace[0.3em]
\hdashline
\addlinespace[0.3em]
Non-top & Behavior & Behavior of Middle Managers Monitoring Their Supervisors' Reactions (管理職の上司への顔色伺い行動) & This refers to the behavioral pattern in which middle managers and leaders proceed with their work while being attentive to the intentions and facial expressions of higher-ranking executives. (管理職やリーダー層が上位役職者の意向や表情を気にしながら仕事を進める行動パターンを示す。) & -3.27 (4.79) & 11.83 (7.63) & 16.39 (10.89) & -0.98 (0.28)*** & -0.95 (0.40)** & -1.04 (0.56)* & 1268 \\
\bottomrule
\end{tabular}
\end{adjustbox}
\end{table*}


\begin{table*}[t]
\centering
\scriptsize
\setlength{\tabcolsep}{2pt}
\renewcommand{\arraystretch}{0.92}
\ContinuedFloat
\caption{Relationship between leadership topics and firm outcomes (continued)}
\begin{adjustbox}{max width=\textwidth}
\begin{tabular}{>{\raggedright\arraybackslash}p{0.07\textwidth}>{\raggedright\arraybackslash}p{0.08\textwidth}>{\raggedright\arraybackslash}p{0.18\textwidth}>{\raggedright\arraybackslash}p{0.30\textwidth}>{\centering\arraybackslash}p{0.06\textwidth}>{\centering\arraybackslash}p{0.06\textwidth}>{\centering\arraybackslash}p{0.06\textwidth}>{\centering\arraybackslash}p{0.06\textwidth}>{\centering\arraybackslash}p{0.06\textwidth}>{\centering\arraybackslash}p{0.06\textwidth}>{\centering\arraybackslash}p{0.05\textwidth}}
\toprule
Type. & Char. & Translated Topic Name (Original Topic Name) & Translated Topic Description (Original Topic Description) & \multicolumn{3}{c}{ROA} & \multicolumn{3}{c}{Employees Morale} & N \\
\cmidrule(lr){5-7}\cmidrule(lr){8-10}
 &  &  &  & $t=5$ & $t=10$ & $t=15$ & $t=5$ & $t=10$ & $t=15$ &  \\
\midrule
Non-top & Behavior & Behaviors of Not Leaving Work on Time (定時退社しない行動) & Reviews concerning managerial behaviors of working overtime or not leaving work at the scheduled time. (管理職が定時で帰らず残業をする、または早く帰らない行動に関する口コミ。) & -7.88 (25.27) & -39.72 (75.87) & -78.84 (138.69) & -0.72 (1.49) & -1.60 (4.03) & 4.04 (7.13) & 24 \\
\addlinespace[0.3em]
\hdashline
\addlinespace[0.3em]
Non-top & Behavior & Calm Response (穏やかな対応) & Char.s of a supervisor's behavior in responding calmly and without reprimanding subordinates for their mistakes or failure to meet targets. (上司が部下の失敗や数字未達に対して叱責しない、穏やかに接する行動の特徴) & -28.68 (35.51) & -51.67 (117.83) & -67.74 (120.24) & -1.29 (2.10) & -3.41 (6.26) & -4.17 (6.18) & 24 \\
\addlinespace[0.3em]
\hdashline
\addlinespace[0.3em]
Non-top & Behavior & Careful Guidance and Development (丁寧な指導と育成) & The behavior in which a superior dedicates time and effort to sincerely teach subordinates and actively engages in their development. (上司が部下に対して時間や手間をかけて親身に教え、育成に積極的に取り組む行動) & -7.34 (6.06) & -19.66 (10.39)* & -19.39 (14.97) & 0.73 (0.36)** & 0.68 (0.55) & 0.72 (0.76) & 828 \\
\addlinespace[0.3em]
\hdashline
\addlinespace[0.3em]
Non-top & Behavior & Cautious Approval Process (慎重な承認プロセス) & A behavioral pattern in which supervisors and managers require multiple stages of approval and the preparation of detailed documentation, thereby advancing decision-making with careful deliberation. (上司や管理職が複数段階の承認や詳細な資料作成を求め、慎重に意思決定を進める行動様式) & 4.86 (17.76) & 1.32 (25.96) & -0.26 (38.42) & 0.33 (1.05) & 0.78 (1.38) & 0.58 (1.97) & 199 \\
\addlinespace[0.3em]
\hdashline
\addlinespace[0.3em]
Non-top & Behavior & Cautious and Deliberate Decision-Making (慎重で遅い意思決定) & A tendency among executive members to withhold information, resulting in delayed or carefully considered decision-making processes. (役員層が情報を出し渋り、意思決定を遅らせたり慎重に行う様子) & -14.93 (19.85) & -52.38 (33.59) & -48.50 (45.66) & -2.20 (1.17)* & -2.57 (1.77) & -2.19 (2.31) & 83 \\
\addlinespace[0.3em]
\hdashline
\addlinespace[0.3em]
Non-top & Behavior & Challenge Inhibition (挑戦阻害) & Behaviors by supervisors or managers that suppress subordinates' willingness to take on challenges, deny opportunities for challenge, or reject such attempts. (上司や管理職が部下の挑戦意欲を抑えたり、挑戦の機会を与えない、または拒否する行動) & 28.30 (16.09)* & 39.15 (33.41) & 8.15 (41.37) & -1.16 (0.95) & 2.61 (1.78) & 0.75 (2.13) & 88 \\
\addlinespace[0.3em]
\hdashline
\addlinespace[0.3em]
Non-top & Behavior & Challenges in Accepting Opinions of Young Employees (若手意見の受容困難) & Reviews concerning behavioral patterns in which management finds it difficult to accept or tends to disregard the opinions of young employees. (管理層が若手社員の意見を聞き入れにくい、または軽視する行動パターンに関する口コミ。) & -38.37 (46.72) & -51.06 (59.96) & -112.99 (86.20) & 1.88 (2.76) & 2.79 (3.19) & -3.31 (4.43) & 22 \\
\addlinespace[0.3em]
\hdashline
\addlinespace[0.3em]
Non-top & Behavior & Challenges in Developing Young Employees and Negative Behaviors (若手育成の課題と否定的行動) & Reviews concerning issues such as managerial reluctance or obstructive behaviors toward the development of young employees, as well as problems within the developmental environment. (管理職が若手育成に消極的、または妨げる行動をとっていることや育成環境の問題に関する口コミ。) & 12.49 (27.45) & 43.12 (46.69) & 32.42 (57.60) & -1.70 (1.62) & 3.03 (2.47) & 3.01 (2.94) & 49 \\
\addlinespace[0.3em]
\hdashline
\addlinespace[0.3em]
Non-top & Behavior & Challenges in Taking Childcare Leave and Insufficient Work-Life Balance (育児休暇取得の困難と不十分な両立) & This topic addresses the difficulties faced by managerial staff in taking childcare leave, the inability to effectively balance childcare and work responsibilities, and the general reluctance to utilize childcare leave. (管理職が育児休暇を取得しにくい、または育児と仕事の両立ができていない、あるいは育休取得に消極的な状況に関する内容。) & -67.66 (31.20)** & -9.36 (51.83) & 20.66 (78.62) & 2.41 (1.84) & 2.20 (2.76) & 8.43 (4.04)** & 30 \\
\addlinespace[0.3em]
\hdashline
\addlinespace[0.3em]
Non-top & Behavior & Challenges of Work-Life Balance (ワークライフバランスの困難) & Managers face difficulties in securing personal time and taking leave due to heavy workloads and demanding responsibilities. (管理職が多忙や役割の重さによりプライベートの時間をほとんど持てず、休みも取りづらい状況。) & 6.65 (10.51) & -8.89 (18.02) & -14.87 (24.61) & 0.10 (0.62) & -0.93 (0.95) & -0.64 (1.26) & 321 \\
\addlinespace[0.3em]
\hdashline
\addlinespace[0.3em]
Non-top & Behavior & Classroom Management with a Sales-Oriented Focus (営業重視の教室運営) & This topic concerns the management of classrooms by directors or managers who prioritize sales activities and revenue generation in their operational policies. (教室長や室長が営業活動や売上重視の方針で教室運営を行うことに関する内容) & 38.13 (24.13) & 63.96 (37.34)* & 60.09 (38.14) & -2.06 (1.43) & -0.39 (1.99) & -1.07 (1.96) & 17 \\
\addlinespace[0.3em]
\hdashline
\addlinespace[0.3em]
Non-top & Behavior & Close Communication Between Supervisors and Subordinates (上司と部下の密接なコミュニケーション) & A behavioral approach in which supervisors reduce the interpersonal distance with their subordinates and actively engage in open dialogue and information sharing. (上司が部下と距離感を縮め、風通しの良い対話や情報共有を積極的に行う行動様式を指す。) & 3.99 (6.09) & 7.75 (11.35) & 7.84 (14.92) & 1.20 (0.36)*** & 1.36 (0.60)** & 2.00 (0.76)*** & 706 \\
\addlinespace[0.3em]
\hdashline
\addlinespace[0.3em]
Non-top & Behavior & Coercion of Unpaid Overtime Work (サービス残業強要) & Reviews concerning managerial behaviors that compel or tacitly permit subordinates to engage in unpaid or excessive overtime work. (管理職が部下にサービス残業や過剰な残業を強制または黙認する行動に関する口コミ) & 8.30 (7.78) & 3.55 (14.13) & -3.19 (20.59) & -0.13 (0.46) & -0.38 (0.74) & -2.46 (1.06)** & 340 \\
\addlinespace[0.3em]
\hdashline
\addlinespace[0.3em]
Non-top & Behavior & Compliance with Instructions and Strict Conduct of Middle Management (中間管理職の指示遵守と厳格な言動) & This topic illustrates the behavioral patterns of middle managers who faithfully adhere to directives from top management and adopt a strict tone and detailed criticisms toward their subordinates. (経営トップの指示に忠実に従い、部下に対して厳しい口調や細かい指摘を行う中間管理職の行動パターンを示す。) & 16.03 (13.79) & 5.13 (20.95) & -8.19 (30.59) & 0.02 (0.81) & -0.78 (1.11) & -0.54 (1.56) & 135 \\
\addlinespace[0.3em]
\hdashline
\addlinespace[0.3em]
Non-top & Behavior & Comprehensive Support from Supervisors (上司の手厚いサポート) & A behavioral pattern of supervisors characterized by approachability when subordinates face difficulties, and proactive provision of assistance and follow-up support. (部下が困ったときに相談しやすく、積極的に支援やフォローを行う上司の行動パターンを示す。) & -6.61 (6.08) & -2.03 (10.58) & -1.75 (14.39) & 0.43 (0.36) & 0.76 (0.56) & 1.08 (0.74) & 780 \\
\addlinespace[0.3em]
\hdashline
\addlinespace[0.3em]
Non-top & Behavior & Compulsory Participation in Social Drinking Events (強制的な飲み会参加) & A behavioral pattern in which managers compel subordinates to attend drinking parties or client entertainment, thereby exerting pressure and imposing burdens on them. (管理職が部下に飲み会や接待への参加を強制し、プレッシャーや負担をかける行動パターン。) & -2.36 (10.43) & -5.41 (19.50) & 5.59 (29.88) & -0.65 (0.62) & -1.24 (1.04) & -1.44 (1.54) & 154 \\
\addlinespace[0.3em]
\hdashline
\addlinespace[0.3em]
Non-top & Behavior & Consideration and Preferential Treatment of Female Employees (女性社員への配慮と優遇) & The actual practices whereby managers provide consideration, kind treatment, and preferential treatment to female employees. (管理職が女性社員に対して配慮や優しい対応、優遇的な扱いを行う実態。) & -2.89 (11.16) & -4.65 (21.64) & -22.35 (30.02) & 0.38 (0.66) & 0.92 (1.15) & 1.06 (1.54) & 192 \\
\addlinespace[0.3em]
\hdashline
\addlinespace[0.3em]
Non-top & Behavior & Consideration and Thoughtfulness (配慮と気遣い) & Actions whereby a superior attentively considers the physical health, mental well-being, and family circumstances of subordinates, and proactively demonstrates care and concern. (上司が部下の体調やメンタル、家庭事情に対して細やかに配慮し、積極的に気遣う行動) & -5.25 (9.76) & -15.05 (18.62) & 4.31 (25.38) & 1.07 (0.58)* & 1.53 (0.99) & 1.63 (1.30) & 251 \\
\bottomrule
\end{tabular}
\end{adjustbox}
\end{table*}


\begin{table*}[t]
\centering
\scriptsize
\setlength{\tabcolsep}{2pt}
\renewcommand{\arraystretch}{0.92}
\ContinuedFloat
\caption{Relationship between leadership topics and firm outcomes (continued)}
\begin{adjustbox}{max width=\textwidth}
\begin{tabular}{>{\raggedright\arraybackslash}p{0.07\textwidth}>{\raggedright\arraybackslash}p{0.08\textwidth}>{\raggedright\arraybackslash}p{0.18\textwidth}>{\raggedright\arraybackslash}p{0.30\textwidth}>{\centering\arraybackslash}p{0.06\textwidth}>{\centering\arraybackslash}p{0.06\textwidth}>{\centering\arraybackslash}p{0.06\textwidth}>{\centering\arraybackslash}p{0.06\textwidth}>{\centering\arraybackslash}p{0.06\textwidth}>{\centering\arraybackslash}p{0.06\textwidth}>{\centering\arraybackslash}p{0.05\textwidth}}
\toprule
Type. & Char. & Translated Topic Name (Original Topic Name) & Translated Topic Description (Original Topic Description) & \multicolumn{3}{c}{ROA} & \multicolumn{3}{c}{Employees Morale} & N \\
\cmidrule(lr){5-7}\cmidrule(lr){8-10}
 &  &  &  & $t=5$ & $t=10$ & $t=15$ & $t=5$ & $t=10$ & $t=15$ &  \\
\midrule
Non-top & Behavior & Consideration for Women Who Do Not Meet Sales Targets (目標未達の女性への配慮) & Behavioral patterns of considerate responses that avoid pressuring women who have not achieved their sales targets. (営業目標未達の女性に対して詰めることをしない、配慮的な対応の行動パターン) & 106.88 (201.32) & 81.87 (203.11) & 65.07 (207.22) & 10.86 (11.90) & 7.79 (10.80) & 6.36 (10.65) & 4 \\
\addlinespace[0.3em]
\hdashline
\addlinespace[0.3em]
Non-top & Behavior & Denial and Suppression of Transfer Requests (異動希望の否認と抑制) & The actual situation in which supervisors deny transfer requests, make transfer decisions based on personal interests or moods, and engage in negative behaviors such as undermining employees' transfer aspirations. (上司が異動希望を認めず、私利私欲や気分で異動を決めたり、希望を潰すなど否定的な対応を行う実態。) & -4.76 (12.31) & 11.18 (21.50) & 3.96 (28.49) & 0.09 (0.73) & -1.50 (1.14) & -1.35 (1.46) & 192 \\
\addlinespace[0.3em]
\hdashline
\addlinespace[0.3em]
Non-top & Behavior & Deterioration of the Workplace Environment (職場環境の悪化) & Reviews concerning the impact of supervisors' behaviors and attitudes on the workplace atmosphere and overall ease of work. (上司の行動や態度が職場の雰囲気や働きやすさを悪くする影響についての口コミ) & 27.19 (13.66)** & 33.56 (21.01) & 36.99 (32.46) & 0.27 (0.81) & 0.37 (1.11) & 0.24 (1.64) & 163 \\
\addlinespace[0.3em]
\hdashline
\addlinespace[0.3em]
Non-top & Behavior & Differences in Responses and Rigidity (対応の差異と硬直性) & There are significant variations in responses and guidance depending on the supervisor, which in some cases appear rigid and inconsistent. (上司によって対応や指導に大きな差があり、場合によっては硬直的で一貫性がない様子) & 4.38 (17.86) & -16.50 (29.43) & -4.22 (44.33) & -0.89 (1.06) & 0.08 (1.56) & -0.01 (2.28) & 106 \\
\addlinespace[0.3em]
\hdashline
\addlinespace[0.3em]
Non-top & Behavior & Disadvantages Faced by Subordinates Who Are Disfavored by Their Superiors (気に入られない部下の不利益) & Reviews concerning actions in which subordinates who are not favored by their superiors suffer disadvantages such as demotion or transfer. (上司に気に入られない部下が降格や出向など不利益を被る行動に関する口コミ) & 10.57 (22.36) & 10.97 (36.16) & 24.08 (48.13) & -0.14 (1.32) & -0.20 (1.92) & 0.31 (2.47) & 64 \\
\addlinespace[0.3em]
\hdashline
\addlinespace[0.3em]
Non-top & Behavior & Discriminatory Treatment and Marginalization of Female Employees (女性社員への冷遇と軽視) & The reality of negative attitudes exhibited by management toward female employees, including cold treatment, disregard, and withholding of responsibilities. (管理職が女性社員に対して冷たい対応や軽視、責任を与えないなどの否定的な態度を取る実態。) & 4.20 (11.16) & 12.39 (20.32) & -8.18 (25.79) & -0.31 (0.65) & 1.13 (1.02) & 1.05 (1.22) & 196 \\
\addlinespace[0.3em]
\hdashline
\addlinespace[0.3em]
Non-top & Behavior & Early Development of Young Leaders (若手リーダーの早期育成) & Reviews concerning developmental environments where young individuals are entrusted with responsibilities as project or team leaders early on, allowing them to gain experience as leaders. (若いうちからプロジェクトやチームの責任者を任され、リーダーとしての経験を積む育成環境に関する口コミ。) & 41.58 (11.39)*** & 45.76 (15.48)*** & 56.63 (20.37)*** & 2.89 (0.67)*** & 2.80 (0.82)*** & 3.00 (1.04)*** & 344 \\
\addlinespace[0.3em]
\hdashline
\addlinespace[0.3em]
Non-top & Behavior & Efforts in Working Hours Management and Overtime Reduction (労働時間管理と残業削減の取り組み) & Reviews concerning initiatives by leadership to manage working hours, reduce overtime, and promote efficient work practices. (リーダー層が労働時間を管理し、残業を減らそうとする取り組みや効率的な働き方に関する口コミ) & -6.61 (9.31) & -7.62 (17.89) & -19.91 (27.70) & -0.39 (0.55) & 0.72 (0.95) & -0.09 (1.41) & 282 \\
\addlinespace[0.3em]
\hdashline
\addlinespace[0.3em]
Non-top & Behavior & Encouragement of Timely Departure from Work (定時退社推奨行動) & Reviews concerning actions and directives whereby managers proactively leave work on time or early and encourage their subordinates to do the same. (管理職が率先して定時や早めの時間に帰宅し、部下にも早く帰るよう促す行動や指示に関する口コミ。) & -0.29 (10.56) & -7.18 (19.44) & -8.50 (26.11) & -0.58 (0.62) & -1.05 (1.03) & -1.37 (1.34) & 202 \\
\addlinespace[0.3em]
\hdashline
\addlinespace[0.3em]
Non-top & Behavior & Ensuring Work-Life Balance (ワークライフバランスの確保) & The situation in which managerial staff strive to balance professional responsibilities with personal life, achieving the utilization of leave and flexible working arrangements. (管理職が仕事とプライベートの両立を図り、休暇や柔軟な働き方を実現している状況。) & 1.12 (17.53) & -2.38 (28.93) & -19.90 (34.18) & 2.57 (1.04)** & 3.09 (1.54)** & 3.34 (1.75)* & 95 \\
\addlinespace[0.3em]
\hdashline
\addlinespace[0.3em]
Non-top & Behavior & Evaluation Management Based on Discretion (裁量による評価運用) & Cases in which supervisors conduct evaluations based on their own discretion and judgment in the operation of the evaluation system. (上司が評価制度の運用において自分の裁量や判断で評価を行うケース。) & -10.86 (25.39) & -22.83 (84.93) & -37.60 (86.84) & -0.56 (1.50) & 5.94 (4.51) & 4.13 (4.46) & 24 \\
\addlinespace[0.3em]
\hdashline
\addlinespace[0.3em]
Non-top & Behavior & Evaluation Management Based on Standards (基準に基づく評価運用) & Cases in which supervisors conduct evaluations fairly in accordance with evaluation criteria and systems. (上司が評価基準や制度に沿って公平に評価を行うケース。) & -42.52 (54.60) & 53.03 (189.22) & 35.76 (193.03) & 1.43 (3.23) & 3.09 (10.06) & 8.81 (9.92) & 11 \\
\addlinespace[0.3em]
\hdashline
\addlinespace[0.3em]
Non-top & Behavior & Excessive Micromanagement (過剰なマイクロマネジメント) & A behavioral pattern in which middle managers or supervisors provide overly detailed instructions and conduct excessive monitoring, resulting in an undue level of control over subordinates. (中間管理職や上司が細かすぎる指示や監視を行い、部下に対して過度に管理を行う行動パターン) & 45.16 (38.48) & 58.38 (52.34) & 83.62 (70.21) & 3.36 (2.24) & 2.34 (2.70) & -0.65 (3.43) & 56 \\
\addlinespace[0.3em]
\hdashline
\addlinespace[0.3em]
Non-top & Behavior & Excessive Workload Burden (過剰な業務負担) & A situation in which leaders or supervisors assign an excessive amount of work to capable subordinates, resulting in chronic busyness and overtime. (リーダーや上司が能力のある部下に過度に仕事を割り振り、多忙や残業が常態化している状況) & 13.36 (8.21) & 23.93 (14.78) & 29.62 (19.47) & -0.18 (0.49) & -1.00 (0.78) & -1.75 (1.00)* & 459 \\
\addlinespace[0.3em]
\hdashline
\addlinespace[0.3em]
Non-top & Behavior & Excessive Workload and Management Burden among Managers (管理職の過重業務とマネジメント負担) & This topic describes the situation in which managers face intense workloads and diverse responsibilities, struggling to balance operational tasks with managerial duties. (管理職が激務や多様な業務負担に直面し、実務とマネジメントの両立に苦労している状況を表す。) & -3.41 (6.04) & -5.82 (10.67) & -21.26 (15.76) & -0.76 (0.35)** & -1.36 (0.56)** & -2.33 (0.80)*** & 784 \\
\addlinespace[0.3em]
\hdashline
\addlinespace[0.3em]
Non-top & Behavior & Excessive Workload and Unreasonable Directives (過剰な業務負荷と無理な指示) & A behavioral pattern in which management imposes unrealistic deadlines and excessive tasks without considering the actual conditions on the ground. (管理層が現場の状況を考慮せず、無理な納期や過剰な作業を強要する行動パターン。) & -2.95 (10.56) & -17.21 (15.29) & -31.12 (23.28) & -1.06 (0.62)* & -1.32 (0.81) & -2.55 (1.19)** & 315 \\
\addlinespace[0.3em]
\hdashline
\addlinespace[0.3em]
Non-top & Behavior & Exercise of Discretionary Authority (裁量権の行使) & Opinions regarding the scope and manner in which store managers and leaders exercise their discretionary authority. (店長やリーダーが持つ裁量権の範囲や行使の仕方に関する意見) & 5.51 (11.92) & 13.25 (19.59) & 7.12 (26.88) & -1.11 (0.69) & -1.38 (1.04) & -0.34 (1.38) & 171 \\
\addlinespace[0.3em]
\hdashline
\addlinespace[0.3em]
Non-top & Behavior & Exercise of Discretionary Authority (裁量権の行使) & The behavior or tendency of supervisors and managers to actively exercise discretion in decisions regarding salary increases and promotions. (上司や管理職が昇給や昇進の判断において裁量を積極的に行使する行動や傾向。) & 35.74 (18.28)* & 35.01 (29.43) & -7.84 (42.20) & -0.46 (1.08) & -0.32 (1.57) & -1.39 (2.17) & 83 \\
\bottomrule
\end{tabular}
\end{adjustbox}
\end{table*}


\begin{table*}[t]
\centering
\scriptsize
\setlength{\tabcolsep}{2pt}
\renewcommand{\arraystretch}{0.92}
\ContinuedFloat
\caption{Relationship between leadership topics and firm outcomes (continued)}
\begin{adjustbox}{max width=\textwidth}
\begin{tabular}{>{\raggedright\arraybackslash}p{0.07\textwidth}>{\raggedright\arraybackslash}p{0.08\textwidth}>{\raggedright\arraybackslash}p{0.18\textwidth}>{\raggedright\arraybackslash}p{0.30\textwidth}>{\centering\arraybackslash}p{0.06\textwidth}>{\centering\arraybackslash}p{0.06\textwidth}>{\centering\arraybackslash}p{0.06\textwidth}>{\centering\arraybackslash}p{0.06\textwidth}>{\centering\arraybackslash}p{0.06\textwidth}>{\centering\arraybackslash}p{0.06\textwidth}>{\centering\arraybackslash}p{0.05\textwidth}}
\toprule
Type. & Char. & Translated Topic Name (Original Topic Name) & Translated Topic Description (Original Topic Description) & \multicolumn{3}{c}{ROA} & \multicolumn{3}{c}{Employees Morale} & N \\
\cmidrule(lr){5-7}\cmidrule(lr){8-10}
 &  &  &  & $t=5$ & $t=10$ & $t=15$ & $t=5$ & $t=10$ & $t=15$ &  \\
\midrule
Non-top & Behavior & Existence of Harassment Behaviors (ハラスメント行動の存在) & Reviews concerning actual instances and attitudes related to harassment behaviors, such as power harassment and sexual harassment, perpetrated by managerial staff. (管理職によるパワハラやセクハラなどのハラスメント行動が実際に行われている事例や態度に関する口コミ。) & 5.88 (7.62) & 12.77 (12.63) & -0.30 (17.96) & -0.91 (0.45)** & -1.02 (0.67) & -0.07 (0.92) & 363 \\
\addlinespace[0.3em]
\hdashline
\addlinespace[0.3em]
Non-top & Behavior & Fair Evaluation (公正な評価) & A process in which managers assess employees from multiple perspectives in an impartial manner, conducting interviews and evaluations involving multiple evaluators to make appropriate personnel decisions. (管理職が多面的かつ公平に従業員を評価し、面談や複数人での評価を通じて適切な人事判断を行うプロセス。) & 1.53 (7.40) & 29.74 (14.44)** & 34.59 (18.44)* & 0.91 (0.44)** & 0.94 (0.76) & 0.93 (0.95) & 442 \\
\addlinespace[0.3em]
\hdashline
\addlinespace[0.3em]
Non-top & Behavior & Fair Evaluation (公正な評価) & Reviews concerning supervisors' actions in fairly assessing subordinates' achievements and efforts, and appropriately reflecting them. (上司が部下の成果や努力を正当に評価し、適切に反映させる行動に関する口コミ) & 5.79 (6.35) & 12.25 (12.13) & 16.69 (15.51) & -0.46 (0.38) & 0.54 (0.64) & 1.27 (0.80) & 535 \\
\addlinespace[0.3em]
\hdashline
\addlinespace[0.3em]
Non-top & Behavior & Fairness and Bias in Evaluations (評価の公平性と偏り) & Reviews indicate that evaluations and personnel transfers are influenced by supervisors' discretion and preferences, resulting in unfairness and bias. (上長の裁量や好みによって評価や人事異動が左右され、不公平や偏りが生じているという口コミ。) & -6.46 (31.04) & -5.81 (34.68) & 19.17 (63.09) & -0.96 (1.83) & -3.29 (1.84)* & -2.26 (3.24) & 24 \\
\addlinespace[0.3em]
\hdashline
\addlinespace[0.3em]
Non-top & Behavior & Fairness and Proper Implementation of Evaluations (評価の公正性と適正運用) & Reviews concerning the positive aspects of managerial staff conducting evaluations appropriately and implementing salary increases and promotions justly. (管理職が評価を適切に行い、昇給や昇格を正当に実施しているという肯定的な側面に関する口コミ。) & -7.11 (72.17) & -17.26 (72.82) & -24.59 (74.30) & 7.20 (4.27)* & 8.91 (3.87)** & 7.83 (3.81)** & 8 \\
\addlinespace[0.3em]
\hdashline
\addlinespace[0.3em]
Non-top & Behavior & Faithful Top-Down Execution (忠実なトップダウン遂行) & A behavioral pattern in which managers unquestioningly adhere to and faithfully transmit and implement directives and policies from upper management. (管理職が上層部の指示や方針に対して疑問を持たず、忠実に従い伝達・実行する行動様式。) & 20.61 (9.23)** & 36.78 (17.44)** & 51.25 (22.35)** & -0.97 (0.54)* & -1.55 (0.92)* & -2.84 (1.13)** & 321 \\
\addlinespace[0.3em]
\hdashline
\addlinespace[0.3em]
Non-top & Behavior & Favoritism (お気に入り優遇) & Reviews concerning behaviors in which supervisors show preferential treatment toward favored subordinates in matters such as promotion, salary increases, and performance evaluations. (上司が気に入った部下を昇進・昇給・評価などで優遇する行動に関する口コミ) & 3.78 (5.40) & -2.65 (8.93) & 1.64 (12.17) & -0.26 (0.32) & -0.71 (0.47) & -1.33 (0.61)** & 987 \\
\addlinespace[0.3em]
\hdashline
\addlinespace[0.3em]
Non-top & Behavior & Flattering Behavior Toward Superiors (上司へのごますり行動) & A topic illustrating behavioral patterns of ingratiation and appeasement that leaders adopt to gain favor with their superiors and executives. (リーダー層が上司や役員に気に入られるために取るごますりやご機嫌取りの行動パターンを示すトピック。) & 4.29 (7.05) & -9.47 (11.93) & -1.02 (16.78) & -1.55 (0.42)*** & -1.09 (0.63)* & -0.38 (0.85) & 560 \\
\addlinespace[0.3em]
\hdashline
\addlinespace[0.3em]
Non-top & Behavior & Flexibility in Taking Leave (休暇取得の自由度) & Reviews concerning the behaviors and attitudes of store managers and assistant managers in granting employees leave freely and facilitating the approval of requested days off. (店長や副店長が従業員に対して休暇を自由に与えたり、希望休を通しやすくする行動や態度に関する口コミ。) & 14.98 (12.55) & 0.51 (23.73) & -24.70 (30.94) & -0.15 (0.72) & -0.27 (1.26) & -0.70 (1.59) & 111 \\
\addlinespace[0.3em]
\hdashline
\addlinespace[0.3em]
Non-top & Behavior & Flexible Adaptation (柔軟な対応) & The manner in which a supervisor flexibly adjusts their responses and guidance according to the situation and department. (上司が状況や部門に応じて柔軟に対応や指導を変える様子) & 3.67 (41.51) & 32.13 (56.57) & 64.18 (69.96) & 0.44 (2.45) & -0.06 (3.01) & 3.72 (3.60) & 45 \\
\addlinespace[0.3em]
\hdashline
\addlinespace[0.3em]
Non-top & Behavior & Flexible Evaluation and Feedback (柔軟な評価とフィードバック) & Reviews concerning supervisors who explain evaluation criteria and conduct assessments flexibly through interviews and feedback. (上司が評価基準を説明し、面談やフィードバックを通じて柔軟に評価を行う行動に関する口コミ。) & -8.70 (13.58) & -13.59 (24.19) & -26.77 (31.00) & 0.86 (0.80) & 0.85 (1.29) & 0.11 (1.59) & 166 \\
\addlinespace[0.3em]
\hdashline
\addlinespace[0.3em]
Non-top & Behavior & Flexible Management of Paid Leave Requests (有給休暇取得の柔軟な対応) & This illustrates supervisors' flexible approval of subordinates' paid leave applications, including responsiveness to both prior consultations and sudden requests. (上司が部下の有給休暇申請を柔軟に認め、事前相談や急な申請にも対応している様子を示す。) & -0.10 (5.16) & 7.21 (8.30) & 4.58 (11.34) & -0.14 (0.30) & -0.01 (0.44) & -1.13 (0.58)* & 1061 \\
\addlinespace[0.3em]
\hdashline
\addlinespace[0.3em]
Non-top & Behavior & Flexible Top-Down Directives (柔軟なトップダウン指示) & A behavioral approach in which supervisors or leaders issue top-down instructions while allowing flexibility by delegating specific actions to the operational level. (上司やリーダーがトップダウンで指示を出すが、具体的行動は現場に任せるなど柔軟性を持つ行動様式) & -2.06 (11.86) & -9.66 (17.98) & -8.73 (27.25) & -0.35 (0.70) & -1.86 (0.95)* & -2.46 (1.40)* & 272 \\
\addlinespace[0.3em]
\hdashline
\addlinespace[0.3em]
Non-top & Behavior & Flexible Work Schedule Adjustment (柔軟な勤務調整) & Actions taken by supervisors and managers to flexibly adjust workload and shifts, taking into consideration the preferences and personal circumstances of their subordinates. (上司や管理職が部下の希望やプライベートに配慮し、業務量やシフトを柔軟に調整する行動。) & 3.06 (8.88) & -12.88 (14.04) & -32.03 (20.73) & 0.85 (0.52) & 0.87 (0.75) & 2.01 (1.06)* & 333 \\
\addlinespace[0.3em]
\hdashline
\addlinespace[0.3em]
Non-top & Behavior & Formal Directive Transmission (形式的指示伝達) & A situation in which middle managers relay instructions from upper management directly to the operational level without exercising autonomous judgment or demonstrating leadership. (中間管理職が上層部の指示をそのまま現場に伝え、主体的な判断やリーダーシップを発揮しない状況。) & 1.60 (9.83) & 19.51 (20.25) & 18.10 (26.78) & -1.56 (0.58)*** & -2.67 (1.07)** & -1.33 (1.37) & 251 \\
\addlinespace[0.3em]
\hdashline
\addlinespace[0.3em]
Non-top & Behavior & Formalistic or Inadequate Interviews (形骸化・不十分な面談) & Behavioral patterns in which supervisors omit interviews, conduct them merely as a formality, or where the quality of the interviews is perceived to be low. (上司が面談を省略したり、形だけで実施しない、または面談の質が低いと感じられる行動パターン。) & -5.48 (14.46) & 41.05 (30.05) & 33.49 (33.02) & 0.32 (0.85) & 0.78 (1.60) & -0.21 (1.70) & 131 \\
\addlinespace[0.3em]
\hdashline
\addlinespace[0.3em]
Non-top & Behavior & Frequent Changes in Policies and Instructions (方針や指示の頻繁な変更) & Reviews concerning supervisors who frequently alter policies and instructions based on their mood or circumstances. (上司が気分や状況に応じて方針や指示を頻繁に変える行動に関する口コミ。) & 4.33 (9.08) & -16.12 (14.19) & -25.01 (20.35) & -1.60 (0.54)*** & -2.39 (0.75)*** & -2.61 (1.04)** & 362 \\
\addlinespace[0.3em]
\hdashline
\addlinespace[0.3em]
Non-top & Behavior & Frequent Holiday Work Attendance (頻繁な休日出勤) & A situation in which managerial staff frequently work on holidays, including Saturdays, Sundays, and public holidays, thereby continuing business operations. (管理職が土日祝日を含む休日に頻繁に出勤し、業務を継続している状況) & 18.38 (9.73)* & 24.25 (17.22) & -3.18 (24.46) & 0.31 (0.57) & -0.55 (0.90) & -0.15 (1.24) & 364 \\
\bottomrule
\end{tabular}
\end{adjustbox}
\end{table*}


\begin{table*}[t]
\centering
\scriptsize
\setlength{\tabcolsep}{2pt}
\renewcommand{\arraystretch}{0.92}
\ContinuedFloat
\caption{Relationship between leadership topics and firm outcomes (continued)}
\begin{adjustbox}{max width=\textwidth}
\begin{tabular}{>{\raggedright\arraybackslash}p{0.07\textwidth}>{\raggedright\arraybackslash}p{0.08\textwidth}>{\raggedright\arraybackslash}p{0.18\textwidth}>{\raggedright\arraybackslash}p{0.30\textwidth}>{\centering\arraybackslash}p{0.06\textwidth}>{\centering\arraybackslash}p{0.06\textwidth}>{\centering\arraybackslash}p{0.06\textwidth}>{\centering\arraybackslash}p{0.06\textwidth}>{\centering\arraybackslash}p{0.06\textwidth}>{\centering\arraybackslash}p{0.06\textwidth}>{\centering\arraybackslash}p{0.05\textwidth}}
\toprule
Type. & Char. & Translated Topic Name (Original Topic Name) & Translated Topic Description (Original Topic Description) & \multicolumn{3}{c}{ROA} & \multicolumn{3}{c}{Employees Morale} & N \\
\cmidrule(lr){5-7}\cmidrule(lr){8-10}
 &  &  &  & $t=5$ & $t=10$ & $t=15$ & $t=5$ & $t=10$ & $t=15$ &  \\
\midrule
Non-top & Behavior & Frequent Implementation of Working on Holidays (休日出勤の頻繁な実施) & A collection of texts concerning the situation in which managerial staff frequently work on holidays and continue their duties. (管理職が休日に頻繁に出勤し、業務を継続している状況に関する文章群) & 6.53 (6.22) & 17.74 (11.14) & 9.26 (15.35) & -0.92 (0.37)** & -0.97 (0.59) & -0.99 (0.79) & 750 \\
\addlinespace[0.3em]
\hdashline
\addlinespace[0.3em]
Non-top & Behavior & Guidance on Overtime Hours Management (残業時間管理の指導) & Patterns of instructions and supervisory actions undertaken by supervisors to ensure that subordinates' overtime hours remain within prescribed limits. (上司が部下の残業時間を規定内に抑えるために行う指示や監督の行動パターン。) & -7.05 (9.55) & -1.63 (14.58) & -7.30 (20.88) & -0.32 (0.56) & -0.53 (0.77) & -1.29 (1.07) & 347 \\
\addlinespace[0.3em]
\hdashline
\addlinespace[0.3em]
Non-top & Behavior & Harassment Prevention and Mitigation (ハラスメント対策と抑制) & Reviews concerning cases where management actively eliminates power harassment, responds strictly to harassment issues, and thoroughly implements educational measures. (管理職がパワハラ行為を排除し、ハラスメントに対して厳しく対応したり教育を徹底している事例に関する口コミ。) & -1.06 (23.44) & 39.59 (33.61) & 83.47 (44.32)* & -0.53 (1.38) & 2.17 (1.78) & -0.43 (2.28) & 80 \\
\addlinespace[0.3em]
\hdashline
\addlinespace[0.3em]
Non-top & Behavior & Holiday and After-Hours Communication Responsibilities (休日対応と連絡業務) & This pattern describes the behavior in which leaders receive communications during holidays and nighttime, and subsequently attend work or take necessary actions. (リーダーが休日や夜間にも連絡を受け、出勤や対応を行う行動パターンを示す。) & -5.17 (7.11) & 1.62 (11.07) & 12.18 (16.46) & -0.33 (0.42) & -0.57 (0.59) & -1.26 (0.84) & 544 \\
\addlinespace[0.3em]
\hdashline
\addlinespace[0.3em]
Non-top & Behavior & Improvement of the Workplace Environment (職場環境の改善) & Reviews regarding the actions taken by supervisors to enhance the workplace atmosphere and ease of work, as well as their impact. (上司が職場の雰囲気や働きやすさを良くするために取る行動や影響についての口コミ) & -0.67 (13.24) & 15.14 (20.14) & 24.36 (27.97) & -1.16 (0.78) & -2.80 (1.06)*** & -2.79 (1.42)** & 190 \\
\addlinespace[0.3em]
\hdashline
\addlinespace[0.3em]
Non-top & Behavior & Inconsistency in Evaluation Criteria and Discretionary Practices (評価基準の不統一と裁量行使) & Reviews concerning the tendency of supervisors and managers to conduct evaluations based on personal discretion and preferences without standardizing evaluation criteria. (上長や管理職が評価基準を統一せず、個人の裁量や好みで評価を行う傾向に関する口コミ。) & 31.90 (28.17) & 68.54 (40.39)* & 357.41 (97.38)*** & -0.05 (1.65) & 0.03 (2.11) & -1.59 (4.62) & 28 \\
\addlinespace[0.3em]
\hdashline
\addlinespace[0.3em]
Non-top & Behavior & Indifference and Aloof Behavior (無関心と冷淡な対応) & Behaviors in which a superior displays indifference and aloofness toward subordinates, such as showing a lack of consideration, looking at them coldly, or making unpleasant facial expressions. (上司が部下に対して気を遣わず、冷たい目で見たり嫌な顔をするなど無関心・冷淡な態度を示す行動) & -7.09 (32.49) & 20.23 (61.40) & 11.73 (96.64) & 0.20 (1.92) & -1.77 (3.26) & -3.51 (4.97) & 30 \\
\addlinespace[0.3em]
\hdashline
\addlinespace[0.3em]
Non-top & Behavior & Indifferent Managerial Response to Work Duties (無関心な管理職の業務対応) & This describes the behavioral patterns of managers who show indifference toward their subordinates' tasks and circumstances, rarely engaging in consultation or follow-up. (部下の業務や状況に無関心で、相談やフォローをほとんど行わない管理職の行動パターンを示す。) & 1.34 (4.31) & 7.18 (7.21) & -5.17 (10.28) & -0.73 (0.25)*** & -1.00 (0.38)*** & -0.83 (0.53) & 1373 \\
\addlinespace[0.3em]
\hdashline
\addlinespace[0.3em]
Non-top & Behavior & Initiatives to Reduce Overtime Work (残業抑制の取り組み) & Management and supervisory actions undertaken by managers to decrease their subordinates' overtime hours and to avoid unnecessary overtime work. (管理職が部下の残業時間を減らし、無駄な残業を避けるために行う管理や指導の行動) & 2.61 (11.91) & -10.14 (18.00) & -33.07 (25.27) & 1.68 (0.70)** & 1.46 (0.95) & 1.65 (1.28) & 303 \\
\addlinespace[0.3em]
\hdashline
\addlinespace[0.3em]
Non-top & Behavior & Instructions and Responses Regarding Work on Holidays and Overtime (休日・時間外勤務の指示と対応) & Reviews concerning the behavioral patterns in which supervisors contact subordinates and instruct them to work during holidays or outside regular working hours, and the corresponding responses. (上司が休日や時間外に部下へ連絡や出勤を指示し、対応を求める行動パターンに関する口コミ。) & -17.82 (15.35) & -41.34 (24.62)* & -91.60 (34.25)*** & -1.16 (0.91) & -0.93 (1.31) & -3.13 (1.76)* & 148 \\
\addlinespace[0.3em]
\hdashline
\addlinespace[0.3em]
Non-top & Behavior & Insufficient Development and Restriction of Discretion (育成不足と裁量制限) & A situation in which young employees are not entrusted with responsible tasks, resulting in inadequate development and limited delegation of discretion. (若手社員に責任ある仕事を任せず、育成や裁量付与が不足している状況) & 3.11 (21.26) & -15.03 (38.54) & -77.06 (57.71) & 1.40 (1.26) & 2.67 (2.04) & -2.12 (2.95) & 57 \\
\addlinespace[0.3em]
\hdashline
\addlinespace[0.3em]
Non-top & Behavior & Insufficient Development of Subordinates by Supervisors (上司の部下育成不足) & This refers to situations where supervisors fail to adequately provide education and career guidance to their subordinates, thereby lacking in offering opportunities for their growth. (上司が部下の教育やキャリア指導を十分に行わず、成長機会を提供できていない状況を示す。) & -10.39 (6.16)* & -24.47 (11.49)** & -42.57 (15.34)*** & -0.73 (0.36)** & 0.12 (0.61) & 0.20 (0.79) & 754 \\
\addlinespace[0.3em]
\hdashline
\addlinespace[0.3em]
Non-top & Behavior & Insufficient Leadership Skills Among Managers (管理職の指導力不足) & This refers to a situation where leaders inadequately provide instructions and follow-up to their subordinates, resulting in insufficient operational management and support for employee development. (リーダー層が部下への指示やフォローを十分に行わず、業務管理や成長支援が不十分な状況を示す。) & -3.66 (4.44) & -8.90 (7.69) & -21.72 (10.89)** & -0.96 (0.26)*** & -0.84 (0.41)** & -0.03 (0.56) & 1458 \\
\addlinespace[0.3em]
\hdashline
\addlinespace[0.3em]
Non-top & Behavior & Insufficient Supervision and Lack of Developmental Environment (指導不足と育成環境の欠如) & A situation in which supervisors are unable to allocate sufficient time for instructing and developing subordinates, resulting in an inadequate educational environment. (上司が部下への指導や育成に時間を割けず、教育環境が不十分である状況) & -17.85 (13.11) & -41.97 (22.58)* & -47.64 (31.99) & -1.06 (0.77) & -2.40 (1.20)** & -2.75 (1.63)* & 180 \\
\addlinespace[0.3em]
\hdashline
\addlinespace[0.3em]
Non-top & Behavior & Insufficient and Restrictive Delegation of Authority (権限委譲の不十分・制限的対応) & Behavioral patterns in which supervisors do not assign responsible tasks to subordinates, refrain from entrusting them, or impose limitations on their discretion. (上司が部下に責任ある仕事を与えない、任せない、または裁量を制限する行動パターン) & 20.21 (21.70) & 33.67 (26.76) & 50.96 (40.30) & -0.36 (1.25) & 0.39 (1.42) & 1.30 (2.07) & 94 \\
\addlinespace[0.3em]
\hdashline
\addlinespace[0.3em]
Non-top & Behavior & Issues Related to Discretionary Authority (裁量権の問題点) & A collection of texts highlighting concerns that salary increases and promotion decisions made at the discretion of supervisors can be unfair and arbitrary. (上司の裁量による昇給昇進決定が不公平や独断的であることを指摘する文章群。) & 15.65 (39.35) & -54.79 (90.89) & -67.81 (92.75) & 4.08 (2.33)* & 1.81 (4.83) & 0.18 (4.77) & 18 \\
\addlinespace[0.3em]
\hdashline
\addlinespace[0.3em]
Non-top & Behavior & Issues in Shift Management (シフト管理の問題点) & Problematic behaviors in shift scheduling, such as supervisors or store managers monopolizing shift assignments and failing to consider employees' personal circumstances. (上司や店長がシフトを私物化したり、従業員のプライベートを考慮しないなど、勤務調整に問題がある行動。) & -14.99 (19.45) & 0.70 (31.00) & -17.66 (39.77) & -2.71 (1.15)** & -4.58 (1.65)*** & -4.85 (2.04)** & 67 \\
\addlinespace[0.3em]
\hdashline
\addlinespace[0.3em]
Non-top & Behavior & Lack of Guidance and Negative Attitudes (指導の不足や否定的態度) & A behavioral pattern in which senior employees or supervisors fail to provide adequate instruction or support to junior staff or newcomers, exhibiting reluctance to teach and causing confusion or unreasonable demands. (先輩社員や上司が後輩や新人に対して指導やサポートを十分に行わず、教えようとしない、困惑や無茶振りがある行動パターン。) & -16.14 (31.01) & 23.29 (38.91) & 15.22 (57.89) & -2.01 (1.83) & -4.86 (2.07)** & -2.91 (2.97) & 41 \\
\bottomrule
\end{tabular}
\end{adjustbox}
\end{table*}


\begin{table*}[t]
\centering
\scriptsize
\setlength{\tabcolsep}{2pt}
\renewcommand{\arraystretch}{0.92}
\ContinuedFloat
\caption{Relationship between leadership topics and firm outcomes (continued)}
\begin{adjustbox}{max width=\textwidth}
\begin{tabular}{>{\raggedright\arraybackslash}p{0.07\textwidth}>{\raggedright\arraybackslash}p{0.08\textwidth}>{\raggedright\arraybackslash}p{0.18\textwidth}>{\raggedright\arraybackslash}p{0.30\textwidth}>{\centering\arraybackslash}p{0.06\textwidth}>{\centering\arraybackslash}p{0.06\textwidth}>{\centering\arraybackslash}p{0.06\textwidth}>{\centering\arraybackslash}p{0.06\textwidth}>{\centering\arraybackslash}p{0.06\textwidth}>{\centering\arraybackslash}p{0.06\textwidth}>{\centering\arraybackslash}p{0.05\textwidth}}
\toprule
Type. & Char. & Translated Topic Name (Original Topic Name) & Translated Topic Description (Original Topic Description) & \multicolumn{3}{c}{ROA} & \multicolumn{3}{c}{Employees Morale} & N \\
\cmidrule(lr){5-7}\cmidrule(lr){8-10}
 &  &  &  & $t=5$ & $t=10$ & $t=15$ & $t=5$ & $t=10$ & $t=15$ &  \\
\midrule
Non-top & Behavior & Leadership Attitudes in Protecting Junior Employees (若手を守るリーダーの姿勢) & Behaviors and attitudes of leaders who prioritize preventing failures and safeguarding junior employees. (若手社員の失敗を防ぎ、守ることを重視するリーダーの行動や態度。) & 6.44 (36.14) & 6.58 (41.29) & 52.40 (68.01) & 1.82 (2.14) & 0.96 (2.20) & 2.80 (3.50) & 32 \\
\addlinespace[0.3em]
\hdashline
\addlinespace[0.3em]
Non-top & Behavior & Leniency and Favoritism in Evaluations (評価の甘さと優遇) & Reviews concerning supervisors exhibiting lenient evaluations and preferential treatment toward specific subordinates. (上司が特定の部下に対して甘い評価や優遇を行う行動に関する口コミ。) & -7.52 (16.86) & 3.95 (35.18) & -2.66 (49.64) & 0.14 (1.00) & -0.83 (1.87) & 0.45 (2.55) & 72 \\
\addlinespace[0.3em]
\hdashline
\addlinespace[0.3em]
Non-top & Behavior & Long Working Hours Among Managerial Staff (管理職の長時間労働) & This refers to the situation in which managerial staff are frequently required to work overtime late at night and on holidays, resulting in prolonged working hours. (管理職が深夜や休日まで頻繁に残業し、長時間勤務を強いられている状況を示す。) & 20.62 (11.33)* & 43.05 (16.21)*** & 39.33 (21.63)* & -1.52 (0.67)** & -2.05 (0.86)** & -1.39 (1.11) & 302 \\
\addlinespace[0.3em]
\hdashline
\addlinespace[0.3em]
Non-top & Behavior & Long Working Hours Among Managerial Staff (管理職の長時間労働) & A situation in which managerial personnel routinely engage in extended overtime and after-hours work, with such practices becoming the standard basis for evaluation and work style. (管理職層が長時間の残業や時間外労働を常態的に行い、それが評価や働き方の前提となっている状況。) & -0.25 (5.19) & 3.55 (8.97) & 8.40 (12.20) & -0.60 (0.31)* & -1.16 (0.48)** & -2.27 (0.62)*** & 1185 \\
\addlinespace[0.3em]
\hdashline
\addlinespace[0.3em]
Non-top & Behavior & Long Working Hours and Holiday Work Among Managers (役職者の長時間労働と休日出勤) & Employee testimonials illustrating the reality of managers being compelled to engage in long working hours, including late-night overtime and holiday work. (役職者が深夜残業や休日出勤を含む長時間労働を強いられる実態を示す口コミ。) & 14.88 (8.13)* & 28.52 (13.24)** & 26.86 (17.93) & -0.37 (0.47) & -0.44 (0.70) & -2.90 (0.92)*** & 627 \\
\addlinespace[0.3em]
\hdashline
\addlinespace[0.3em]
Non-top & Behavior & Low Ratings and Unfairness (評価の低さと不公平さ) & Reviews concerning supervisors who fail to fairly assess the achievements and efforts of their subordinates, resulting in low evaluations and disadvantages. (上司が部下の成果や努力を正当に評価せず、低い評価や不利益を与える行動に関する口コミ。) & 13.87 (8.03)* & 26.61 (12.97)** & 16.11 (17.39) & -0.15 (0.47) & -0.76 (0.69) & -1.27 (0.89) & 422 \\
\addlinespace[0.3em]
\hdashline
\addlinespace[0.3em]
Non-top & Behavior & Management Dysfunction Among Supervisors (管理職のマネジメント不全) & A situation in which supervisors fail to adequately manage operations and guide subordinates, resulting in a lack of presence and insufficient delegation of authority. (管理職が業務管理や部下指導を適切に行えず、存在感や権限委譲が不足している状況を指す。) & -7.61 (3.87)** & -11.34 (6.43)* & -24.92 (8.62)*** & -1.09 (0.23)*** & -1.44 (0.34)*** & -1.78 (0.44)*** & 2008 \\
\addlinespace[0.3em]
\hdashline
\addlinespace[0.3em]
Non-top & Behavior & Management Dysfunction Among Supervisors (管理職のマネジメント不全) & This term refers to situations in which supervisors fail to adequately understand tasks, coordinate activities, and support subordinates, resulting in ineffective management. (管理職が業務把握や調整、部下支援を十分に行わず、適切なマネジメントができていない状況を指す。) & 4.25 (10.46) & 17.15 (15.62) & 13.58 (20.22) & 0.15 (0.62) & 0.13 (0.83) & -0.01 (1.04) & 331 \\
\addlinespace[0.3em]
\hdashline
\addlinespace[0.3em]
Non-top & Behavior & Management and Leadership by Passive and Authoritarian Store Managers (消極的・独裁的な店長の運営と指導) & A collection of texts illustrating negative aspects such as store managers being passive in subordinate development or exhibiting authoritarian attitudes. (店長が部下の育成に消極的であったり、独裁的な態度をとるなど、否定的な側面を示す文章群。) & -4.62 (6.18) & 11.50 (10.55) & 11.16 (12.20) & -0.09 (0.36) & 0.34 (0.56) & 1.24 (0.62)** & 416 \\
\addlinespace[0.3em]
\hdashline
\addlinespace[0.3em]
Non-top & Behavior & Management of Paid Leave Utilization (有給休暇取得管理) & Reviews concerning supervisors' management of subordinates' paid leave usage and their actions or guidance aimed at promoting systematic and planned utilization. (上司が部下の有給休暇取得状況を管理し、計画的な消化を促す行動や指導に関する口コミ。) & -6.93 (6.12) & -5.79 (8.67) & -9.55 (11.53) & -0.16 (0.36) & -0.76 (0.46)* & -1.00 (0.59)* & 904 \\
\addlinespace[0.3em]
\hdashline
\addlinespace[0.3em]
Non-top & Behavior & Management-Focused Managerial Roles (マネジメント専念型管理職) & Reviews concerning the work style in which managers disengage from operational tasks to concentrate exclusively on management duties, delegating practical work to their subordinates. (管理職が実務から離れ、マネジメント業務に専念し実務を部下に任せる働き方に関する口コミ) & 21.89 (19.99) & 51.65 (27.75)* & 85.50 (39.83)** & 1.68 (1.18) & 0.08 (1.47) & -1.19 (2.05) & 109 \\
\addlinespace[0.3em]
\hdashline
\addlinespace[0.3em]
Non-top & Behavior & Managerial Support for Encouraging Timely Departure (帰宅促進の上司対応) & Employee feedback regarding supervisors' proactive encouragement and approval of early or on-time departure, reflecting attitudes and behaviors that promote leaving work promptly. (上司が部下の早めの退社や定時退社を積極的に推奨・許可し、帰宅を促す態度や行動に関する口コミ) & -1.45 (9.28) & 4.90 (16.97) & -5.34 (22.89) & 0.34 (0.55) & 0.07 (0.90) & -0.45 (1.18) & 284 \\
\addlinespace[0.3em]
\hdashline
\addlinespace[0.3em]
Non-top & Behavior & Managerial Task Coordination and Interpersonal Behavior (マネージャーの業務調整と対人行動) & This topic illustrates the behavioral patterns of managers as they oversee and coordinate tasks while actively engaging with team members and striving to create a conducive work environment. (マネージャーが業務の管理や調整を行いながら、メンバーとの関わり方や働きやすい環境づくりに努める行動パターンを示す。) & 14.69 (13.56) & 62.11 (23.01)*** & 89.76 (30.25)*** & 2.63 (0.80)*** & 1.68 (1.22) & 3.74 (1.55)** & 206 \\
\addlinespace[0.3em]
\hdashline
\addlinespace[0.3em]
Non-top & Behavior & Managers Who Do Not Work on Holidays (休日出勤しない管理職) & While most leaders at the section manager level and above generally do not work on weekends, there are exceptional circumstances in which they may be required to do so. (課長以上のリーダーの多くは基本的に土日に出勤しないが、例外的に出勤する場合がある状況) & 27.84 (32.39) & 95.63 (52.95)* & 108.11 (65.22)* & 0.69 (1.91) & -0.89 (2.82) & -4.66 (3.35) & 49 \\
\addlinespace[0.3em]
\hdashline
\addlinespace[0.3em]
Non-top & Behavior & Mandatory Holiday Duty (強制的な休日対応) & A behavioral pattern in which managers compel employees to respond during holidays or late at night, enforcing attendance or communication. (管理職が休日や深夜に強制的に対応を求め、出勤や連絡を強要する行動パターン。) & 9.63 (19.28) & 53.91 (26.41)** & 68.44 (37.61)* & -0.96 (1.13) & -1.02 (1.39) & 1.43 (1.90) & 106 \\
\addlinespace[0.3em]
\hdashline
\addlinespace[0.3em]
Non-top & Behavior & Meeting Management and Participant Burden (会議運営と参加負担) & The participation of managers in numerous meetings, focusing on the perceived burden related to the frequency and duration of meetings, as well as the actual practices of meeting management. (管理職が多くの会議に参加し、会議の頻度や時間に対する負担感や運営の実態) & -12.67 (15.97) & -7.00 (24.85) & 38.57 (38.13) & 1.42 (0.94) & 1.19 (1.32) & 0.56 (1.96) & 154 \\
\addlinespace[0.3em]
\hdashline
\addlinespace[0.3em]
Non-top & Behavior & Meeting Preparation and Document Creation (会議準備と資料作成) & The reality that managers spend a significant amount of time preparing materials and making arrangements for meetings. (管理職が会議のために資料作成や準備に多くの時間を費やす実態) & 29.51 (32.92) & -9.01 (40.82) & 44.94 (60.02) & -1.04 (1.95) & -2.64 (2.17) & -2.27 (3.08) & 32 \\
\addlinespace[0.3em]
\hdashline
\addlinespace[0.3em]
Non-top & Behavior & Merit-Based Evaluation in Promotions (昇進における実力評価) & The actual practice of promoting leaders based on their abilities, achievements, and work outcomes. (リーダーの昇進が実力や成果、仕事の結果に基づいて行われる実態) & 4.79 (10.48) & 12.60 (18.98) & 45.48 (26.49)* & -1.07 (0.61)* & 0.94 (1.01) & 0.56 (1.36) & 281 \\
\bottomrule
\end{tabular}
\end{adjustbox}
\end{table*}


\begin{table*}[t]
\centering
\scriptsize
\setlength{\tabcolsep}{2pt}
\renewcommand{\arraystretch}{0.92}
\ContinuedFloat
\caption{Relationship between leadership topics and firm outcomes (continued)}
\begin{adjustbox}{max width=\textwidth}
\begin{tabular}{>{\raggedright\arraybackslash}p{0.07\textwidth}>{\raggedright\arraybackslash}p{0.08\textwidth}>{\raggedright\arraybackslash}p{0.18\textwidth}>{\raggedright\arraybackslash}p{0.30\textwidth}>{\centering\arraybackslash}p{0.06\textwidth}>{\centering\arraybackslash}p{0.06\textwidth}>{\centering\arraybackslash}p{0.06\textwidth}>{\centering\arraybackslash}p{0.06\textwidth}>{\centering\arraybackslash}p{0.06\textwidth}>{\centering\arraybackslash}p{0.06\textwidth}>{\centering\arraybackslash}p{0.05\textwidth}}
\toprule
Type. & Char. & Translated Topic Name (Original Topic Name) & Translated Topic Description (Original Topic Description) & \multicolumn{3}{c}{ROA} & \multicolumn{3}{c}{Employees Morale} & N \\
\cmidrule(lr){5-7}\cmidrule(lr){8-10}
 &  &  &  & $t=5$ & $t=10$ & $t=15$ & $t=5$ & $t=10$ & $t=15$ &  \\
\midrule
Non-top & Behavior & Negative and Restrictive Attitudes Toward Working on Holidays (休日出勤の否定的・抑制的態度) & A collection of texts concerning managerial attitudes that discourage working on holidays or refrain from working to the extent of sacrificing personal days off. (管理職が休日出勤を控えたり、休日を潰してまで働かないという否定的または抑制的な態度に関する文章群) & -6.80 (24.40) & -16.18 (30.02) & -23.39 (57.00) & -1.95 (1.43) & -3.24 (1.57)** & -6.47 (2.79)** & 45 \\
\addlinespace[0.3em]
\hdashline
\addlinespace[0.3em]
Non-top & Behavior & Negative and Uncooperative Responses (否定的・非協力的対応) & Behaviors and statements in which supervisors demonstrate a lack of understanding toward the taking of childcare leave or maternity leave, exhibiting negative attitudes or inappropriate conduct. (上司が育児休業や産休取得に対して理解が乏しく、否定的な態度や不適切な言動を示す行動や言動。) & -34.45 (14.39)** & -3.47 (24.94) & -14.54 (32.42) & -0.88 (0.85) & 0.36 (1.31) & 0.06 (1.63) & 105 \\
\addlinespace[0.3em]
\hdashline
\addlinespace[0.3em]
Non-top & Behavior & Neglect and Misconception of Management Duties (マネジメント業務の軽視・誤認) & A collection of texts that exhibit a negative or passive stance, such as misconstruing management duties as mere coordination or workload management, demonstrating a lack of management awareness, or being overwhelmed by operational tasks. (マネジメント業務を単なる調整や繁閑管理と誤認したり、マネジメント意識が希薄であったり、業務に忙殺されているといった否定的・消極的なスタンスを示す文章群。) & 117.20 (37.41)*** & 127.84 (59.30)** & 122.03 (70.79)* & 2.61 (2.20) & -1.95 (3.10) & -4.52 (3.56) & 32 \\
\addlinespace[0.3em]
\hdashline
\addlinespace[0.3em]
Non-top & Behavior & Negligence of Duties Among Middle Management (中間管理職の職務怠慢) & Reviews describing instances where middle managers, such as section chiefs and department heads, fail to fulfill their responsibilities and exhibit a passive attitude toward managing subordinates and executing tasks. (課長や部長などの中間管理職が責任を果たさず、部下の管理や業務遂行に消極的な様子を示す口コミ。) & -40.26 (20.92)* & -20.52 (29.02) & -31.10 (37.83) & -2.11 (1.24)* & -0.53 (1.54) & 0.82 (1.94) & 144 \\
\addlinespace[0.3em]
\hdashline
\addlinespace[0.3em]
Non-top & Behavior & Non-utilization and Difficulty in Taking Paid Leave (有給取得しない・取得困難) & Situations and behaviors in which managers do not take paid leave or are unable to do so due to reasons such as excessive overtime and heavy workload. (管理職が残業過多や業務負荷などの理由で有給休暇を取得しない、または取得できていない状況や行動。) & 27.25 (14.49)* & 38.19 (25.01) & 71.37 (40.96)* & -0.43 (0.86) & -0.31 (1.33) & -1.82 (2.10) & 142 \\
\addlinespace[0.3em]
\hdashline
\addlinespace[0.3em]
Non-top & Behavior & Normalization of Extended Overtime Work (長時間残業の常態化) & A situation in which managerial staff are compelled to work long hours during busy periods and on a regular basis, resulting in a chronic accumulation of excessive overtime. (管理職が繁忙期や日常的に長時間残業を強いられ、慢性的に多くの残業時間をこなしている状況) & 6.63 (6.50) & 12.37 (10.62) & 1.54 (14.51) & -0.73 (0.38)* & -1.35 (0.56)** & -2.60 (0.74)*** & 835 \\
\addlinespace[0.3em]
\hdashline
\addlinespace[0.3em]
Non-top & Behavior & Normalization of Extended Working Hours (長時間勤務の常態化) & A situation in which store managers are compelled to work long hours, including overtime and holiday shifts, leading to significant exhaustion. (店舗管理職が長時間勤務や残業、休日出勤を強いられ、疲弊している状況) & 10.86 (5.66)* & 9.78 (10.01) & 12.65 (11.81) & -0.10 (0.33) & 0.51 (0.53) & 0.45 (0.60) & 442 \\
\addlinespace[0.3em]
\hdashline
\addlinespace[0.3em]
Non-top & Behavior & Opacity in Evaluation and Discretion (評価の不透明性と裁量) & Reviews concerning the issues arising from managers obscuring evaluation criteria and conducting assessments based on discretion and subjective judgment. (管理職が評価基準を曖昧にし、裁量や主観で評価を行う問題点に関する口コミ。) & 9.56 (9.14) & 16.37 (15.75) & -6.95 (20.80) & -0.24 (0.54) & -0.95 (0.83) & -1.72 (1.05) & 409 \\
\addlinespace[0.3em]
\hdashline
\addlinespace[0.3em]
Non-top & Behavior & Overtime Management and Adjustment (残業管理と調整) & Supervisors manage overtime hours and, when necessary, adjust workloads or approve requests, thereby authorizing and supporting overtime work. (上長が残業時間を管理し、必要に応じて業務量調整や申請承認を行い、残業を認めたりフォローする行動。) & -3.74 (14.97) & -33.33 (26.33) & -6.90 (45.82) & -0.52 (0.88) & -2.66 (1.40)* & -0.39 (2.36) & 92 \\
\addlinespace[0.3em]
\hdashline
\addlinespace[0.3em]
Non-top & Behavior & Overtime Reduction Guidance (残業抑制指導) & Supervisors strictly manage and instruct employees to reduce overtime, implementing actions and policies that prohibit or limit overtime work. (上長が残業を減らすために厳しく管理・指導し、残業を許さないまたは制限する行動や方針。) & 2.05 (26.59) & -18.72 (41.04) & -85.99 (84.93) & 0.94 (1.57) & -2.04 (2.18) & -2.87 (4.37) & 38 \\
\addlinespace[0.3em]
\hdashline
\addlinespace[0.3em]
Non-top & Behavior & Parental Leave Acquisition and Continuity of Professional Duties (育児休暇取得と職務継続) & This topic addresses the realities of managerial employees who continue to perform their professional roles actively by taking maternity and parental leave while utilizing reduced working hours and returning to work. (管理職が産休・育休を取得しながら時短勤務や復職を通じて職務を継続し、活躍している実態に関する内容。) & 17.33 (16.15) & 50.20 (27.01)* & 73.06 (34.23)** & 2.93 (0.95)*** & 4.15 (1.41)*** & 3.26 (1.72)* & 175 \\
\addlinespace[0.3em]
\hdashline
\addlinespace[0.3em]
Non-top & Behavior & Passive Attitude Toward Accepting Opinions (意見受容の消極的態度) & A behavioral pattern in which leaders do not fully accept their subordinates' opinions and respond in a passive or limited manner. (リーダーが部下の意見を十分に受け入れず、消極的または限定的に対応する行動パターン) & -13.39 (41.06) & 15.12 (65.39) & 33.75 (129.82) & -0.44 (2.43) & -2.33 (3.48) & -8.03 (6.67) & 16 \\
\addlinespace[0.3em]
\hdashline
\addlinespace[0.3em]
Non-top & Behavior & Passive Career Support (消極的なキャリア支援) & Reviews describing supervisors who render career support perfunctory, fail to conduct interviews, or exhibit passive and negative behaviors toward the career development of their subordinates. (上司がキャリア支援を形骸化したり、面談を行わなかったり、部下のキャリア開発に消極的・否定的な行動をとる口コミ。) & -1.84 (14.93) & -28.29 (26.43) & -10.40 (42.98) & -0.08 (0.88) & 0.32 (1.41) & 1.94 (2.21) & 113 \\
\addlinespace[0.3em]
\hdashline
\addlinespace[0.3em]
Non-top & Behavior & Passive Communication (消極的なコミュニケーション) & A situation in which managers do not engage in sufficient communication with employees, resulting in a distant relationship and a lack of mutual understanding. (管理職が従業員と十分にコミュニケーションを取らず、距離が遠かったり意思疎通が取れていない様子。) & 42.70 (28.89) & 1.54 (64.85) & -14.15 (66.24) & 2.99 (1.70)* & -0.11 (3.36) & -1.95 (3.40) & 56 \\
\addlinespace[0.3em]
\hdashline
\addlinespace[0.3em]
Non-top & Behavior & Passive Decision-Making (消極的な意思決定) & The tendency of middle managers to delay decisions or avoid making independent judgments, instead requiring confirmation from higher management. (中間管理職が判断を遅らせたり、自己判断を避けて上層部の確認を必要とする様子) & -6.86 (17.70) & 6.65 (23.50) & 16.91 (32.00) & -0.38 (1.04) & -1.57 (1.25) & 0.03 (1.64) & 158 \\
\addlinespace[0.3em]
\hdashline
\addlinespace[0.3em]
Non-top & Behavior & Passive Store Management (消極的な店舗管理) & Patterns of passive management behavior characterized by area managers neglecting store visits and guidance, thereby leaving on-site issues unaddressed. (エリアマネージャーが店舗訪問や指導を怠り、現場の問題を放置するなど消極的な管理行動のパターン) & -5.28 (19.69) & 20.18 (34.52) & -41.58 (53.00) & -0.56 (1.15) & 2.30 (1.76) & -0.51 (2.50) & 44 \\
\addlinespace[0.3em]
\hdashline
\addlinespace[0.3em]
Non-top & Behavior & Passive and Inadequate Management Behaviors (消極的・不十分なマネジメント行動) & Behavioral patterns of supervisors demonstrating passive and insufficient management, such as providing inadequate guidance and support to subordinates or delegating tasks irresponsibly. (部下への指導や支援が不足していたり、業務を丸投げするなど消極的・不十分なマネジメントを示す上長の行動パターン。) & 10.31 (6.56) & 14.48 (11.80) & 19.23 (15.08) & -0.64 (0.39) & -0.23 (0.63) & -1.16 (0.77) & 684 \\
\addlinespace[0.3em]
\hdashline
\addlinespace[0.3em]
Non-top & Behavior & Passive and Negative Impacts of Externally Hired Managers (外部出身管理職の消極的・否定的影響) & Situations in which managers recruited from outside the organization exhibit passive attitudes in the workplace and exert detrimental effects on the organization. (外部から採用された管理職が企業の現場で消極的な態度をとったり、組織に悪影響を及ぼしている状況。) & 34.64 (59.61) & 68.98 (74.59) & 114.08 (142.62) & 0.00 (3.52) & -1.23 (3.97) & -5.68 (7.33) & 16 \\
\bottomrule
\end{tabular}
\end{adjustbox}
\end{table*}


\begin{table*}[t]
\centering
\scriptsize
\setlength{\tabcolsep}{2pt}
\renewcommand{\arraystretch}{0.92}
\ContinuedFloat
\caption{Relationship between leadership topics and firm outcomes (continued)}
\begin{adjustbox}{max width=\textwidth}
\begin{tabular}{>{\raggedright\arraybackslash}p{0.07\textwidth}>{\raggedright\arraybackslash}p{0.08\textwidth}>{\raggedright\arraybackslash}p{0.18\textwidth}>{\raggedright\arraybackslash}p{0.30\textwidth}>{\centering\arraybackslash}p{0.06\textwidth}>{\centering\arraybackslash}p{0.06\textwidth}>{\centering\arraybackslash}p{0.06\textwidth}>{\centering\arraybackslash}p{0.06\textwidth}>{\centering\arraybackslash}p{0.06\textwidth}>{\centering\arraybackslash}p{0.06\textwidth}>{\centering\arraybackslash}p{0.05\textwidth}}
\toprule
Type. & Char. & Translated Topic Name (Original Topic Name) & Translated Topic Description (Original Topic Description) & \multicolumn{3}{c}{ROA} & \multicolumn{3}{c}{Employees Morale} & N \\
\cmidrule(lr){5-7}\cmidrule(lr){8-10}
 &  &  &  & $t=5$ & $t=10$ & $t=15$ & $t=5$ & $t=10$ & $t=15$ &  \\
\midrule
Non-top & Behavior & Passive and Problem-Avoidant On-Site Management (消極的・問題回避的な現場管理) & Behaviors exhibited by site supervisors or leaders who avoid addressing troubles and management responsibilities, or who display harsh attitudes and dissatisfaction. (現場責任者やリーダーがトラブル対応や管理を避けたり、厳しい態度や不満を示す行動) & -7.44 (36.93) & -75.88 (51.40) & -90.88 (53.85)* & 1.74 (2.18) & 3.06 (2.73) & 2.83 (2.77) & 47 \\
\addlinespace[0.3em]
\hdashline
\addlinespace[0.3em]
Non-top & Behavior & Performance Evaluations Influenced by Supervisors' Personal Preferences (上司の好き嫌いによる人事評価) & Reviews concerning the tendency of supervisors to affect subordinates' evaluations based on personal likes and emotions. (上司が個人的な好みや感情で部下の評価を左右する傾向についての口コミ。) & 3.56 (8.74) & 7.84 (13.44) & 2.23 (18.81) & -0.66 (0.52) & -1.11 (0.71) & -2.97 (0.96)*** & 404 \\
\addlinespace[0.3em]
\hdashline
\addlinespace[0.3em]
Non-top & Behavior & Personnel Evaluation Authority of Direct Supervisors (直属上司の人事評価権限) & Refers to the role of direct supervisors and managerial personnel in ultimately determining individual personnel evaluations and leading the evaluation process. (直属の上司や管理職が個人の人事評価を最終的に決定し、評価プロセスを主導する役割を指す。) & -0.35 (6.79) & 3.48 (12.55) & -7.86 (16.51) & -0.42 (0.40) & -0.22 (0.66) & -0.68 (0.84) & 508 \\
\addlinespace[0.3em]
\hdashline
\addlinespace[0.3em]
Non-top & Behavior & Political and Personal Factors in Promotion (昇進における政治的・個人的要因) & The reality that leaders' promotions are influenced by political and personal factors such as favoritism, factionalism, and personal preferences. (リーダーの昇進が気に入られ方や派閥、好き嫌いなどの政治的・個人的要因に左右される実態) & 5.54 (5.63) & 17.09 (9.14)* & 18.92 (12.52) & -0.35 (0.33) & -0.55 (0.48) & -1.35 (0.64)** & 1009 \\
\addlinespace[0.3em]
\hdashline
\addlinespace[0.3em]
Non-top & Behavior & Practice-Oriented Playing Manager (実務重視のプレイングマネージャー) & Reviews concerning a working style in which managers actively engage in practical tasks and player duties, participating directly in on-site operations and sales activities. (管理職が積極的に実務やプレイヤー業務を行い、現場での作業や営業活動に従事している働き方に関する口コミ) & 7.76 (7.93) & 22.28 (15.51) & 61.82 (24.54)** & 0.53 (0.47) & 1.85 (0.82)** & -0.25 (1.25) & 386 \\
\addlinespace[0.3em]
\hdashline
\addlinespace[0.3em]
Non-top & Behavior & Preference Bias in Supervisors' Evaluations (上司の評価における気に入り重視) & Supervisors tend to prioritize personal favoritism over performance in their personnel evaluations of subordinates, with assessments influenced by factors such as closeness and ingratiation. (上司が部下の人事評価で成績よりも気に入るかどうかを重視し、親密さや媚びに基づいて評価を左右する傾向を示す。) & -15.80 (13.36) & -10.52 (24.89) & 6.67 (34.97) & -0.77 (0.79) & -0.54 (1.32) & -1.49 (1.80) & 120 \\
\addlinespace[0.3em]
\hdashline
\addlinespace[0.3em]
Non-top & Behavior & Presence or Absence of Overtime and Efforts to Reduce It (残業の有無・削減努力) & Among managerial staff, there are individuals who do not work overtime, and the current situation involves managing tasks within the policy aimed at reducing overtime hours. (管理職の中には残業をしない人もおり、残業を減らす方針の中で業務をやりくりしている状況) & -10.92 (17.77) & -9.12 (38.29) & 41.82 (60.92) & -0.50 (1.05) & -1.52 (2.04) & -2.34 (3.13) & 63 \\
\addlinespace[0.3em]
\hdashline
\addlinespace[0.3em]
Non-top & Behavior & Pressure from Supervisors Emphasizing Numerical Targets (数字重視の上司のプレッシャー) & A behavioral pattern in which supervisors strongly demand the achievement of numerical goals from their subordinates, applying rigorous pressure and frequent scrutiny on a daily basis. (上司が部下に対して数字の達成を強く求め、日々厳しい圧力や詰めをかける行動パターン。) & -2.54 (5.13) & 1.63 (8.56) & -2.02 (11.04) & -0.82 (0.30)*** & -1.11 (0.45)** & -1.80 (0.56)*** & 1002 \\
\addlinespace[0.3em]
\hdashline
\addlinespace[0.3em]
Non-top & Behavior & Proactive Acceptance of Opinions from Junior Employees (若手意見の積極的受容) & Reviews concerning behavioral patterns in which management actively listens to and utilizes the opinions and proposals of junior employees. (管理層が若手社員の意見や提案を積極的に聞き入れ、活用する行動パターンに関する口コミ。) & -2.51 (17.11) & -7.20 (26.53) & -7.96 (38.10) & 0.90 (1.01) & 0.05 (1.41) & -0.15 (1.95) & 96 \\
\addlinespace[0.3em]
\hdashline
\addlinespace[0.3em]
Non-top & Behavior & Proactive Appointment and Advancement of Women in Managerial Positions (女性管理職の積極的登用と活躍) & Positive feedback regarding initiatives and actual practices aimed at actively appointing women to managerial roles and promoting their success. (女性を管理職に積極的に登用し、活躍を推進する取り組みや実態に関する肯定的な口コミ。) & 1.91 (10.84) & -6.50 (19.47) & -24.87 (28.95) & 0.94 (0.64) & 0.37 (1.03) & -0.95 (1.49) & 233 \\
\addlinespace[0.3em]
\hdashline
\addlinespace[0.3em]
Non-top & Behavior & Proactive Behavior of Team Leaders and Organizational Management (チームリーダーの主体的な行動と組織運営) & This topic illustrates the behavioral patterns of team leaders who demonstrate proactivity and engage those around them to unify the organization and drive the achievement of goals. (チームリーダーが主体性を持ち、周囲を巻き込みながら組織をまとめて目標達成を推進する行動パターンを示す。) & 13.79 (5.02)*** & 13.28 (7.97)* & 25.22 (11.06)** & 0.51 (0.30)* & 0.17 (0.42) & -0.11 (0.57) & 1145 \\
\addlinespace[0.3em]
\hdashline
\addlinespace[0.3em]
Non-top & Behavior & Proactive Career Support (積極的なキャリア支援) & Reviews concerning supervisors' proactive efforts to support their subordinates' career development, aiming to facilitate growth and fulfill aspirations through consultations and interviews. (上司が部下のキャリア形成を積極的に支援し、相談や面談を通じて成長や希望を叶えようとする行動に関する口コミ。) & 22.80 (8.55)*** & 51.75 (11.95)*** & 62.21 (14.52)*** & 2.58 (0.50)*** & 3.20 (0.63)*** & 3.38 (0.74)*** & 639 \\
\addlinespace[0.3em]
\hdashline
\addlinespace[0.3em]
Non-top & Behavior & Proactive Communication (積極的なコミュニケーション) & The manner in which management actively engages in dialogue with employees, fostering an atmosphere conducive to open conversation and the exchange of ideas. (管理職が従業員と積極的に対話し、気軽に話せる雰囲気や意見交換を促進している様子。) & 24.97 (13.21)* & 25.66 (23.84) & 48.07 (33.01) & 0.50 (0.78) & -1.23 (1.27) & 0.31 (1.70) & 171 \\
\addlinespace[0.3em]
\hdashline
\addlinespace[0.3em]
Non-top & Behavior & Proactive Decision-Making (積極的な意思決定) & The manner in which middle managers independently make judgments and execute decisions and directives swiftly and assertively. (中間管理職が自ら判断し、迅速かつ強く意思決定や指示を行う様子) & -12.37 (11.69) & -10.80 (18.89) & -19.35 (26.07) & -0.78 (0.69) & -2.44 (1.00)** & -2.43 (1.33)* & 266 \\
\addlinespace[0.3em]
\hdashline
\addlinespace[0.3em]
Non-top & Behavior & Proactive Delegation of Authority (権限委譲の積極的実施) & A behavioral pattern in which supervisors actively entrust subordinates with responsible tasks and discretionary authority. (上司が部下に責任ある仕事や裁量を積極的に任せる行動パターン) & -21.39 (7.91)*** & -19.91 (13.55) & -46.61 (19.28)** & 0.87 (0.47)* & -0.77 (0.72) & 0.02 (0.99) & 511 \\
\addlinespace[0.3em]
\hdashline
\addlinespace[0.3em]
Non-top & Behavior & Proactive Development and Delegation of Discretion (積極的な育成と裁量付与) & An active approach to development that assigns responsible tasks and grants discretion to young employees in order to foster their growth. (若手社員に責任ある仕事や裁量を与え、成長を促す積極的な育成の取り組み) & 4.73 (12.07) & -13.64 (21.31) & -36.95 (25.85) & 1.15 (0.71) & 2.86 (1.13)** & 2.69 (1.32)** & 199 \\
\addlinespace[0.3em]
\hdashline
\addlinespace[0.3em]
Non-top & Behavior & Proactive Goal Setting and Progress Management (積極的な目標設定と進捗管理) & Reviews concerning managerial behaviors involving regular meetings to actively engage in goal setting, progress monitoring, evaluation, and feedback. (管理職が定期的に面談を行い、目標設定や進捗確認、評価・フィードバックを積極的に実施している行動に関する口コミ。) & -11.02 (9.11) & 3.72 (14.98) & -14.47 (19.34) & 1.60 (0.53)*** & 2.88 (0.78)*** & 3.20 (0.98)*** & 433 \\
\addlinespace[0.3em]
\hdashline
\addlinespace[0.3em]
Non-top & Behavior & Proactive Growth Support (積極的な成長支援) & This topic pertains to the actions taken by supervisors and leaders to actively support the growth and career development of their subordinates, providing opportunities for training and skill enhancement. (上司やリーダーが部下の成長やキャリア開発を積極的に支援し、育成やスキルアップの機会を提供する行動に関する内容。) & 11.94 (7.65) & 20.29 (13.60) & 13.88 (17.17) & 1.23 (0.45)*** & 2.92 (0.72)*** & 2.80 (0.87)*** & 499 \\
\bottomrule
\end{tabular}
\end{adjustbox}
\end{table*}


\begin{table*}[t]
\centering
\scriptsize
\setlength{\tabcolsep}{2pt}
\renewcommand{\arraystretch}{0.92}
\ContinuedFloat
\caption{Relationship between leadership topics and firm outcomes (continued)}
\begin{adjustbox}{max width=\textwidth}
\begin{tabular}{>{\raggedright\arraybackslash}p{0.07\textwidth}>{\raggedright\arraybackslash}p{0.08\textwidth}>{\raggedright\arraybackslash}p{0.18\textwidth}>{\raggedright\arraybackslash}p{0.30\textwidth}>{\centering\arraybackslash}p{0.06\textwidth}>{\centering\arraybackslash}p{0.06\textwidth}>{\centering\arraybackslash}p{0.06\textwidth}>{\centering\arraybackslash}p{0.06\textwidth}>{\centering\arraybackslash}p{0.06\textwidth}>{\centering\arraybackslash}p{0.06\textwidth}>{\centering\arraybackslash}p{0.05\textwidth}}
\toprule
Type. & Char. & Translated Topic Name (Original Topic Name) & Translated Topic Description (Original Topic Description) & \multicolumn{3}{c}{ROA} & \multicolumn{3}{c}{Employees Morale} & N \\
\cmidrule(lr){5-7}\cmidrule(lr){8-10}
 &  &  &  & $t=5$ & $t=10$ & $t=15$ & $t=5$ & $t=10$ & $t=15$ &  \\
\midrule
Non-top & Behavior & Proactive Guidance and Support (積極的な指導とサポート) & A behavioral pattern in which senior employees and supervisors actively provide education and care to junior staff and newcomers, offering thorough and empathetic instruction. (先輩社員や上司が後輩や新人に対して積極的に教育や面倒見を行い、丁寧かつ親身に指導する行動パターン。) & 3.36 (12.47) & -3.47 (22.93) & -13.98 (29.16) & -0.36 (0.74) & -1.23 (1.22) & -0.11 (1.49) & 196 \\
\addlinespace[0.3em]
\hdashline
\addlinespace[0.3em]
Non-top & Behavior & Proactive Influence of Externally Recruited Managers (外部出身管理職の積極的影響力) & The situation in which managers hired from outside the organization actively influence corporate policies and on-site operations, driving reforms and changes. (外部から採用された管理職が企業の方針や現場運営に積極的に影響を与え、改革や変化を推進している状況。) & -8.43 (28.04) & 99.61 (94.57) & 87.79 (96.49) & -1.80 (1.66) & 1.36 (5.03) & -0.82 (4.96) & 20 \\
\addlinespace[0.3em]
\hdashline
\addlinespace[0.3em]
Non-top & Behavior & Proactive Management Behaviors (積極的なマネジメント行動) & Behavioral patterns of supervisors who actively engage in supporting subordinate development, coordinating tasks, and facilitating communication, providing assistance and guidance. (部下の成長支援や業務調整、コミュニケーションに積極的に関与し、支援や指導を行う上長の行動パターン。) & 6.58 (3.84)* & 13.17 (6.20)** & 15.06 (8.46)* & 0.79 (0.23)*** & 1.38 (0.33)*** & 2.43 (0.43)*** & 2153 \\
\addlinespace[0.3em]
\hdashline
\addlinespace[0.3em]
Non-top & Behavior & Proactive Management Implementation (積極的なマネジメント実施) & A collection of texts demonstrating the proactive attitude of managers and leaders who independently carry out management duties and actively engage in organizational operations, subordinate supervision, and problem-solving. (管理職やリーダーが主体的にマネジメント業務を遂行し、組織運営や部下管理、課題解決に積極的に取り組む姿勢を示す文章群。) & 35.64 (20.32)* & 43.77 (25.29)* & 55.31 (36.54) & 1.36 (1.20) & 0.69 (1.35) & 0.47 (1.88) & 139 \\
\addlinespace[0.3em]
\hdashline
\addlinespace[0.3em]
Non-top & Behavior & Proactive On-Site Management (積極的な現場管理) & Actions undertaken by site supervisors and leaders to actively manage the site, including providing guidance, addressing issues, and attending to the well-being of workers. (現場責任者やリーダーが積極的に現場を管理し、指導やトラブル対応、作業員のケアを行う行動) & -10.34 (11.56) & -18.84 (18.46) & -50.23 (30.45)* & -0.77 (0.68) & -1.72 (0.98)* & -2.06 (1.55) & 209 \\
\addlinespace[0.3em]
\hdashline
\addlinespace[0.3em]
Non-top & Behavior & Proactive Regular Meetings (積極的な定期面談) & A behavioral pattern in which supervisors conduct regular meetings to actively engage in evaluating subordinates, discussing career development, and providing feedback. (上司が定期的に面談を実施し、部下の評価やキャリア相談、フィードバックを積極的に行う行動パターン。) & 6.40 (6.81) & 24.15 (10.84)** & 27.18 (14.19)* & 1.74 (0.40)*** & 3.23 (0.57)*** & 4.05 (0.72)*** & 750 \\
\addlinespace[0.3em]
\hdashline
\addlinespace[0.3em]
Non-top & Behavior & Proactive Store Management (積極的な店舗管理) & A behavioral pattern in which area managers frequently visit stores to actively conduct evaluations, provide instructions, and offer follow-up support. (エリアマネージャーが店舗を頻繁に訪問し、評価や指示、フォローを積極的に行う行動パターン) & 14.77 (17.83) & -3.97 (36.14) & -19.68 (42.40) & -0.45 (1.05) & 1.38 (1.92) & 0.85 (2.18) & 77 \\
\addlinespace[0.3em]
\hdashline
\addlinespace[0.3em]
Non-top & Behavior & Proactive Store Manager Operations and Leadership (積極的な店長の運営と指導) & A collection of texts illustrating store managers taking active responsibility for store operations, staff training, and shift scheduling, demonstrating a positive and forward-looking approach. (店長が積極的に店舗運営やスタッフ教育、シフト作成などに責任を持ち、前向きに行動している様子を示す文章群。) & 0.30 (4.11) & -0.57 (7.11) & 0.24 (8.80) & -0.07 (0.24) & -0.01 (0.38) & 0.63 (0.45) & 732 \\
\addlinespace[0.3em]
\hdashline
\addlinespace[0.3em]
Non-top & Behavior & Proactive Supervisors in Employee Consultations (相談に積極的な上司) & Behavioral patterns of supervisors who attentively respond to their subordinates' consultations and actively provide support and interventions. (部下の相談に親身に応じ、積極的に支援や対応を行う上司の行動パターン) & 2.14 (4.97) & 17.84 (9.24)* & 22.32 (13.40)* & 0.71 (0.29)** & 0.92 (0.49)* & 1.84 (0.69)*** & 1154 \\
\addlinespace[0.3em]
\hdashline
\addlinespace[0.3em]
Non-top & Behavior & Proactive Support for Young Employee Development (若手育成の積極的支援) & Reviews concerning the behaviors and attitudes of managers who actively engage in the development and support of young employees. (管理職が若手社員の育成やサポートに積極的に取り組んでいる行動や姿勢に関する口コミ。) & 45.30 (50.23) & 231.39 (109.42)** & 214.66 (111.68)* & 5.86 (2.97)** & -1.28 (5.82) & -4.08 (5.75) & 21 \\
\addlinespace[0.3em]
\hdashline
\addlinespace[0.3em]
Non-top & Behavior & Proactive Utilization of Paid Leave (積極的な有給取得) & Behaviors and circumstances in which managers take paid leave themselves proactively and encourage their subordinates to do the same. (管理職が自ら率先して有給休暇を取得し、部下にも取得を促進している行動や状況。) & 2.02 (8.79) & 12.83 (14.40) & 10.23 (21.13) & -0.03 (0.52) & 1.19 (0.76) & 1.25 (1.08) & 370 \\
\addlinespace[0.3em]
\hdashline
\addlinespace[0.3em]
Non-top & Behavior & Proactive Utilization of Telecommuting (在宅勤務の積極的活用) & The active authorization and utilization of telecommuting and flexible working hours by management, promoting flexible work styles. (管理職が在宅勤務やフレックス勤務を積極的に許可・活用し、柔軟な働き方を推進している様子。) & 33.48 (15.71)** & 42.93 (19.84)** & 53.24 (26.54)** & 0.69 (0.93) & 0.15 (1.06) & 0.69 (1.36) & 219 \\
\addlinespace[0.3em]
\hdashline
\addlinespace[0.3em]
Non-top & Behavior & Proactive Utilization of Transfers and Relocations (異動・転勤の積極的活用) & This refers to situations in which managerial personnel voluntarily apply for transfers or relocations, or actively use these opportunities to gain experience. (管理職が異動や転勤を志願したり、経験を積むために活用している状況を示す。) & -61.67 (33.18)* & -101.56 (58.38)* & -34.29 (85.13) & -4.51 (1.96)** & -4.31 (3.10) & -3.96 (4.38) & 34 \\
\addlinespace[0.3em]
\hdashline
\addlinespace[0.3em]
Non-top & Behavior & Proactively Delegating Work to Junior Employees (若手への積極的な仕事任せ) & Leaders' actions of actively assigning responsible tasks and opportunities to junior employees to promote their growth. (若手社員に責任ある仕事やチャンスを積極的に与え、成長を促すリーダーの行動。) & 9.25 (13.71) & 43.57 (24.91)* & 33.91 (30.14) & 0.94 (0.81) & 1.94 (1.32) & 1.78 (1.54) & 181 \\
\addlinespace[0.3em]
\hdashline
\addlinespace[0.3em]
Non-top & Behavior & Project Advancement (プロジェクト推進) & Reviews concerning leaders' proactive efforts in advancing projects, including their management and coordination activities. (リーダーがプロジェクトを積極的に推進し、管理や調整を行う行動に関する口コミ) & 21.35 (6.83)*** & 32.39 (12.65)** & 36.08 (16.48)** & 0.98 (0.40)** & 1.72 (0.67)*** & 2.11 (0.84)** & 581 \\
\addlinespace[0.3em]
\hdashline
\addlinespace[0.3em]
Non-top & Behavior & Project Constraints (プロジェクト制約) & Reviews concerning situations where the leader's discretion and role are limited, with a focus on schedule management and detailed inspections. (リーダーの裁量や役割が限定的で、日程管理や細かいチェックに注力する状況に関する口コミ) & 31.80 (18.84)* & 58.53 (35.75) & 49.64 (40.26) & 1.00 (1.11) & 3.21 (1.89)* & 3.83 (2.05)* & 99 \\
\addlinespace[0.3em]
\hdashline
\addlinespace[0.3em]
Non-top & Behavior & Promotion of Paid Leave Utilization (有給取得促進) & The actions taken by supervisors to actively encourage their subordinates to take paid leave, including recommending and managing its usage. (上長が部下に対して積極的に有給休暇の取得を促し、取得を推奨・管理する行動。) & -10.24 (12.78) & -6.40 (16.21) & -9.85 (21.51) & -0.27 (0.76) & -0.70 (0.86) & -1.26 (1.11) & 259 \\
\addlinespace[0.3em]
\hdashline
\addlinespace[0.3em]
Non-top & Behavior & Promotion of Paid Leave Utilization (有給取得促進) & Actions and attitudes whereby managers and supervisors actively encourage their subordinates to take paid leave and establish an environment conducive to its utilization. (管理職や上司が部下に対して積極的に有給休暇の取得を促し、取得環境を整備する行動や態度。) & -29.95 (21.83) & -49.37 (30.29) & -62.06 (36.09)* & -0.48 (1.29) & -1.27 (1.61) & -1.42 (1.86) & 114 \\
\bottomrule
\end{tabular}
\end{adjustbox}
\end{table*}


\begin{table*}[t]
\centering
\scriptsize
\setlength{\tabcolsep}{2pt}
\renewcommand{\arraystretch}{0.92}
\ContinuedFloat
\caption{Relationship between leadership topics and firm outcomes (continued)}
\begin{adjustbox}{max width=\textwidth}
\begin{tabular}{>{\raggedright\arraybackslash}p{0.07\textwidth}>{\raggedright\arraybackslash}p{0.08\textwidth}>{\raggedright\arraybackslash}p{0.18\textwidth}>{\raggedright\arraybackslash}p{0.30\textwidth}>{\centering\arraybackslash}p{0.06\textwidth}>{\centering\arraybackslash}p{0.06\textwidth}>{\centering\arraybackslash}p{0.06\textwidth}>{\centering\arraybackslash}p{0.06\textwidth}>{\centering\arraybackslash}p{0.06\textwidth}>{\centering\arraybackslash}p{0.06\textwidth}>{\centering\arraybackslash}p{0.05\textwidth}}
\toprule
Type. & Char. & Translated Topic Name (Original Topic Name) & Translated Topic Description (Original Topic Description) & \multicolumn{3}{c}{ROA} & \multicolumn{3}{c}{Employees Morale} & N \\
\cmidrule(lr){5-7}\cmidrule(lr){8-10}
 &  &  &  & $t=5$ & $t=10$ & $t=15$ & $t=5$ & $t=10$ & $t=15$ &  \\
\midrule
Non-top & Behavior & Prompt Approval Response (迅速な承認対応) & A behavioral pattern in which supervisors promptly carry out approvals and authorizations, facilitating decision-making with a sense of urgency. (上司が承認や決裁を迅速に行い、スピード感を持って意思決定を進める行動様式) & -37.07 (26.00) & -63.01 (54.18) & -64.79 (61.97) & -1.83 (1.54) & -0.75 (2.88) & 0.01 (3.19) & 41 \\
\addlinespace[0.3em]
\hdashline
\addlinespace[0.3em]
Non-top & Behavior & Protective Actions in the Workplace (現場保護行動) & A situation in which middle managers protect frontline operations from unreasonable demands imposed by upper management and attempt effective responses while caught in a dilemma. (中間管理職が上層部の無理難題から現場を守り、板挟みの中で実効的な対応を試みる状況。) & 16.59 (23.10) & 9.56 (32.67) & -18.38 (50.69) & -1.20 (1.37) & -1.09 (1.74) & -0.01 (2.61) & 90 \\
\addlinespace[0.3em]
\hdashline
\addlinespace[0.3em]
Non-top & Behavior & Recommendation and Support for Training (研修推奨・支援) & Reviews concerning supervisors' proactive encouragement and support of their subordinates' participation in training programs. (上司が部下の研修受講を積極的に推奨し、支援する行動に関する口コミ。) & 12.99 (13.30) & 16.61 (19.95) & 44.32 (27.93) & -0.45 (0.79) & -0.84 (1.06) & -0.14 (1.43) & 185 \\
\addlinespace[0.3em]
\hdashline
\addlinespace[0.3em]
Non-top & Behavior & Rejection of Unpaid Overtime Work (サービス残業否定) & Reviews indicating that managers do not compel their subordinates to engage in unpaid overtime work or force them to work beyond regular hours. (管理職が部下にサービス残業を強いない、または残業を強要しないとする口コミ) & -2.27 (15.83) & -7.48 (26.45) & -5.12 (34.42) & -1.86 (0.93)** & -1.95 (1.40) & -1.37 (1.77) & 95 \\
\addlinespace[0.3em]
\hdashline
\addlinespace[0.3em]
Non-top & Behavior & Reluctant Attitude Toward Development (育成に消極的な態度) & This refers to situations where leaders or supervisors lack sufficient time or capacity to adequately support the growth and development of their subordinates, resulting in insufficient developmental support. (リーダーや上司が部下の育成に十分な時間や余裕がなく、成長支援が不十分であることを示す内容。) & 47.22 (29.12) & 52.16 (43.60) & 60.01 (93.80) & -0.57 (1.62) & -4.29 (2.32)* & -2.65 (4.82) & 33 \\
\addlinespace[0.3em]
\hdashline
\addlinespace[0.3em]
Non-top & Behavior & Responsibilities and Management in Classroom Administration (教室運営の責任と管理) & This topic addresses the roles of classroom and school directors in overseeing classroom operations, managing staff, and guiding students, emphasizing their accountability in executing these duties. (教室長や校舎長が教室運営やスタッフ管理、生徒指導に責任を持ち、業務を遂行する役割に関する内容) & 14.35 (7.15)** & 12.39 (11.55) & 19.55 (14.02) & -0.86 (0.42)** & -0.20 (0.61) & -1.04 (0.72) & 100 \\
\addlinespace[0.3em]
\hdashline
\addlinespace[0.3em]
Non-top & Behavior & Restrictions and Management of Leave Acquisition (休暇取得の制限・管理) & Reviews concerning managerial attitudes such as store managers or supervisors imposing restrictions on taking leave, denying requested days off, or making decisions at their discretion. (店長や上司が休暇取得に制限をかけたり、希望休を通さない、裁量で決めるなど管理的な態度に関する口コミ。) & 13.96 (9.23) & 8.42 (18.72) & -5.50 (24.76) & 0.63 (0.54) & 1.30 (0.99) & 3.17 (1.25)** & 160 \\
\addlinespace[0.3em]
\hdashline
\addlinespace[0.3em]
Non-top & Behavior & Restrictions and Negative Attitudes Toward Telecommuting (在宅勤務の制限・否定的対応) & The phenomenon whereby managers impose limitations on the use of telecommuting and flexible work arrangements or exhibit negative attitudes, thereby creating barriers to their utilization. (管理職が在宅勤務やフレックス勤務の利用を制限したり否定的な態度を示し、利用しにくくしている様子。) & -2.93 (11.24) & 7.87 (21.97) & 5.10 (33.39) & -0.54 (0.66) & -0.53 (1.14) & -0.87 (1.63) & 192 \\
\addlinespace[0.3em]
\hdashline
\addlinespace[0.3em]
Non-top & Behavior & Restrictions on Paid Leave Acquisition (有給取得制限) & Actions taken by supervisors to restrict or respond negatively to the acquisition of paid leave, including refusing to allow exceptions, thereby suppressing the use of paid leave. (上長が有給休暇の取得に対して制限や否定的な対応を行い、例外を認めないなど取得を抑制する行動。) & 2.09 (17.78) & 11.46 (37.42) & 11.18 (76.57) & -0.73 (1.05) & -0.43 (1.99) & -3.48 (3.93) & 54 \\
\addlinespace[0.3em]
\hdashline
\addlinespace[0.3em]
Non-top & Behavior & Restrictions on Taking Paid Leave (有給休暇取得の制限) & Reviews concerning behaviors and attitudes of supervisors and leaders that restrict subordinates from taking paid leave, thereby making it difficult for them to utilize their entitled time off. (上司やリーダーが部下の有給休暇取得を制限し、取得を困難にする行動や態度に関する口コミ。) & -3.69 (6.39) & 22.37 (11.49)* & 20.86 (14.08) & -1.37 (0.38)*** & -1.98 (0.61)*** & -2.45 (0.72)*** & 593 \\
\addlinespace[0.3em]
\hdashline
\addlinespace[0.3em]
Non-top & Behavior & Rigid and Conservative Attitudes Toward Policies and Instructions (方針や指示の固定・保守的態度) & Reviews concerning supervisors who maintain outdated methods without change, exhibiting behavior that rigidly preserves policies and instructions. (上司が古いやり方を変えず、方針や指示を固定的に維持する行動に関する口コミ。) & 9.27 (27.68) & -26.27 (32.65) & -4.35 (60.18) & 0.25 (1.64) & -0.35 (1.74) & 0.61 (3.09) & 44 \\
\addlinespace[0.3em]
\hdashline
\addlinespace[0.3em]
Non-top & Behavior & Ritualistic Goal Management and Neglect (形骸化した目標管理と放置) & Reviews concerning situations where goals are set but subsequent follow-up and progress monitoring are inadequate, resulting in superficial evaluations and a hollowed-out management process. (目標設定は行うものの、その後のフォローや進捗管理が不十分で、評価が雑であったり、管理が形骸化している状況に関する口コミ。) & -12.52 (16.24) & 16.60 (35.83) & 22.23 (45.40) & -0.39 (0.96) & -0.13 (1.88) & -1.36 (2.29) & 87 \\
\addlinespace[0.3em]
\hdashline
\addlinespace[0.3em]
Non-top & Behavior & Severe Reprimand Behaviors (厳しい叱責行動) & Char.s of behaviors in which supervisors or leaders engage in strong reprimands, such as being overbearing, intimidating, or shouting. (上司やリーダーが高圧的、威圧的、怒鳴るなど強い叱責を行う行動の特徴) & 2.97 (5.17) & 3.96 (8.70) & -8.10 (11.96) & -0.49 (0.31) & -0.79 (0.46)* & -1.23 (0.61)** & 894 \\
\addlinespace[0.3em]
\hdashline
\addlinespace[0.3em]
Non-top & Behavior & Special Treatment and Perceived Lack of Competence Among Female Managers (女性管理職の特別扱いや能力不足の指摘) & Negative reviews suggesting that female managers receive preferential treatment disproportionate to their abilities, or that they have been promoted against their own will. (女性管理職が能力に見合わず特別扱いされている、または本人の意思に反して昇進しているという否定的な口コミ。) & -23.62 (29.76) & -39.97 (32.67) & -42.36 (54.36) & 0.29 (1.76) & 0.03 (1.74) & 3.11 (2.79) & 46 \\
\addlinespace[0.3em]
\hdashline
\addlinespace[0.3em]
Non-top & Behavior & Status of Paid Leave Utilization Among Managers (管理職の有給休暇取得状況) & Reviews concerning the difficulties managers face in taking paid leave and the constraints affecting their ability to utilize vacation time. (管理職が有給休暇を取りにくく、休暇取得が制約されている実態についての口コミ。) & 10.34 (7.55) & 3.85 (11.86) & 10.78 (18.78) & -0.83 (0.45)* & -1.25 (0.63)** & -2.24 (0.96)** & 445 \\
\addlinespace[0.3em]
\hdashline
\addlinespace[0.3em]
Non-top & Behavior & Store Manager's Holiday Duties (店長の休日対応) & Reviews concerning the actual conditions under which store managers handle tasks such as answering phone calls and reporting to work even on holidays. (店長が休日でも電話対応や出勤などの業務を行う実態に関する口コミ) & -4.90 (7.97) & -1.32 (15.82) & -1.56 (18.58) & -1.46 (0.47)*** & -1.38 (0.84)* & -1.36 (0.95) & 215 \\
\addlinespace[0.3em]
\hdashline
\addlinespace[0.3em]
Non-top & Behavior & Strict Evaluation and Assessment (厳格な評価と査定) & Reviews concerning supervisors conducting rigorous evaluations and assessments, including the implementation of salary reductions and stringent bonus appraisals. (上司が厳しく評価や査定を行い、減給処分やボーナスの厳しい査定を実施する行動に関する口コミ。) & -17.74 (20.27) & -43.62 (30.08) & -71.18 (49.74) & -2.24 (1.20)* & -3.59 (1.60)** & -2.73 (2.56) & 61 \\
\addlinespace[0.3em]
\hdashline
\addlinespace[0.3em]
Non-top & Behavior & Strict Evaluation and Supervision (厳格な評価と指導) & An approach in which branch managers and supervisors emphasize performance and discipline through rigorous guidance and critical oversight. (支店長や上司が厳しい指導や詰める行動を通じて業績や規律を重視する態度) & -15.72 (7.46)** & -21.65 (11.28)* & -32.20 (14.97)** & -1.32 (0.44)*** & -1.29 (0.60)** & -1.75 (0.77)** & 329 \\
\bottomrule
\end{tabular}
\end{adjustbox}
\end{table*}


\begin{table*}[t]
\centering
\scriptsize
\setlength{\tabcolsep}{2pt}
\renewcommand{\arraystretch}{0.92}
\ContinuedFloat
\caption{Relationship between leadership topics and firm outcomes (continued)}
\begin{adjustbox}{max width=\textwidth}
\begin{tabular}{>{\raggedright\arraybackslash}p{0.07\textwidth}>{\raggedright\arraybackslash}p{0.08\textwidth}>{\raggedright\arraybackslash}p{0.18\textwidth}>{\raggedright\arraybackslash}p{0.30\textwidth}>{\centering\arraybackslash}p{0.06\textwidth}>{\centering\arraybackslash}p{0.06\textwidth}>{\centering\arraybackslash}p{0.06\textwidth}>{\centering\arraybackslash}p{0.06\textwidth}>{\centering\arraybackslash}p{0.06\textwidth}>{\centering\arraybackslash}p{0.06\textwidth}>{\centering\arraybackslash}p{0.05\textwidth}}
\toprule
Type. & Char. & Translated Topic Name (Original Topic Name) & Translated Topic Description (Original Topic Description) & \multicolumn{3}{c}{ROA} & \multicolumn{3}{c}{Employees Morale} & N \\
\cmidrule(lr){5-7}\cmidrule(lr){8-10}
 &  &  &  & $t=5$ & $t=10$ & $t=15$ & $t=5$ & $t=10$ & $t=15$ &  \\
\midrule
Non-top & Behavior & Strict Supervision and Reprimand (厳しい指導と叱責) & Behavioral patterns in which managers exert intense pressure and issue reprimands in response to failure to meet sales targets. (営業目標未達に対して管理職が強いプレッシャーや叱責を行う行動パターン) & 4.78 (9.46) & 18.31 (16.11) & 3.17 (22.89) & 0.03 (0.56) & 0.33 (0.86) & -0.48 (1.17) & 294 \\
\addlinespace[0.3em]
\hdashline
\addlinespace[0.3em]
Non-top & Behavior & Supervisors Reluctant to Engage in Consultations (相談に消極的な上司) & Behavioral patterns of supervisors who are reluctant or unresponsive to consultations from their subordinates. (部下の相談に対して消極的、または対応しない上司の行動パターン) & 8.26 (18.29) & 2.24 (34.33) & -5.53 (44.31) & -1.13 (1.08) & 1.27 (1.83) & 0.80 (2.28) & 74 \\
\addlinespace[0.3em]
\hdashline
\addlinespace[0.3em]
Non-top & Behavior & Supervisory Responses to Inhibiting Employees' Timely Departure (帰宅阻害の上司対応) & Reviews concerning supervisors who do not permit subordinates to leave work early or on time, thereby obstructing their departure or creating an atmosphere that makes leaving difficult. (上司が部下の早めの退社や定時退社を許さず、帰宅を阻害したり帰りにくい雰囲気を作る態度や行動に関する口コミ) & -6.47 (10.85) & -6.55 (18.18) & 7.35 (28.40) & -0.72 (0.64) & -1.05 (0.97) & -2.84 (1.46)* & 218 \\
\addlinespace[0.3em]
\hdashline
\addlinespace[0.3em]
Non-top & Behavior & Support for Challenging Initiatives (挑戦支援) & Actions whereby supervisors respect their subordinates' willingness to take on challenges and proactively provide opportunities to engage in new tasks and assignments. (上司が部下の挑戦意欲を尊重し、新しい業務や課題に取り組む機会を積極的に提供する行動) & 0.49 (9.71) & -14.99 (14.45) & -15.09 (18.86) & 1.72 (0.57)*** & 0.96 (0.76) & 0.57 (0.96) & 398 \\
\addlinespace[0.3em]
\hdashline
\addlinespace[0.3em]
Non-top & Behavior & Supportive Attitude of Branch Managers (支店長の支援的態度) & Supportive behavioral patterns exhibited by branch managers, such as encouraging subordinates to take paid leave and actively engaging in their evaluation and promotion processes. (支店長が部下の有給取得を促したり、評価や昇格に積極的に関与する支援的な行動パターン) & -14.38 (7.38)* & -23.48 (12.75)* & -41.10 (17.59)** & -0.32 (0.44) & -0.34 (0.67) & -0.73 (0.90) & 278 \\
\addlinespace[0.3em]
\hdashline
\addlinespace[0.3em]
Non-top & Behavior & Suppression of Paid Leave Utilization (有給取得抑制) & A situation in which managers or supervisors do not encourage the use of paid leave, thereby creating an environment where taking paid leave is difficult, or exhibit a passive attitude toward employees taking such leave. (管理職や上司が有給休暇の取得を促さず、取得しにくい環境を作る、または取得に消極的な態度。) & 3.08 (68.39) & -8.71 (69.01) & -56.68 (85.81) & -0.59 (4.04) & -1.56 (3.67) & 0.12 (4.41) & 11 \\
\addlinespace[0.3em]
\hdashline
\addlinespace[0.3em]
Non-top & Behavior & The Coercion and Reality of Long Working Hours Overtime (長時間残業の強制と実態) & Reviews concerning the actual conditions in which leadership personnel are compelled to work long hours of overtime or force their subordinates to do so. (リーダー層が長時間残業を強いられたり、部下に残業を強制する実態に関する口コミ) & 0.34 (6.59) & 3.16 (11.25) & 7.26 (15.22) & -0.33 (0.39) & -0.55 (0.60) & -0.96 (0.78) & 741 \\
\addlinespace[0.3em]
\hdashline
\addlinespace[0.3em]
Non-top & Behavior & The Presence of Subjective Evaluation (主観的な評価の存在) & Opinions regarding the tendency of supervisors to make evaluations and promotion decisions based on intuition, personal preferences, and individual discretion. (上司が感覚や好み、個人的な裁量で評価や昇格判断を行う傾向に関する意見) & -1.86 (7.60) & 4.88 (12.01) & -0.87 (16.17) & 0.61 (0.45) & -0.04 (0.63) & -1.11 (0.82) & 583 \\
\addlinespace[0.3em]
\hdashline
\addlinespace[0.3em]
Non-top & Behavior & The Reality of Frequent Transfers and Reassignments (頻繁な異動・転勤の実態) & This refers to the situation in which managerial personnel undergo frequent transfers and reassignments within a short period. (管理職が短期間で頻繁に異動や転勤を繰り返す状況を示す。) & 10.20 (10.44) & -6.03 (16.82) & -5.25 (26.23) & -1.61 (0.62)*** & -1.71 (0.89)* & -1.20 (1.33) & 288 \\
\addlinespace[0.3em]
\hdashline
\addlinespace[0.3em]
Non-top & Behavior & Training Restrictions and Passive Responses (研修制限・消極的対応) & Reviews concerning behaviors in which supervisors or managers impose limitations on training participation or exhibit passive attitudes toward it. (上司や管理職が研修受講に対して制限を設けたり、消極的な態度を示す行動に関する口コミ。) & 12.66 (27.57) & -0.17 (31.76) & 13.81 (42.06) & -3.32 (1.63)** & -3.37 (1.69)** & -2.92 (2.16) & 74 \\
\addlinespace[0.3em]
\hdashline
\addlinespace[0.3em]
Non-top & Behavior & Understanding and Cooperative Support (理解と協力的対応) & Behaviors and statements in which supervisors demonstrate understanding and provide cooperative support regarding the taking of childcare leave and maternity leave. (上司が育児休業や産休取得に対して理解を示し、協力的に対応する行動や言動。) & 0.87 (11.57) & 0.06 (17.92) & 4.68 (25.19) & 0.56 (0.68) & 0.83 (0.95) & 0.62 (1.29) & 275 \\
\addlinespace[0.3em]
\hdashline
\addlinespace[0.3em]
Non-top & Behavior & Unfair Evaluation (不公正な評価) & A situation in which managerial personnel make arbitrary and biased judgments in evaluations, leading to issues such as demotions, manipulation of assessments, and lack of transparency in evaluation criteria. (管理職が評価において恣意的、偏った判断を行い、降格や評価の操作、評価基準の不透明さが問題となっている状況。) & 20.41 (12.14)* & 19.00 (18.54) & 13.35 (27.73) & 0.80 (0.72) & 1.74 (0.99)* & 1.30 (1.42) & 206 \\
\addlinespace[0.3em]
\hdashline
\addlinespace[0.3em]
Non-top & Behavior & Unfair Evaluation Practices (不公正な評価行動) & Instances in which leaders conduct evaluations lacking fairness based on relationships, personal preferences, or self-interest. (リーダーが関係性や好み、自己都合に基づき、公正さを欠いた評価を行う行動の事例) & 13.10 (11.68) & 33.63 (20.52) & 30.71 (26.39) & -1.04 (0.69) & -0.30 (1.09) & -0.68 (1.36) & 236 \\
\addlinespace[0.3em]
\hdashline
\addlinespace[0.3em]
Non-top & Behavior & Unfair Evaluations (不公平な評価) & Reviews concerning biases in evaluations and salary increases based on store managers' personal preferences and interpersonal compatibility. (店長の個人的な好みや相性に基づく評価や昇給の偏りに関する口コミ) & 1.63 (11.25) & -1.13 (14.40) & -11.63 (28.07) & 0.69 (0.66) & -0.48 (0.76) & 0.06 (1.41) & 127 \\
\addlinespace[0.3em]
\hdashline
\addlinespace[0.3em]
Non-top & Behavior & Unfair Evaluations and Instances of Undervaluation (評価の不公平・低評価事例) & Opinions concerning cases where supervisors' evaluations are unfair, resulting in low ratings or unjust treatment that do not correspond to an individual's abilities or achievements. (上司の評価が不公平であり、能力や成果に見合わない低評価や不当な扱いを受けるケースに関する意見) & -0.54 (31.90) & 5.33 (58.94) & -2.00 (71.74) & 2.40 (1.89) & 1.58 (3.13) & 3.25 (3.69) & 27 \\
\addlinespace[0.3em]
\hdashline
\addlinespace[0.3em]
Non-top & Behavior & Unfair and Inadequate Performance Evaluations (評価の不公平・不十分) & Reviews concerning supervisors who do not fairly assess their subordinates' performance, resulting in evaluations that are difficult to reflect accurately or are handled in an unjust manner. (上司が部下の業績を正当に評価せず、評価に反映しにくい、または不公平な対応をすることに関する口コミ) & -9.61 (19.26) & -28.99 (28.83) & -32.76 (42.57) & 0.70 (1.14) & 0.50 (1.53) & 0.24 (2.19) & 85 \\
\addlinespace[0.3em]
\hdashline
\addlinespace[0.3em]
Non-top & Behavior & Variability in Supervisors' Evaluation Criteria (上司の評価基準のばらつき) & Refers to the situation where significant differences and biases arise in personnel evaluations due to differing evaluation criteria and implementation methods among supervisors. (上司ごとに異なる評価基準や運用方法により、人事評価に大きな差異や偏りが生じる状況を指す。) & 22.62 (12.68)* & 28.45 (21.24) & 55.55 (36.10) & -0.79 (0.75) & 0.13 (1.12) & 0.41 (1.81) & 146 \\
\addlinespace[0.3em]
\hdashline
\addlinespace[0.3em]
Non-top & Behavior & Voluntary Holiday Response (自主的な休日対応) & A pattern in which managers proactively make contact or take action during holidays or late at night, responding according to emergencies or operational necessities. (管理職が自主的に休日や深夜に連絡や対応を行い、緊急時や業務上の必要性に応じて行動するパターン。) & -8.96 (11.80) & -21.30 (17.46) & 10.75 (26.09) & 0.14 (0.70) & -0.65 (0.92) & -0.51 (1.33) & 230 \\
\bottomrule
\end{tabular}
\end{adjustbox}
\end{table*}


\begin{table*}[t]
\centering
\scriptsize
\setlength{\tabcolsep}{2pt}
\renewcommand{\arraystretch}{0.92}
\ContinuedFloat
\caption{Relationship between leadership topics and firm outcomes (continued)}
\begin{adjustbox}{max width=\textwidth}
\begin{tabular}{>{\raggedright\arraybackslash}p{0.07\textwidth}>{\raggedright\arraybackslash}p{0.08\textwidth}>{\raggedright\arraybackslash}p{0.18\textwidth}>{\raggedright\arraybackslash}p{0.30\textwidth}>{\centering\arraybackslash}p{0.06\textwidth}>{\centering\arraybackslash}p{0.06\textwidth}>{\centering\arraybackslash}p{0.06\textwidth}>{\centering\arraybackslash}p{0.06\textwidth}>{\centering\arraybackslash}p{0.06\textwidth}>{\centering\arraybackslash}p{0.06\textwidth}>{\centering\arraybackslash}p{0.05\textwidth}}
\toprule
Type. & Char. & Translated Topic Name (Original Topic Name) & Translated Topic Description (Original Topic Description) & \multicolumn{3}{c}{ROA} & \multicolumn{3}{c}{Employees Morale} & N \\
\cmidrule(lr){5-7}\cmidrule(lr){8-10}
 &  &  &  & $t=5$ & $t=10$ & $t=15$ & $t=5$ & $t=10$ & $t=15$ &  \\
\midrule
Non-top & Behavior & Voluntary Participation in Social Interactions (自主的な交流参加) & A behavioral pattern in which managers actively and voluntarily engage in social activities such as drinking parties and golf, aiming to build relationships with their subordinates. (管理職が飲み会やゴルフなどの交流に積極的かつ自主的に参加し、部下との関係構築を図る行動パターン。) & -1.98 (9.78) & -7.94 (15.51) & 0.81 (21.74) & 0.55 (0.58) & 0.57 (0.82) & 0.40 (1.12) & 223 \\
\addlinespace[0.3em]
\hdashline
\addlinespace[0.3em]
Non-top & Behavior & Work Style Innovations and Constraints (働き方の工夫と制約) & Descriptions concerning managerial efforts to mitigate workload, adjustments to working hours, and awareness aimed at avoiding excessive work. (管理職が業務負荷を抑えようとする努力や、勤務時間調整、ハードワーク回避の意識に関する記述) & 18.95 (15.66) & 16.27 (25.00) & -15.81 (35.48) & 0.11 (0.92) & 1.01 (1.32) & 0.93 (1.81) & 155 \\
\addlinespace[0.3em]
\hdashline
\addlinespace[0.3em]
Non-top & Behavior & Workload Intensity (業務負荷の重さ) & A description concerning the busyness, volume of tasks, exhaustion, and high levels of burden experienced by managerial staff. (管理職が抱える多忙さや業務量の多さ、疲弊や負荷の高さに関する記述) & 7.47 (4.72) & 5.23 (8.24) & 3.58 (11.61) & -0.58 (0.28)** & -0.40 (0.44) & -0.62 (0.59) & 1324 \\
\addlinespace[0.3em]
\hdashline
\addlinespace[0.3em]
Top & Ability & Advanced Management Skills and Foresight of Top Executives (経営トップの高度な経営力と先見性) & An evaluation demonstrating the superior leadership, expertise, foresight, and multifaceted practical abilities of presidents and chairpersons. (社長や会長の優れたリーダーシップ、専門知識、先見性、そして多方面にわたる実践的能力を示す評価。) & 17.59 (4.66)*** & 27.88 (8.58)*** & 50.92 (13.17)*** & 1.97 (0.27)*** & 2.91 (0.45)*** & 1.79 (0.68)*** & 930 \\
\addlinespace[0.3em]
\hdashline
\addlinespace[0.3em]
Top & Ability & Deficiency in Executive Leadership Competencies (経営トップの能力不足) & This refers to a situation characterized by a general lack of managerial capabilities among top executives, including strategic planning, marketing, and business development. (経営トップ層の戦略立案やマーケティング、事業推進などの経営能力が全般的に不足している状況を示す。) & -15.48 (5.70)*** & -6.66 (11.93) & 6.74 (17.80) & -0.94 (0.34)*** & -1.37 (0.63)** & -2.42 (0.91)*** & 660 \\
\addlinespace[0.3em]
\hdashline
\addlinespace[0.3em]
Top & Ability & High Leadership Competence (高いリーダーシップ能力) & Evaluation of the superior leadership qualities and high capabilities of top management. (経営トップの優れたリーダーシップや能力の高さに関する評価) & 23.38 (5.23)*** & 45.04 (9.27)*** & 75.18 (14.15)*** & 2.40 (0.31)*** & 3.40 (0.49)*** & 3.32 (0.73)*** & 748 \\
\addlinespace[0.3em]
\hdashline
\addlinespace[0.3em]
Top & Ability & Inadequate Competence of Top Management (経営トップの能力不足) & This refers to a situation where the president or chairman lacks sufficient expertise and practical experience, resulting in challenges related to managerial skills and business execution capabilities. (社長や会長の専門知識や実務経験が不足し、経営手腕や事業推進力に課題がある状況を示す。) & -20.57 (13.90) & 0.11 (23.74) & 8.49 (33.79) & -0.70 (0.82) & -0.10 (1.26) & -1.56 (1.74) & 177 \\
\addlinespace[0.3em]
\hdashline
\addlinespace[0.3em]
Top & Ability & Lack of Charisma (カリスマ性の欠如) & A negative evaluation indicating that the top management lacks or is deficient in charisma. (経営トップにカリスマ性がない、または不足しているとする否定的評価) & -16.43 (40.72) & -128.96 (52.00)** & -95.99 (63.01) & 3.41 (2.41) & 3.56 (2.77) & 4.73 (3.24) & 43 \\
\addlinespace[0.3em]
\hdashline
\addlinespace[0.3em]
Top & Ability & Lack of Specialized Knowledge Among Top Management (経営トップの専門知識不足) & This refers to the situation in which top management lacks sufficient specialized knowledge and practical skills across diverse fields such as IT, marketing, and technological development. (経営トップ層がITやマーケティング、技術開発など多様な分野での専門知識や実践スキルを十分に持っていない状況を指す。) & -7.99 (8.82) & -10.01 (18.96) & 43.39 (32.33) & -1.69 (0.52)*** & -1.95 (0.99)** & -2.96 (1.66)* & 256 \\
\addlinespace[0.3em]
\hdashline
\addlinespace[0.3em]
Top & Ability & Outstanding Leadership Abilities (優れたリーダーシップ能力) & Positive evaluations regarding the superior leadership abilities of top management, including expertise, decision-making skills, and strategic planning. (経営トップの専門知識や意思決定力、戦略立案などの優れたリーダーシップ能力に関する肯定的評価) & 23.83 (5.06)*** & 41.47 (8.64)*** & 70.18 (13.16)*** & 2.36 (0.30)*** & 3.02 (0.46)*** & 3.01 (0.67)*** & 845 \\
\addlinespace[0.3em]
\hdashline
\addlinespace[0.3em]
Top & Ability & Weaknesses in Leadership (リーダーシップの弱さ) & An evaluation of the deficiencies and challenges in the leadership capabilities of top management. (経営トップのリーダーシップ能力の不足や課題に関する評価) & 15.92 (25.73) & -93.90 (44.18)** & -49.20 (59.00) & 0.24 (1.52) & 1.28 (2.35) & 4.62 (3.03) & 58 \\
\addlinespace[0.3em]
\hdashline
\addlinespace[0.3em]
Top & Attitude & Apathetic Management Attitude (無関心な経営態度) & An attitude in which top management shows a lack of interest or concern toward the workplace and business operations, refraining from active involvement or efforts to implement improvements. (経営トップが現場や事業に対して興味や関心を持たず、積極的な関与や改善を行わない態度。) & 2.47 (14.69) & 19.19 (26.60) & 38.06 (33.19) & -2.10 (0.86)** & -0.25 (1.40) & -1.08 (1.68) & 163 \\
\addlinespace[0.3em]
\hdashline
\addlinespace[0.3em]
Top & Attitude & Attitude of Disregard Towards Employees (従業員軽視姿勢) & An internal disposition wherein top management neglects, undervalues, or dismisses employees without consideration. (経営トップが従業員を顧みず、軽視したり切り捨てる態度を持つ内面的な姿勢) & 9.30 (21.32) & 43.11 (47.53) & 41.89 (51.22) & -1.92 (1.25) & 0.38 (2.48) & -1.73 (2.57) & 79 \\
\addlinespace[0.3em]
\hdashline
\addlinespace[0.3em]
Top & Attitude & Attitudes of Executive Officers and Power Structures (経営役員の態度と権力構造) & Employees' evaluations of the authoritarian and anachronistic attitudes of executive officers and the top-down power structures. (経営役員の権威主義的で時代錯誤な態度やトップダウンの権力構造に関する従業員の評価。) & -4.73 (9.22) & -13.35 (16.26) & -33.94 (23.12) & -0.47 (0.55) & -1.64 (0.86)* & -1.92 (1.19) & 315 \\
\addlinespace[0.3em]
\hdashline
\addlinespace[0.3em]
Top & Attitude & Authoritative Leadership of Top Management (経営トップの権威的リーダーシップ) & This refers to a leadership style in which top management holds strong authority and demonstrates absolute directives and attitudes in a unilateral manner. (経営トップが強い権限を持ち、ワンマンで絶対的な指示や態度を示すリーダーシップの特徴を表す。) & 1.59 (3.44) & 29.03 (6.96)*** & 35.68 (9.86)*** & -0.50 (0.20)** & 0.23 (0.37) & -0.00 (0.51) & 1541 \\
\addlinespace[0.3em]
\hdashline
\addlinespace[0.3em]
Top & Attitude & Autocratic Leadership Style of Top Management (経営トップのワンマン経営姿勢) & The manner in which top management exhibits strong assertiveness and a self-centered attitude, demonstrating a conservative and autocratic management style. (経営トップが強い発言力と自己優先的な態度を持ち、保守的かつワンマンな経営スタイルを示す様子。) & -8.95 (5.21)* & 18.43 (10.59)* & 31.32 (15.91)** & -0.94 (0.31)*** & 0.05 (0.56) & -1.06 (0.82) & 785 \\
\addlinespace[0.3em]
\hdashline
\addlinespace[0.3em]
Top & Attitude & Autocratic Leadership in Top Management (ワンマン経営の経営トップ) & This topic characterizes situations where top management holds strong authority and exhibits an autocratic management style and organizational culture. (経営トップが強い権限を持ち、独断的な経営姿勢や体質を示す特徴を表すトピックです。) & -14.51 (6.43)** & 27.80 (12.62)** & 60.11 (20.67)*** & -0.95 (0.38)** & 0.74 (0.67) & -0.58 (1.06) & 558 \\
\addlinespace[0.3em]
\hdashline
\addlinespace[0.3em]
Top & Attitude & Cautious and Constrained Attitude Toward Transformation (変革への慎重・制約的姿勢) & An attitude wherein top management seeks transformation but remains cautious about taking risks and faces constraints in the implementation of change. (経営トップが変革を求めつつも、リスクを取ることに慎重であったり、変革の実行に制約がある態度。) & -8.25 (34.51) & -118.24 (62.32)* & -377.17 (99.44)*** & 1.27 (2.04) & -1.14 (3.32) & -1.59 (5.13) & 18 \\
\bottomrule
\end{tabular}
\end{adjustbox}
\end{table*}


\begin{table*}[t]
\centering
\scriptsize
\setlength{\tabcolsep}{2pt}
\renewcommand{\arraystretch}{0.92}
\ContinuedFloat
\caption{Relationship between leadership topics and firm outcomes (continued)}
\begin{adjustbox}{max width=\textwidth}
\begin{tabular}{>{\raggedright\arraybackslash}p{0.07\textwidth}>{\raggedright\arraybackslash}p{0.08\textwidth}>{\raggedright\arraybackslash}p{0.18\textwidth}>{\raggedright\arraybackslash}p{0.30\textwidth}>{\centering\arraybackslash}p{0.06\textwidth}>{\centering\arraybackslash}p{0.06\textwidth}>{\centering\arraybackslash}p{0.06\textwidth}>{\centering\arraybackslash}p{0.06\textwidth}>{\centering\arraybackslash}p{0.06\textwidth}>{\centering\arraybackslash}p{0.06\textwidth}>{\centering\arraybackslash}p{0.05\textwidth}}
\toprule
Type. & Char. & Translated Topic Name (Original Topic Name) & Translated Topic Description (Original Topic Description) & \multicolumn{3}{c}{ROA} & \multicolumn{3}{c}{Employees Morale} & N \\
\cmidrule(lr){5-7}\cmidrule(lr){8-10}
 &  &  &  & $t=5$ & $t=10$ & $t=15$ & $t=5$ & $t=10$ & $t=15$ &  \\
\midrule
Top & Attitude & Cautious and Passive Internal Attitudes (慎重・消極的な内面的姿勢) & Evaluation of top management's cautious, passive, and risk-averse attitudes and decision-making tendencies (経営トップが慎重で消極的、リスク回避的な態度や意思決定傾向に関する評価) & 0.62 (9.29) & 16.69 (19.36) & 51.54 (31.54) & 0.13 (0.55) & 0.09 (1.03) & -0.06 (1.61) & 241 \\
\addlinespace[0.3em]
\hdashline
\addlinespace[0.3em]
Top & Attitude & Challenges in the Awareness and Attitudes of Top Management (経営トップの意識と態度の課題) & Issues related to the awareness and attitudes of top management, including a noted lack of understanding of on-site conditions. (経営トップの意識や姿勢に関する問題点や現場理解の不足が指摘されている内容。) & -7.36 (4.98) & 0.98 (8.86) & 5.49 (13.05) & -1.25 (0.29)*** & -1.44 (0.47)*** & -2.75 (0.66)*** & 1104 \\
\addlinespace[0.3em]
\hdashline
\addlinespace[0.3em]
Top & Attitude & Charisma of Top Management (経営トップのカリスマ性) & A topic that represents the intrinsic characteristics of leadership based on the strong charisma, appeal, and personality of executives. (経営者の強いカリスマ性や魅力、人柄に基づくリーダーシップの内面的特徴を表すトピック。) & 8.48 (4.90)* & 22.92 (9.27)** & 40.08 (15.08)*** & 1.52 (0.29)*** & 2.52 (0.49)*** & 2.25 (0.77)*** & 757 \\
\addlinespace[0.3em]
\hdashline
\addlinespace[0.3em]
Top & Attitude & Emphasis on Work-Life Balance (ワークライフバランス重視) & The management's commitment to prioritizing the harmony between work and personal life, as well as respecting employees' private time. (経営トップが仕事と生活の調和やプライベートの尊重を重視する姿勢) & 40.52 (21.68)* & 58.96 (54.60) & 72.22 (59.35) & 3.94 (1.26)*** & 4.30 (2.67) & 2.80 (2.96) & 37 \\
\addlinespace[0.3em]
\hdashline
\addlinespace[0.3em]
Top & Attitude & Employee-Centric Attitude (従業員思いやり姿勢) & An intrinsic mindset wherein top management values employees and demonstrates compassion and empathy towards them. (経営トップが従業員を大切にし、思いやりや寄り添う態度を持つ内面的な姿勢) & 17.54 (7.92)** & 25.14 (13.43)* & 26.54 (19.35) & 1.95 (0.47)*** & 3.27 (0.71)*** & 3.47 (0.99)*** & 402 \\
\addlinespace[0.3em]
\hdashline
\addlinespace[0.3em]
Top & Attitude & Founder’s Emphasis on Ideological Principles (理念重視の創業者姿勢) & The founder holds progressive and egalitarian principles and values the sharing of corporate culture and ideology. (創業者が先進的で平等な理念や考えを持ち、企業文化や理念の共有を大切にしている様子。) & -6.55 (14.14) & 51.69 (36.56) & 95.29 (61.04) & 2.27 (0.83)*** & -0.39 (1.85) & -0.88 (3.14) & 67 \\
\addlinespace[0.3em]
\hdashline
\addlinespace[0.3em]
Top & Attitude & Lack of Sense of Urgency and Passive Attitude (危機感欠如と消極的姿勢) & A situation in which top management lacks a sense of urgency and exhibits a passive stance toward problem-solving and organizational growth. (経営トップが危機感を持たず、問題解決や成長に対して消極的な態度を示している状況) & -13.71 (8.28)* & -4.63 (14.91) & -20.71 (22.45) & -1.80 (0.49)*** & -1.16 (0.78) & -1.99 (1.12)* & 382 \\
\addlinespace[0.3em]
\hdashline
\addlinespace[0.3em]
Top & Attitude & Maintaining a Sense of Urgency and a Proactive Attitude (危機感保持と前向き姿勢) & A situation in which top management maintains a sense of urgency and demonstrates an attitude of striving for improvement and innovation without being complacent with the current state of the company. (経営トップが危機感を持ち、会社の現状に満足せず改善や革新を目指す姿勢を示している状況) & -20.45 (40.11) & -4.95 (69.92) & 35.94 (107.14) & -1.41 (2.37) & 2.11 (3.72) & 2.04 (5.51) & 36 \\
\addlinespace[0.3em]
\hdashline
\addlinespace[0.3em]
Top & Attitude & Management Orientation Focused on Short-Term Profits (短期利益重視の経営姿勢) & A management approach in which top executives prioritize immediate profits and shareholder value, exhibiting an attitude that lacks a long-term perspective and consideration for the customer viewpoint. (経営トップが目先の利益や株主価値を優先し、長期的視点や顧客視点を欠いた態度を示すこと。) & 7.11 (9.45) & 31.81 (22.58) & 38.22 (29.16) & -1.22 (0.56)** & -1.43 (1.17) & -2.87 (1.45)** & 250 \\
\addlinespace[0.3em]
\hdashline
\addlinespace[0.3em]
Top & Attitude & Misogyny and Lack of Understanding (女性蔑視・理解不足) & Reviews indicating that management exhibits male chauvinistic attitudes and demonstrates a negative or insufficient understanding of women's active participation. (経営層に男尊女卑的な態度や女性活躍に対する否定的・理解不足の姿勢が見られる口コミ。) & 25.50 (18.78) & 44.67 (31.03) & 89.12 (45.11)** & 0.93 (1.11) & 1.25 (1.65) & -1.36 (2.32) & 89 \\
\addlinespace[0.3em]
\hdashline
\addlinespace[0.3em]
Top & Attitude & Proactive Approach to Driving Transformation (変革推進の積極姿勢) & The attitude of top management actively engaging in transformation and challenges with the aim of enhancing corporate value. (経営トップが変革やチャレンジに積極的に取り組み、企業価値向上を目指す姿勢。) & 11.42 (10.45) & 11.99 (17.03) & -17.63 (23.73) & -0.40 (0.62) & 0.29 (0.90) & -0.02 (1.22) & 328 \\
\addlinespace[0.3em]
\hdashline
\addlinespace[0.3em]
Top & Attitude & Proactive Internal Attitude (積極的な内面的姿勢) & An evaluation of the management leadership's proactive, challenging, and leadership-driven approach, demonstrating positive decision-making. (経営トップが積極的、挑戦的、リーダーシップを発揮し、前向きな意思決定を行う姿勢に関する評価) & 18.91 (9.86)* & 49.33 (18.06)*** & 69.61 (25.19)*** & 1.57 (0.58)*** & 1.80 (0.95)* & 2.28 (1.26)* & 287 \\
\addlinespace[0.3em]
\hdashline
\addlinespace[0.3em]
Top & Attitude & Promotion of Women's Participation and Advancement (女性活躍推進) & Employee reviews reflecting the management's proactive stance and attitude toward promoting women's active participation and their appointment to managerial positions. (経営層が女性の活躍や管理職登用に積極的である姿勢や態度を示す口コミ。) & 7.80 (26.44) & 9.94 (39.36) & 3.12 (67.30) & 3.73 (1.56)** & 4.40 (2.09)** & 7.73 (3.45)** & 65 \\
\addlinespace[0.3em]
\hdashline
\addlinespace[0.3em]
Top & Attitude & Promotion of Young Talent Challenges (若手挑戦推進) & The management's proactive approach to embracing challenges and opinions from young employees, fostering their development and advancement within the organization. (経営トップが若手の挑戦や意見を積極的に受け入れ、育成や登用を促進する姿勢) & -5.58 (34.50) & 74.63 (85.43) & 59.84 (87.23) & 2.50 (2.04) & 3.89 (4.54) & 2.45 (4.48) & 20 \\
\addlinespace[0.3em]
\hdashline
\addlinespace[0.3em]
Top & Attitude & Responsibility Avoidance and Self-Preservation Orientation (責任回避と保身志向) & An attitude in which top management prioritizes protecting their own position and compensation over assuming responsibility. (経営トップが責任を取らず、自身の地位や報酬を守ることを最優先にする態度。) & -38.58 (14.53)*** & -70.37 (25.78)*** & -98.64 (51.79)* & -3.65 (0.86)*** & -6.02 (1.37)*** & -9.80 (2.66)*** & 106 \\
\addlinespace[0.3em]
\hdashline
\addlinespace[0.3em]
Top & Attitude & Strong Founder Control (強力な創業者支配) & A situation in which founders or founding families possess significant influence and authority, dominating management through autocratic leadership or dictatorial attitudes. (創業者や創業家が強い影響力や権力を持ち、ワンマン経営や独裁的な態度で経営を支配している様子。) & 14.59 (10.04) & 12.26 (19.47) & 39.19 (38.29) & 1.23 (0.59)** & 0.08 (1.04) & -3.98 (1.97)** & 207 \\
\addlinespace[0.3em]
\hdashline
\addlinespace[0.3em]
Top & Attitude & Suppression of Young Employees' Initiatives (若手挑戦抑制) & A managerial attitude in which top executives are reluctant or unwilling to accept or permit the opinions and challenges proposed by young employees. (経営トップが若手の意見や挑戦を採用しにくい、または許さない態度を示す姿勢) & 0.00 (0.00) & 0.00 (0.00) & 0.00 (0.00) & 0.00 (0.00) & 0.00 (0.00) & 0.00 (0.00) & 3 \\
\addlinespace[0.3em]
\hdashline
\addlinespace[0.3em]
Top & Attitude & The Attitude of a Autocratic Management Leader (ワンマン経営トップの姿勢) & An attitude characterized by the management leader exerting strong influence and exhibiting a tendency toward an autocratic, top-down management style. (経営トップが強い影響力を持ち、独断的でトップダウンの経営スタイルを取る傾向を示す姿勢。) & -21.80 (7.03)*** & 19.88 (13.40) & 59.77 (25.54)** & -1.32 (0.42)*** & 0.26 (0.71) & -1.66 (1.31) & 438 \\
\bottomrule
\end{tabular}
\end{adjustbox}
\end{table*}


\begin{table*}[t]
\centering
\scriptsize
\setlength{\tabcolsep}{2pt}
\renewcommand{\arraystretch}{0.92}
\ContinuedFloat
\caption{Relationship between leadership topics and firm outcomes (continued)}
\begin{adjustbox}{max width=\textwidth}
\begin{tabular}{>{\raggedright\arraybackslash}p{0.07\textwidth}>{\raggedright\arraybackslash}p{0.08\textwidth}>{\raggedright\arraybackslash}p{0.18\textwidth}>{\raggedright\arraybackslash}p{0.30\textwidth}>{\centering\arraybackslash}p{0.06\textwidth}>{\centering\arraybackslash}p{0.06\textwidth}>{\centering\arraybackslash}p{0.06\textwidth}>{\centering\arraybackslash}p{0.06\textwidth}>{\centering\arraybackslash}p{0.06\textwidth}>{\centering\arraybackslash}p{0.06\textwidth}>{\centering\arraybackslash}p{0.05\textwidth}}
\toprule
Type. & Char. & Translated Topic Name (Original Topic Name) & Translated Topic Description (Original Topic Description) & \multicolumn{3}{c}{ROA} & \multicolumn{3}{c}{Employees Morale} & N \\
\cmidrule(lr){5-7}\cmidrule(lr){8-10}
 &  &  &  & $t=5$ & $t=10$ & $t=15$ & $t=5$ & $t=10$ & $t=15$ &  \\
\midrule
Top & Attitude & Top Management's Innovation Orientation and Philosophy (経営トップの革新志向と理念) & A topic that reflects how the president's challenging and future-oriented attitude, along with the corporate philosophy, strongly influences management. (社長の挑戦的で未来志向な姿勢と企業理念が経営に強く反映されている様子を表すトピック。) & 12.57 (6.10)** & 11.23 (12.05) & 21.04 (17.42) & 1.05 (0.36)*** & 1.48 (0.64)** & 0.77 (0.89) & 649 \\
\addlinespace[0.3em]
\hdashline
\addlinespace[0.3em]
Top & Attitude & Work Priority and Emphasis on Long Working Hours (仕事優先・長時間労働重視) & A management approach in which top executives prioritize work as the central aspect of life and regard long working hours and personal sacrifice as virtues. (経営トップが仕事を生活の中心とし、長時間労働や犠牲を美徳とする姿勢) & 44.29 (20.90)** & 82.24 (27.89)*** & 76.80 (28.50)*** & 4.10 (1.23)*** & 6.06 (1.48)*** & 4.93 (1.46)*** & 62 \\
\addlinespace[0.3em]
\hdashline
\addlinespace[0.3em]
Top & Behavior & Acceptance of Overtime and Holiday Work (残業・休日出勤容認) & Policies and actions whereby top management permits or mandates overtime and holiday work, thereby promoting a return to office attendance and extended working hours. (経営トップが残業や休日出勤を容認・強要し、出社回帰や長時間労働を推進する方針や行動。) & 12.26 (14.51) & 46.62 (25.43)* & 37.21 (29.95) & 0.14 (0.86) & 1.85 (1.35) & 1.20 (1.54) & 136 \\
\addlinespace[0.3em]
\hdashline
\addlinespace[0.3em]
Top & Behavior & Authoritarian Autocratic Management (強権的ワンマン経営) & A management style in which the founding owner or chairman holds strong authority, makes decisions in a top-down manner, and exercises strict control over employees. (創業オーナーや会長が強い権限を持ち、トップダウンで意思決定を行い、従業員を強く統制する経営スタイル。) & 16.55 (8.81)* & 29.74 (16.05)* & 45.34 (27.73) & -0.35 (0.52) & -0.79 (0.85) & -2.97 (1.42)** & 338 \\
\addlinespace[0.3em]
\hdashline
\addlinespace[0.3em]
Top & Behavior & Authoritarian Decision-Making (強権的な意思決定) & Reviews concerning behavioral patterns in which executives possess absolute authority and forcibly enforce instructions and decisions. (役員が絶対的な権限を持ち、指示や決定を強制的に押し通す行動パターンに関する口コミ。) & -17.87 (9.38)* & -27.70 (14.55)* & -55.44 (22.10)** & -1.52 (0.55)*** & -3.41 (0.77)*** & -3.13 (1.13)*** & 346 \\
\addlinespace[0.3em]
\hdashline
\addlinespace[0.3em]
Top & Behavior & Authoritarian Top-Down Management (強権的トップダウン経営) & A leadership style in which top management centralizes authority and unilaterally determines and implements directives and policies. (経営トップが権限を集中させ、指示や方針を一方的に決定し実行するリーダーシップスタイル。) & 8.88 (6.91) & 28.09 (11.64)** & 27.69 (16.33)* & 0.11 (0.41) & 0.70 (0.62) & 0.15 (0.83) & 655 \\
\addlinespace[0.3em]
\hdashline
\addlinespace[0.3em]
Top & Behavior & Autocratic Decision-Making (独断的意思決定) & A behavioral pattern in which top management unilaterally determines policies and decisions, independently setting the direction of the company. (経営トップが一存で方針や意思決定を行い、独断的に企業の方向性を決める行動パターン) & 16.31 (35.99) & -9.41 (54.77) & 23.49 (93.75) & -1.12 (2.13) & -1.46 (2.91) & -5.73 (4.82) & 44 \\
\addlinespace[0.3em]
\hdashline
\addlinespace[0.3em]
Top & Behavior & Autocratic Decision-Making (独断的意思決定) & A behavioral pattern in which senior management prioritizes their own intentions and ideas, forcibly making decisions. (上層部が自分たちの意向や考えを優先し、強引に意思決定を行う行動パターン) & -32.13 (24.04) & -82.10 (43.84)* & -58.78 (54.97) & -3.90 (1.41)*** & -4.70 (2.33)** & -6.30 (2.82)** & 58 \\
\addlinespace[0.3em]
\hdashline
\addlinespace[0.3em]
Top & Behavior & Autocratic Exercise of Power by Top Management (経営トップの独裁的権力行使) & This refers to the situation in which the chairman or president holds strong decision-making authority and influence, privatizes management, and operates the organization in an autocratic manner. (会長や社長が強い決定権と影響力を持ち、経営を私物化し独裁的に運営する様子を示す。) & -6.38 (12.04) & -9.38 (22.43) & -37.67 (27.19) & -0.60 (0.71) & -2.29 (1.19)* & -3.68 (1.40)*** & 197 \\
\addlinespace[0.3em]
\hdashline
\addlinespace[0.3em]
Top & Behavior & Autocratic Management (ワンマン経営) & A management style in which the president makes decisions unilaterally and centralizes all authority. (社長が独断的に意思決定を行い、全ての権限を集中させる経営スタイルを指す。) & 10.32 (4.99)** & 19.28 (9.35)** & 42.34 (14.68)*** & 0.07 (0.29) & 0.42 (0.49) & 0.10 (0.75) & 881 \\
\addlinespace[0.3em]
\hdashline
\addlinespace[0.3em]
Top & Behavior & Autocratic Management System (ワンマン経営体制) & A management style in which the top executive makes decisions unilaterally and surrounds themselves with yes-men. (経営トップが独断的に意思決定を行い、イエスマンを周囲に固める経営スタイルを指す。) & -0.16 (10.44) & 8.02 (19.03) & 9.80 (26.51) & 0.89 (0.62) & 0.10 (1.01) & -1.76 (1.36) & 248 \\
\addlinespace[0.3em]
\hdashline
\addlinespace[0.3em]
Top & Behavior & Behavioral Patterns of Autocratic Executive Leadership (独裁的な経営トップの行動パターン) & Char.s of executive leaders who disregard the opinions of employees and frontline staff, refuse dissent, and make self-centered decisions. (社員や現場の意見を無視し、反論を許さず自己中心的な意思決定を行う経営トップの特徴を示す。) & -2.02 (6.24) & 18.69 (11.21)* & 19.95 (16.28) & -1.49 (0.37)*** & -0.73 (0.60) & -1.31 (0.84) & 595 \\
\addlinespace[0.3em]
\hdashline
\addlinespace[0.3em]
Top & Behavior & Behavioral Patterns of Top Management in Autocratic Leadership (ワンマン経営の経営トップ行動) & This refers to the behavioral pattern in which top management makes decisions in a top-down manner and repeatedly issues autocratic directives and judgments. (経営トップがトップダウンで意思決定を行い、ワンマン的な指示や判断を繰り返す行動パターンを示す。) & 9.84 (7.46) & -3.16 (13.64) & 14.73 (25.24) & -0.37 (0.44) & -0.01 (0.72) & -1.35 (1.28) & 408 \\
\addlinespace[0.3em]
\hdashline
\addlinespace[0.3em]
Top & Behavior & Consensus-Based Decision Making (合議的意思決定) & A behavioral pattern in which top management advances policies and decisions by obtaining the consensus of all senior executives. (経営トップが上層部全員の合意を得て方針や意思決定を進める行動パターン) & 15.90 (31.18) & 86.58 (225.62) & 67.84 (230.15) & 3.29 (1.84)* & -2.29 (12.00) & -5.07 (11.83) & 7 \\
\addlinespace[0.3em]
\hdashline
\addlinespace[0.3em]
Top & Behavior & Decision-Making by the Representative Director and Management Policies (代表取締役の意思決定と経営方針) & This topic illustrates how the representative director makes top-down decisions that strongly influence management policies and organizational operations. (代表取締役がトップダウンで意思決定を行い、経営方針や組織運営に強い影響を与える様子を示すトピック。) & 10.93 (21.08) & 43.79 (32.84) & 71.83 (44.58) & 3.13 (1.25)** & 2.78 (1.75) & 3.43 (2.29) & 66 \\
\addlinespace[0.3em]
\hdashline
\addlinespace[0.3em]
Top & Behavior & Delayed Decision-Making and Actions (遅い意思決定と行動) & A collection of texts pointing out the slow or delayed decision-making and actions of top management. (経営トップの意思決定や行動が遅い、遅延していることを指摘する文章群) & 5.61 (26.74) & -73.69 (43.99)* & -161.77 (64.35)** & 0.65 (1.58) & -1.97 (2.34) & -3.97 (3.31) & 34 \\
\addlinespace[0.3em]
\hdashline
\addlinespace[0.3em]
Top & Behavior & Denial of Female Appointment (女性登用否定) & Cases in which top management and senior executives exhibit negative or passive attitudes toward the appointment of women to managerial positions and the promotion of their active participation. (経営トップや上層部が女性の管理職登用や活躍推進に否定的または消極的な態度を示す事例) & 97.47 (81.21) & 78.51 (81.97) & -1097.85 (680.62) & 3.23 (4.80) & 2.03 (4.36) & 5.62 (35.00) & 6 \\
\addlinespace[0.3em]
\hdashline
\addlinespace[0.3em]
Top & Behavior & Executive Favoritism in Personnel Management (経営トップのえこひいき人事) & Reviews indicating a tendency for top management and executives to determine personnel evaluations and promotions based on personal preferences and biases. (経営トップや役員が個人的な好みや好き嫌いで人事評価や昇格を決定する傾向を示す口コミ。) & -15.93 (39.42) & -31.19 (55.81) & -68.82 (177.83) & -5.59 (2.33)** & -4.02 (2.97) & -3.54 (9.14) & 25 \\
\bottomrule
\end{tabular}
\end{adjustbox}
\end{table*}


\begin{table*}[t]
\centering
\scriptsize
\setlength{\tabcolsep}{2pt}
\renewcommand{\arraystretch}{0.92}
\ContinuedFloat
\caption{Relationship between leadership topics and firm outcomes (continued)}
\begin{adjustbox}{max width=\textwidth}
\begin{tabular}{>{\raggedright\arraybackslash}p{0.07\textwidth}>{\raggedright\arraybackslash}p{0.08\textwidth}>{\raggedright\arraybackslash}p{0.18\textwidth}>{\raggedright\arraybackslash}p{0.30\textwidth}>{\centering\arraybackslash}p{0.06\textwidth}>{\centering\arraybackslash}p{0.06\textwidth}>{\centering\arraybackslash}p{0.06\textwidth}>{\centering\arraybackslash}p{0.06\textwidth}>{\centering\arraybackslash}p{0.06\textwidth}>{\centering\arraybackslash}p{0.06\textwidth}>{\centering\arraybackslash}p{0.05\textwidth}}
\toprule
Type. & Char. & Translated Topic Name (Original Topic Name) & Translated Topic Description (Original Topic Description) & \multicolumn{3}{c}{ROA} & \multicolumn{3}{c}{Employees Morale} & N \\
\cmidrule(lr){5-7}\cmidrule(lr){8-10}
 &  &  &  & $t=5$ & $t=10$ & $t=15$ & $t=5$ & $t=10$ & $t=15$ &  \\
\midrule
Top & Behavior & Flexible Decision-Making and Consideration (柔軟な意思決定と配慮) & Reviews concerning behavioral patterns in which executives make decisions and give instructions while listening to subordinates' opinions and demonstrating consideration. (役員が部下の意見を聞き入れたり、配慮を行いながら意思決定や指示を行う行動パターンに関する口コミ。) & -0.49 (15.85) & 21.84 (37.37) & 12.05 (46.91) & 0.51 (0.94) & -0.59 (1.99) & 0.56 (2.41) & 91 \\
\addlinespace[0.3em]
\hdashline
\addlinespace[0.3em]
Top & Behavior & Frequent Policy Reversals by Top Management (経営トップの頻繁な方針転換) & Reviews concerning the behavioral pattern of top management frequently altering the direction of organizations or projects based on moods or subjective judgments. (経営トップが気分や判断で組織やプロジェクトの方向性を頻繁に変更する行動パターンに関する口コミ。) & -11.34 (6.90) & -18.12 (11.90) & -15.98 (19.13) & -1.35 (0.41)*** & -1.64 (0.63)*** & -2.33 (0.97)** & 718 \\
\addlinespace[0.3em]
\hdashline
\addlinespace[0.3em]
Top & Behavior & Indecisive Decision-Making (優柔不断な意思決定) & Behavioral patterns characterized by executives and senior management delaying decisions or frequently reversing decisions, resulting in inconsistency. (役員や上層部が意思決定を遅らせたり、決定を頻繁に覆すなど一貫性のない行動パターン) & -43.40 (107.00) & -54.41 (107.94) & -60.13 (110.11) & -0.85 (5.86) & -0.94 (5.32) & -1.95 (5.25) & 3 \\
\addlinespace[0.3em]
\hdashline
\addlinespace[0.3em]
Top & Behavior & Influence Behaviors That Prompt Subordinates to Gauge the Reactions of Top Management (経営トップの顔色を伺わせる影響力行動) & This topic examines proactive influence behaviors whereby top management induces subordinates and executives to monitor their reactions and seek to appease their moods. (経営トップが部下や役員に対して顔色を伺わせる、機嫌を取らせるなど影響力を行使する主体的な行動を示す。) & 3.52 (13.21) & 5.40 (24.17) & 29.14 (39.08) & -0.64 (0.78) & 0.63 (1.28) & -0.61 (2.01) & 124 \\
\addlinespace[0.3em]
\hdashline
\addlinespace[0.3em]
Top & Behavior & Irresponsible Management Decisions (無責任な経営判断) & Issues arising when top management disregards input from frontline employees, neglects to take responsibility, leaves problems unaddressed, and fails to make appropriate decisions. (経営トップが現場の声を無視し、責任を取らずに問題を放置したり、適切な判断をしない問題点。) & -17.60 (5.19)*** & -3.61 (8.81) & -4.12 (12.81) & -1.41 (0.31)*** & -2.40 (0.46)*** & -3.21 (0.65)*** & 1057 \\
\addlinespace[0.3em]
\hdashline
\addlinespace[0.3em]
Top & Behavior & Lack of Measures and Insufficient Communication by Senior Management (上層部の無策と情報伝達不足) & This describes a situation in which senior management fails to grasp the conditions on the ground, neglects decision-making and personnel responses, and lacks adequate information sharing. (上層部が現場の状況を把握せず、意思決定や人員対応を怠り情報共有が不足している様子を示す。) & -25.59 (7.81)*** & -17.96 (15.89) & -24.43 (20.57) & -2.77 (0.46)*** & -3.77 (0.84)*** & -5.37 (1.05)*** & 436 \\
\addlinespace[0.3em]
\hdashline
\addlinespace[0.3em]
Top & Behavior & Lack of Reform Capability and Stagnation (改革力の欠如と停滞) & Situations and behaviors characterized by the absence of reform capability among top management, resulting in maintenance of the status quo, stagnation, and misdirection. (経営トップが改革力を欠き、現状維持や停滞、方向性の誤りが見られる行動や状況。) & 4.46 (41.06) & 13.13 (53.52) & 132.75 (115.50) & -0.53 (2.39) & -2.22 (2.78) & 2.62 (5.38) & 43 \\
\addlinespace[0.3em]
\hdashline
\addlinespace[0.3em]
Top & Behavior & Leadership-Driven Management (リーダーシップ発揮型経営) & A management style in which founding owners or chairpersons exercise strong authority while demonstrating leadership, guiding the organization through the establishment and dissemination of corporate philosophy and behavioral principles. (創業オーナーや会長が強い権限を持ちながらも、リーダーシップを発揮し、企業理念や行動原則の構築・浸透を通じて組織を導く経営スタイル。) & 21.73 (13.73) & 44.72 (35.20) & 118.93 (54.73)** & -0.39 (0.81) & 1.60 (1.87) & 1.58 (2.82) & 68 \\
\addlinespace[0.3em]
\hdashline
\addlinespace[0.3em]
Top & Behavior & Limited or Absent Communication (限定的または不在のコミュニケーション) & A behavioral pattern characterized by minimal interaction between the company president and employees, including the president rarely visiting the workplace. (社長が従業員とほとんど関わらず、現場に来ないなどコミュニケーションが限定的な行動様式。) & -4.58 (11.73) & 18.91 (21.24) & 25.05 (32.94) & -0.48 (0.69) & -1.04 (1.12) & -1.04 (1.69) & 202 \\
\addlinespace[0.3em]
\hdashline
\addlinespace[0.3em]
Top & Behavior & Negative Aspects of Top-Down Management (トップダウンの否定的側面) & Reviews expressing critical or excessive concerns regarding decision-making conducted in a top-down manner by senior management. (経営トップがトップダウンで意思決定を行うことに対して否定的・過度と感じる口コミ) & 12.19 (73.70) & -6.22 (74.39) & 58.44 (100.07) & -1.53 (4.36) & -2.81 (3.95) & -5.08 (5.14) & 18 \\
\addlinespace[0.3em]
\hdashline
\addlinespace[0.3em]
Top & Behavior & New Business Development (新規事業推進) & The actions and attitudes of top management actively engaging in and promoting new business initiatives and innovation. (経営トップが新規事業やイノベーションに積極的に取り組み、推進している行動や姿勢) & -6.17 (10.28) & -10.27 (17.49) & -38.84 (30.99) & 1.05 (0.61)* & -0.00 (0.93) & -2.72 (1.58)* & 271 \\
\addlinespace[0.3em]
\hdashline
\addlinespace[0.3em]
Top & Behavior & Organizational Transformation and Innovation Promotion by Top Management (経営トップの組織変革とイノベーション推進) & Refers to innovative initiatives led by top management, including organizational reform, new business development, and digital transformation. (経営トップ層が主導する組織改革や新規事業開拓、デジタルトランスフォーメーションなどの革新的取り組みを指す。) & 20.68 (16.88) & -4.79 (46.59) & 2.34 (58.78) & 0.21 (1.00) & -5.54 (2.45)** & -5.55 (2.97)* & 70 \\
\addlinespace[0.3em]
\hdashline
\addlinespace[0.3em]
Top & Behavior & Passive Deferential Behavior of Top Management (経営トップの受動的忖度行動) & Top management exhibits passive behavior characterized by a lack of initiative, such as gauging the reactions of senior executives or the president and waiting for approval or instructions. (経営トップが上層部や社長の顔色を伺い、承認や指示を待つなど主体性のない受動的な行動を示す。) & 27.12 (20.59) & 28.69 (33.62) & 38.38 (39.77) & -0.07 (1.22) & -1.74 (1.79) & -3.06 (2.04) & 101 \\
\addlinespace[0.3em]
\hdashline
\addlinespace[0.3em]
Top & Behavior & Passive and Delayed Decision-Making (消極的・遅延する意思決定) & Char.s of executive decision-making that are slow, passive, or lack corresponding concrete actions. (経営トップの意思決定が遅い、消極的、または実際の行動が伴わない特徴。) & 13.33 (17.94) & 2.37 (27.21) & -7.64 (45.11) & -0.96 (1.06) & -0.41 (1.45) & 0.00 (2.32) & 92 \\
\addlinespace[0.3em]
\hdashline
\addlinespace[0.3em]
Top & Behavior & Policy Changes by Authoritarian Executive Leadership (独裁的な経営トップの方針変更) & This pattern of behavior is characterized by executive leaders frequently and arbitrarily making significant changes to organizational policies and structures based on a single remark or directive. (経営トップが一言や一声で組織の方針や体制を頻繁かつ思いつきで大きく変える行動パターンを示す。) & 2.26 (4.85) & 14.54 (9.03) & 14.89 (15.62) & -0.54 (0.29)* & 0.04 (0.48) & -1.06 (0.80) & 943 \\
\addlinespace[0.3em]
\hdashline
\addlinespace[0.3em]
Top & Behavior & Preferential Treatment by Top Management (経営トップのえこひいき優遇) & A tendency wherein top management favors certain employees, granting them special treatment through promotions, salary increases, and the assignment of positions. (経営トップが気に入った社員を優遇し、昇進や昇給、役職付与などで特別扱いする傾向。) & -1.97 (8.46) & 14.99 (15.51) & 0.82 (29.91) & -0.08 (0.50) & -1.93 (0.82)** & -3.41 (1.53)** & 362 \\
\addlinespace[0.3em]
\hdashline
\addlinespace[0.3em]
Top & Behavior & Proactive Decision-Making and Leadership (積極的な意思決定とリーダーシップ) & Char.s whereby top management demonstrates strong intent and leadership, actively promoting decision-making and actions. (経営トップが強い意向やリーダーシップを示し、積極的に意思決定や行動を推進する特徴。) & -1.32 (6.39) & 18.75 (11.46) & 71.49 (16.36)*** & 2.09 (0.37)*** & 2.74 (0.60)*** & 3.46 (0.82)*** & 742 \\
\addlinespace[0.3em]
\hdashline
\addlinespace[0.3em]
Top & Behavior & Proactive Dialogue and Communication (積極的な対話とコミュニケーション) & A behavioral approach in which the company president engages in direct dialogue with employees and actively fosters communication. (社長が従業員と直接対話し、積極的にコミュニケーションを取る行動様式。) & 18.24 (4.37)*** & 33.70 (8.66)*** & 34.08 (12.28)*** & 1.98 (0.26)*** & 3.04 (0.46)*** & 4.50 (0.62)*** & 1147 \\
\bottomrule
\end{tabular}
\end{adjustbox}
\end{table*}


\begin{table*}[t]
\centering
\scriptsize
\setlength{\tabcolsep}{2pt}
\renewcommand{\arraystretch}{0.92}
\ContinuedFloat
\caption{Relationship between leadership topics and firm outcomes (continued)}
\begin{adjustbox}{max width=\textwidth}
\begin{tabular}{>{\raggedright\arraybackslash}p{0.07\textwidth}>{\raggedright\arraybackslash}p{0.08\textwidth}>{\raggedright\arraybackslash}p{0.18\textwidth}>{\raggedright\arraybackslash}p{0.30\textwidth}>{\centering\arraybackslash}p{0.06\textwidth}>{\centering\arraybackslash}p{0.06\textwidth}>{\centering\arraybackslash}p{0.06\textwidth}>{\centering\arraybackslash}p{0.06\textwidth}>{\centering\arraybackslash}p{0.06\textwidth}>{\centering\arraybackslash}p{0.06\textwidth}>{\centering\arraybackslash}p{0.05\textwidth}}
\toprule
Type. & Char. & Translated Topic Name (Original Topic Name) & Translated Topic Description (Original Topic Description) & \multicolumn{3}{c}{ROA} & \multicolumn{3}{c}{Employees Morale} & N \\
\cmidrule(lr){5-7}\cmidrule(lr){8-10}
 &  &  &  & $t=5$ & $t=10$ & $t=15$ & $t=5$ & $t=10$ & $t=15$ &  \\
\midrule
Top & Behavior & Proactive Organizational Reform (積極的な組織改革) & Cases in which top management actively promotes organizational reform and transformation of corporate culture, implementing concrete initiatives such as work style reform and improvements in communication openness. (経営トップが積極的に組織改革や企業文化の変革を推進し、働き方改革や風通しの改善など具体的な取り組みを行っている事例。) & 1.15 (11.89) & -3.41 (22.54) & -8.37 (29.50) & 0.32 (0.70) & 1.69 (1.18) & -0.46 (1.48) & 220 \\
\addlinespace[0.3em]
\hdashline
\addlinespace[0.3em]
Top & Behavior & Proactive Promotion of Reform (積極的な改革推進) & Actions whereby top management takes the initiative to actively drive transformation and business restructuring, aiming to foster organizational growth and enhance competitive strength. (経営トップが主体的に変革や事業見直しを積極的に推進し、組織の成長や競争力強化を目指す行動。) & -11.80 (5.57)** & -12.43 (10.64) & -4.84 (13.83) & -0.05 (0.33) & 0.19 (0.56) & 0.26 (0.70) & 913 \\
\addlinespace[0.3em]
\hdashline
\addlinespace[0.3em]
Top & Behavior & Proactive Recruitment of External Talent (外部人材積極登用) & Actions taken by top management to actively recruit external and foreign executives in order to strengthen the organization and promote globalization. (経営トップが外部や外国人幹部を積極的に採用し、組織の強化やグローバル化を推進する行動。) & -40.61 (24.01)* & 68.51 (52.75) & 131.46 (67.81)* & -1.00 (1.42) & 0.06 (2.79) & -2.33 (3.45) & 68 \\
\addlinespace[0.3em]
\hdashline
\addlinespace[0.3em]
Top & Behavior & Proactive Top-Down Leadership (積極的なトップダウン) & Reviews concerning the manner in which senior management actively issues directives and makes decisions through a top-down approach. (経営トップが積極的にトップダウンで指示や意思決定を行う様子に関する口コミ) & 3.07 (4.49) & 9.70 (8.11) & 19.64 (12.32) & -0.07 (0.27) & 0.64 (0.43) & -0.17 (0.63) & 1134 \\
\addlinespace[0.3em]
\hdashline
\addlinespace[0.3em]
Top & Behavior & Proactive but Problematic Managerial Behavior (積極的だが問題のある経営行動) & Cases in which top management takes proactive actions, but those actions involve issues such as vanity, deception, falsification, or unsustainable continuation. (経営トップが積極的に行動しているが、その行動が虚栄心や嘘、偽装、無理な継続など問題を含むケース。) & 2.87 (10.70) & 7.06 (18.43) & 7.60 (27.88) & -2.07 (0.63)*** & -1.83 (0.98)* & -2.39 (1.43)* & 216 \\
\addlinespace[0.3em]
\hdashline
\addlinespace[0.3em]
Top & Behavior & Promotion of Overtime Reduction (残業削減推進) & Policies and actions whereby top management actively promotes the reduction of overtime hours and the reform of work styles. (経営トップが残業時間の削減や働き方改革を積極的に推進する方針や行動。) & 7.46 (22.39) & 27.45 (39.16) & 70.77 (57.44) & 1.85 (1.32) & 0.89 (2.08) & 2.88 (2.95) & 74 \\
\addlinespace[0.3em]
\hdashline
\addlinespace[0.3em]
Top & Behavior & Promotion of Women's Active Participation (女性活躍推進) & Initiatives in which top management actively promotes the appointment of female managers and the creation of a work environment conducive to their success. (経営トップが女性管理職の登用や働きやすい環境づくりを積極的に推進する取り組み) & 17.74 (14.00) & 53.96 (37.75) & 76.11 (47.30) & 2.20 (0.83)*** & 4.24 (2.01)** & 3.66 (2.43) & 89 \\
\addlinespace[0.3em]
\hdashline
\addlinespace[0.3em]
Top & Behavior & Rapid Decision-Making and Action (迅速な意思決定と行動) & A collection of statements characterizing the ability of top management to make decisions with a sense of urgency and swiftly mobilize the organization. (経営トップがスピード感を持って意思決定し、素早く組織を動かす特徴を表す文章群) & -5.69 (5.57) & 23.39 (12.35)* & 36.59 (17.51)** & 1.42 (0.33)*** & 0.97 (0.65) & 0.30 (0.90) & 527 \\
\addlinespace[0.3em]
\hdashline
\addlinespace[0.3em]
Top & Behavior & Rapid Top-Down Decision Making (迅速なトップダウン意思決定) & A behavioral approach in which senior management makes decisions in a top-down manner with a strong sense of urgency, enabling swift and decisive action to drive business initiatives. (経営トップがスピード感を持ってトップダウンで意思決定し、即断即決で事業を推進する行動様式。) & -58.38 (24.26)** & 33.43 (38.86) & 48.25 (52.29) & 1.61 (1.41) & 3.18 (1.99) & 1.68 (2.69) & 62 \\
\addlinespace[0.3em]
\hdashline
\addlinespace[0.3em]
Top & Behavior & Rejection of Utilizing External Human Resources (外部人材活用否定) & Actions by top management that deny the opinions and recruitment of external or foreign executives, emphasizing internal personnel and the existing organizational structure. (経営トップが外部や外国人幹部の意見や採用を否定し、内部人材や既存体制を重視する行動。) & -38.35 (100.63) & -44.03 (101.51) & -53.71 (103.56) & -2.14 (5.95) & -5.09 (5.40) & -5.59 (5.32) & 7 \\
\addlinespace[0.3em]
\hdashline
\addlinespace[0.3em]
Top & Behavior & Reluctance Toward New Business Ventures (新規事業消極姿勢) & Behaviors and attitudes exhibited by top management that demonstrate reluctance or a lack of initiative in promoting new business ventures or innovative initiatives. (経営トップが新規事業や新しい取り組みに対して消極的、または推進しない態度を示している行動や姿勢) & -14.64 (32.08) & 36.42 (69.91) & 55.45 (87.49) & -1.89 (1.90) & -3.57 (3.72) & -4.31 (4.50) & 37 \\
\addlinespace[0.3em]
\hdashline
\addlinespace[0.3em]
Top & Behavior & Restrictions on Approval Authority (決裁権限の制限) & Although top management holds approval authority, there are restrictive behaviors such as withholding approval, delayed approvals, or refusal to approve certain cases. (経営トップが決裁権限を持つものの、承認を出さない、承認が遅い、または一部案件を承認しないなど制限的な様子) & -20.57 (20.45) & 48.38 (28.87)* & 23.84 (43.37) & -0.79 (1.21) & -0.22 (1.54) & 1.01 (2.23) & 72 \\
\addlinespace[0.3em]
\hdashline
\addlinespace[0.3em]
Top & Behavior & Retaliation and Disadvantage by Top Management Favoritism (経営トップのえこひいき報復・不利益) & The tendency of top management to impose disadvantages such as demotion, dismissal, or reassignment on employees they disfavor. (経営トップが気に入らない社員に対して降格、解雇、異動などの不利益を与える傾向。) & 15.84 (19.81) & 30.95 (45.27) & 23.80 (80.72) & -2.23 (1.17)* & -1.62 (2.41) & -5.82 (4.15) & 73 \\
\addlinespace[0.3em]
\hdashline
\addlinespace[0.3em]
Top & Behavior & Stagnation and Passive Attitudes Toward Reform (改革の停滞・消極的姿勢) & A situation in which organizational reform is either stagnant or approached with reluctance, characterized by management's failure to propose fundamental reforms or to implement improvement measures in a top-down manner. (経営トップが抜本的な改革を打ち出せていなかったり、トップダウン的に改善策を行わないなど、組織改革に消極的または停滞している状況。) & -28.78 (45.34) & 6.77 (60.28) & -30.20 (90.39) & -3.47 (2.68) & -4.15 (3.20) & -2.46 (4.65) & 22 \\
\addlinespace[0.3em]
\hdashline
\addlinespace[0.3em]
Top & Behavior & Strong Decision-Making Authority (強い決裁権限) & The situation in which top management holds the ultimate decision-making authority over all matters and exercises significant influence. (経営トップが全ての案件に対して最終決裁権限を持ち、強い影響力を行使している様子) & 16.62 (6.20)*** & 26.74 (9.81)*** & 44.52 (15.27)*** & 0.28 (0.36) & 0.83 (0.52) & 0.94 (0.79) & 706 \\
\addlinespace[0.3em]
\hdashline
\addlinespace[0.3em]
Top & Behavior & Top-Down Directives from Senior Management (経営トップのトップダウン指示) & A behavioral pattern in which senior management exercises strong top-down decision-making and directives, thereby influencing the organizational direction. (経営トップが強力なトップダウンで意思決定や指示を行い、組織の方向性を左右する行動様式。) & 2.18 (2.93) & 11.96 (4.96)** & 18.98 (7.50)** & -0.13 (0.17) & 0.26 (0.26) & -0.28 (0.39) & 2412 \\
\addlinespace[0.3em]
\hdashline
\addlinespace[0.3em]
Top & Behavior & Top-Down Management Directive (トップダウン型経営指示) & A leadership style in which top management issues unilateral directives and controls the organization in a manner akin to a military structure. (経営トップが一方的に指示を出し、組織を軍隊のように統制するリーダーシップスタイル。) & 18.47 (24.41) & 53.90 (53.45) & 120.78 (77.65) & -1.13 (1.44) & 2.30 (2.84) & -2.80 (3.99) & 45 \\
\addlinespace[0.3em]
\hdashline
\addlinespace[0.3em]
Top & Behavior & Top-Down Management Directives (トップダウン型経営指示) & A behavioral pattern in which top management unilaterally issues instructions and orders, providing guidance to the operational level from a hierarchical superior position. (経営トップが一方的に指示や命令を出し、現場に対して上からの指導を行う行動様式を指す。) & 11.61 (29.51) & -4.28 (53.88) & -114.22 (92.46) & -0.79 (1.74) & 0.93 (2.86) & -0.65 (4.75) & 36 \\
\bottomrule
\end{tabular}
\end{adjustbox}
\end{table*}


\begin{table*}[t]
\centering
\scriptsize
\setlength{\tabcolsep}{2pt}
\renewcommand{\arraystretch}{0.92}
\ContinuedFloat
\caption{Relationship between leadership topics and firm outcomes (continued)}
\begin{adjustbox}{max width=\textwidth}
\begin{tabular}{>{\raggedright\arraybackslash}p{0.07\textwidth}>{\raggedright\arraybackslash}p{0.08\textwidth}>{\raggedright\arraybackslash}p{0.18\textwidth}>{\raggedright\arraybackslash}p{0.30\textwidth}>{\centering\arraybackslash}p{0.06\textwidth}>{\centering\arraybackslash}p{0.06\textwidth}>{\centering\arraybackslash}p{0.06\textwidth}>{\centering\arraybackslash}p{0.06\textwidth}>{\centering\arraybackslash}p{0.06\textwidth}>{\centering\arraybackslash}p{0.06\textwidth}>{\centering\arraybackslash}p{0.05\textwidth}}
\toprule
Type. & Char. & Translated Topic Name (Original Topic Name) & Translated Topic Description (Original Topic Description) & \multicolumn{3}{c}{ROA} & \multicolumn{3}{c}{Employees Morale} & N \\
\cmidrule(lr){5-7}\cmidrule(lr){8-10}
 &  &  &  & $t=5$ & $t=10$ & $t=15$ & $t=5$ & $t=10$ & $t=15$ &  \\
\midrule
Top & Behavior & Top-Down Management Policy (トップダウン経営方針) & A management style in which the top executives unilaterally determine policies and measures, directing and controlling the organization. (経営トップが一方的に方針や施策を決定し、組織を指示・管理する経営スタイル。) & 8.88 (12.64) & 20.42 (22.81) & 9.79 (26.84) & 0.82 (0.75) & -0.11 (1.21) & -1.05 (1.38) & 170 \\
\addlinespace[0.3em]
\hdashline
\addlinespace[0.3em]
Top & Behavior & Top-Down Management Structure (トップダウン経営体制) & A management style in which the president or chairman makes decisive decisions and unilaterally directs the organization. (社長や会長が強力に意思決定を行い、組織を一方的に統率する経営スタイル。) & 17.67 (6.97)** & 13.64 (12.35) & 6.49 (17.79) & 0.38 (0.41) & 0.62 (0.66) & 0.06 (0.91) & 527 \\
\addlinespace[0.3em]
\hdashline
\addlinespace[0.3em]
Top & Behavior & Top-Down Policy Shift by Senior Management (経営トップのトップダウン方針転換) & A behavioral pattern in which senior management abruptly changes policies and organizational structures with a single directive, making decisions in a top-down manner. (経営トップが一声で方針や組織体制を急激に変更し、トップダウンで意思決定を行う行動パターン。) & 2.22 (6.92) & 12.15 (11.93) & 25.57 (20.55) & -0.40 (0.41) & 0.29 (0.63) & -0.63 (1.06) & 596 \\
\bottomrule
\end{tabular}
\end{adjustbox}
\vspace{1mm}
\begin{minipage}{\textwidth}
\footnotesize
``Type.'' refers to the leader type (Top or Non-top), and ``Char.'' refers to the leader characteristic (Ability, Attitude, or Behavior). Translated Translated Topic Name (Original Topic Name) and Translated Translated Topic Description (Original Topic Description) are automatically generated English translations produced by GPT-4.1-mini, given the original Japanese topic name and description as input.\textit{Notes:} For each entry, the reported value is the estimated coefficient, with the standard error shown in parentheses.$^{*}$, $^{**}$, and $^{***}$ indicate statistical significance at the 10\%, 5\%, and 1\% levels, respectively. $t$ denotes the annual reviewer-count threshold (minimum number of reviewers per firm-year). $N$ denotes the number of posts assigned to each topic. Topics are included only if they are statistically significant at the 5\% level for both outcomes in at least 0 different thresholds among $t\in\{5,10,15\}$. Topic names may be manually renamed based on the topic descriptions for interpretability.
\end{minipage}
\end{table*}

\end{document}